\documentclass[letterpaper, 10 pt, conference]{ieeeconf}
\usepackage[noadjust]{cite}

\usepackage{times}
\usepackage{epsfig}
\usepackage{graphicx}
\usepackage{amsmath}
\usepackage{amssymb}
\usepackage{caption}
\usepackage{subcaption}
\usepackage{soul}
\usepackage{booktabs}
\usepackage{algorithm}
\usepackage[noend]{algpseudocode}
\usepackage[pagebackref=true,breaklinks=true,letterpaper=true,colorlinks,bookmarks=false]{hyperref}
\usepackage{bm}

\renewcommand{\baselinestretch}{.97}

\IEEEoverridecommandlockouts
\title{\LARGE \bf Learning visual policies for building 3D shape categories}

\author{Alexander Pashevich$^{*}$\\
  Inria$^{\dagger}$\\
  \and
  Igor Kalevatykh$^{*}$\\
  Inria$^{\ddagger}$\\
  \and
  Ivan Laptev\\
  Inria$^{\ddagger}$\\
  \and
  Cordelia Schmid\\
  Inria$^{\ddagger}$\\
}

\begin{document}
\maketitle

\thispagestyle{empty}
\pagestyle{empty}

\renewcommand*{\thefootnote}{\fnsymbol{footnote}}
\footnotetext[1]{Equal contribution.}
\footnotetext[2]{University Grenoble Alpes, Inria, CNRS, Grenoble INP, LJK, 38000 Grenoble, France.}
\footnotetext[3]{Inria, \'{E}cole normale sup\'{e}rieure, CNRS, PSL Research University, 75005 Paris, France.}

\maketitle
\thispagestyle{empty}
\pagestyle{empty}

\begin{abstract}
  %% Motivation (2-3):
  Manipulation and assembly tasks require non-trivial planning of actions depending on the environment and the final goal.
  Previous work in this domain often assembles particular {\em instances} of objects from known sets of primitives.
  %% Solution (2-3):
  In contrast, we aim to handle varying sets of primitives and to construct different objects of a {\em shape category}.
  Given a single object instance of a category, e.g.~an arch, and a binary shape classifier, we learn a visual policy to assemble other instances of the same category.
  %% Contribtuions (2-3):
  In particular, we propose a disassembly procedure and learn a state policy that discovers new object instances and their assembly plans in state space.
  We then render simulated states in the observation space and learn a heatmap representation to predict alternative actions from a given input image.
  %% Results (2-3):
  To validate our approach, we first demonstrate its efficiency for building object categories in state space.
  We then show the success of our visual policies for building arches from different primitives.
  Moreover, we demonstrate (i) the reactive ability of our method to re-assemble objects using additional primitives and (ii) the robust performance of our policy for unseen primitives resembling building blocks used during training.
  Our visual assembly policies are trained with no real images and reach up to 95\% success rate when evaluated on a real robot.
\end{abstract}

\section{Introduction}
\label{sec:intro}

Our daily physical activities such as cooking, dressing or navigation require complex sequences of actions which people successfully and seamlessly plan based on sensory input. Action planning typically depends on the goal and constraints provided by the environment. Despite extensive prior work, existing autonomous agents are still far from the human-level planning performance, especially in unknown and cluttered environments~\cite{Ebert2018VisualFM, Wang2019LearningRM}.

Action planning is a hard problem due to the large action spaces, exponential complexity and partial observability~\cite{rlblogpost, Jaderberg2016ReinforcementLW}. %\cite{offir2020why}
To simplify the problem, existing work on task planing~\cite{garrett2018sampling,nair2018overcoming,srivastava2014combined,toussaint2015logic} typically operates in the state space assuming the full knowledge of the environment.
While such an assumption can be practical in structured and controlled environments, full state reconstruction for common scenes remains a highly challenging problem~\cite{lepetit2018}. 
Arguably, the precise recovery of scene parameters such as its geometry, composition, friction coefficients, etc., is more difficult than the primary planning task itself. 
It is therefore desirable to design sensor-based planning policies that do not rely on explicit scene geometry and full state estimation.

\begin{figure}
  \centering
  %[trim=left bottom right top, clip]
  \includegraphics[trim=120 0 150 50, clip, width=.95\linewidth]{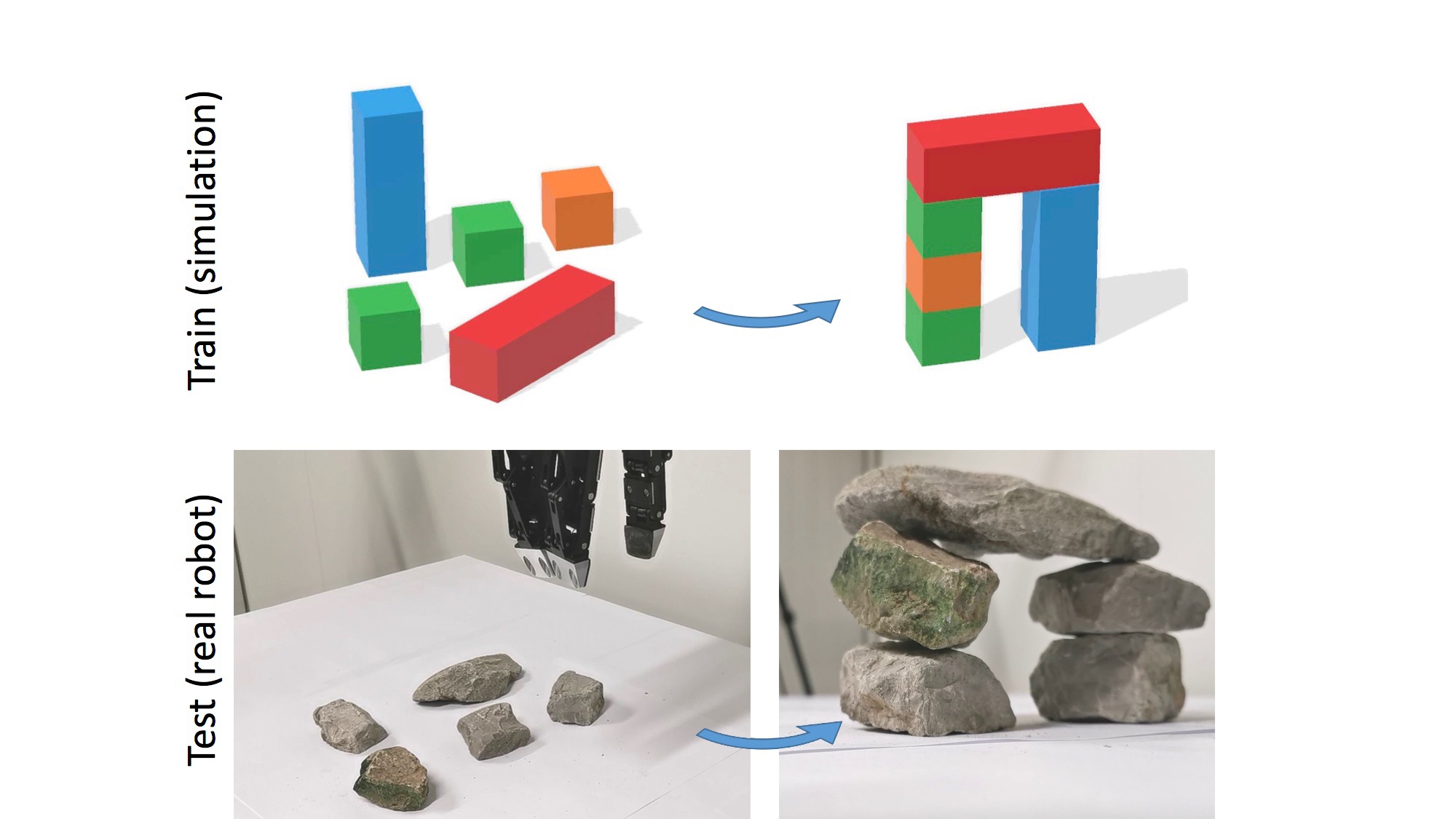}
  \caption{\small Primitives on the left are assembled by our learned policy into arches on the right. We assemble objects of similar shapes in the simulator and learn visual manipulation policies that can build real 3D shapes from unseen primitives resembling building blocks used during training. }
  \vspace{-.3cm}
  \label{fig:teaser}
\end{figure}

\begin{figure*}
  \vspace{-.4cm}
  \centering
  %[trim=left bottom right top, clip]
  \includegraphics[trim=30 110 30 40, clip, width=\linewidth]{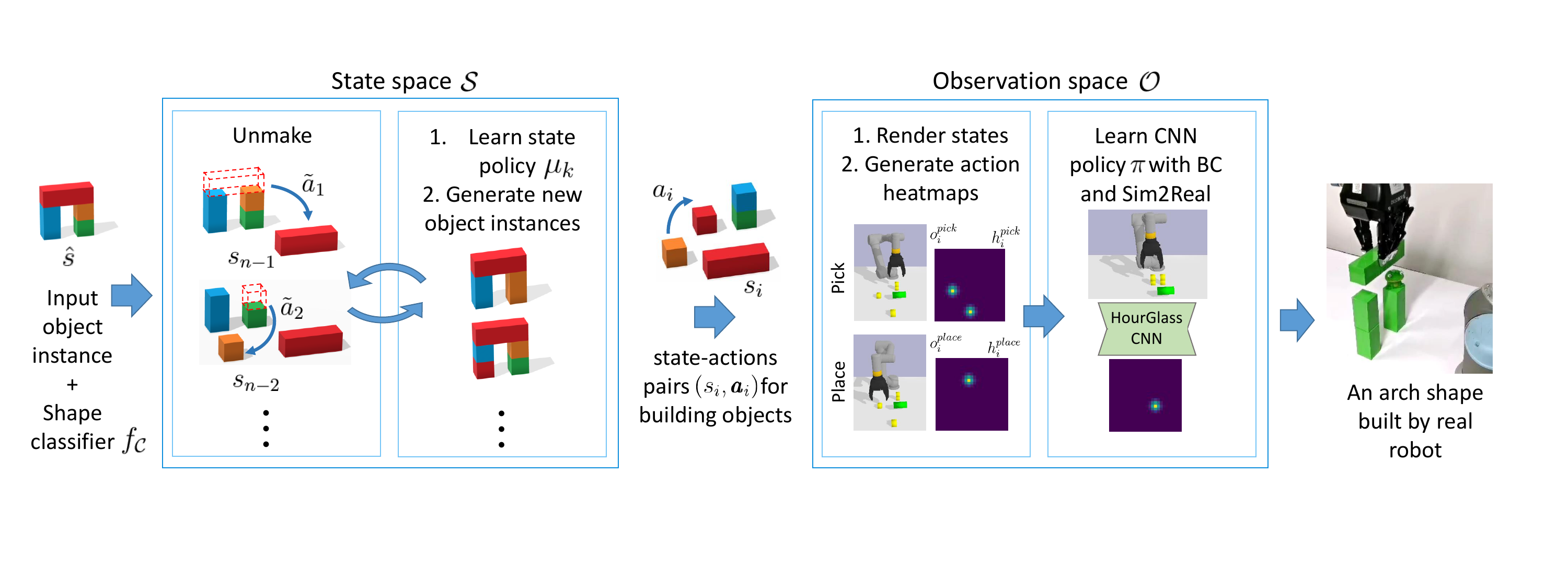}
  \caption{\small Method overview. Given an example object and a shape classifier on the left, our method generates new objects with similar shape and discovers action sequences for building these objects in the state space. Using a large set of generated state-actions pairs $(s_i,\bm{a}_i)$, we render states $s_i$ as realistic observations $o_i$. We also generate 2D heatmaps $(h_i^{\text{pick}},h_i^{\text{place}})$ encoding source and target locations and orientations of one or several primitives. 
  Heatmaps can represent multiple hypothesis for the next action when several identical primitives are used or multiple object instances can be assembled.
  Positions on our 2D heatmaps correspond to positions on the 2D surface of a table, hence, the identified local maxima on heatmaps can be used to control the robot. 
  As the last step of our method, we train a Behaviour Cloning policy $\pi$ to predict $h^{\text{pick}}$ and $h^{\text{place}}$ from observations. We train policies with HourGlass CNN~\cite{newell2016stacked} and sim2real augmentation~\cite{learningsim2real2019}. The learned policy is directly transferred to a real robot which assembles 3D objects from primitives on the table.}
  \label{fig:overview}
  \vspace{-.5cm}
\end{figure*}

Vision-based control policies have recently become popular for robotic manipulation~\cite{agrawal2016learning,Levine2015End-to-EndPolicies,Zeng2019TossingBotLT} and navigation~\cite{codevilla2018end,gandhi2017learning}.
While this line of work shows promise, it has mostly been applied to the low-level motion planning such as predicting next motion direction.
In our work we aim to learn visual policies for high-level task planning.
Given visual input, our policies generate sequences of picking, rotating and placing actions for building 3D shapes.

Object manipulation has a long history in robotics.
In particular, assembling objects from a given set of primitives has been addressed, for example, in~\cite{popov-manipulation, Chen_2019_ICCV}.
Prior work in this domain often aims to build particular object instances for which the structure is pre-defined~\cite{suarez2018can}, specified by demonstrations~\cite{Huang2018NeuralTG, onshot-imitation} or given by a goal image~\cite{janner2019reasoning}.
Here we go further and learn to assemble different objects of a shape category.
Such a task is significantly more complex compared to building particular object instances as it requires generalization to varying sets of building primitives.
Moreover, we show empirically that our method is able to generalize to new primitives unseen during training (see Fig.~\ref{fig:teaser}).

Our approach contains two stages as illustrated in Fig.~\ref{fig:overview}.
In the first stage we discover new object instances and learn their assembling policies in state space.
To this end, we propose a disassembling procedure and generate assembly trajectories by (a)~unbuilding objects and (b)~reverting action sequences.
We then learn a value function and apply it to build new object instances.
We iterate the disassembling and learning steps to obtain a policy assembling a 3D shape.

In the second stage we assemble objects given images of observed scenes.
We render states from assembly trajectories obtained in the first stage and learn visual assembly policies with Behaviour Cloning (BC)~\cite{Pomerleau1989}.
To enable predictions of multiple valid actions at any given time, we propose a heatmap representation for the output of visual policies.
While all our policies are learned in a simulated environment, we enable their direct transfer to a real robot using sim2real augmentation~\cite{learningsim2real2019}.

As main contributions of this work, we 
(i)~propose a novel disassembly algorithm for building shape categories in state space,
(ii)~design and learn visual policies with heatmap outputs to address multi-modality of predicted actions,
(iii)~demonstrate a successful application of the method to a new task of building shape categories on a real UR5 robot.
Moreover, our policies are learned with no human demonstrations, can re-assemble partially built objects, and adapt to unseen primitives resembling building blocks used during training.

\vspace{-.2cm}
\section{Related work}
\label{sec:rw}
\vspace{-.1cm}

Assembly tasks such as constructing IKEA furniture~\cite{suarez2018can} remain to be a hard robotics challenge.
Learning-based methods usually address simpler tasks such as cube stacking~\cite{SACX, ModulatedPH}.
Duan et al.~\cite{onshot-imitation} use demonstrations and attention modules to build a tower instance shown by an expert.
Janner et al.~\cite{janner2019reasoning} learn an object-centric representation of the scene to reproduce a tower instance from a goal image with an MPC-like control. Huang et al.~\cite{Huang2018NeuralTG} train a Graph Network to build a tower instance specified by a demonstration.
We go beyond specific object instances and aim to assemble multiple objects from a given shape category, such as an arch. Moreover, we learn to build objects from different sets of possibly unseen primitives.
To facilitate the learning, we propose to use disassembling to generate assembly trajectories. 
While the idea of reversible actions has been explored e.g., in~\cite{florensa2017reverse, Nair2018TimeRA, Sukhbaatar2017IntrinsicMA, hosu2016playing}, our method differs from the work on disassembling object instances~\cite{disassembly2019} by computing multiple disassembly paths and accounting for alternative valid actions.

Our work is related to methods of Task and Motion Planing (TAMP)~\cite{garrett2015ffrob,lozano2014constraint,srivastava2014combined,toussaint2015logic}. Long-term task planning prohibits costly rollouts, hence, TAMP methods deploy preconditions and postconditions to actions and optimize symbolic planners~\cite{FIKES1971189}. While some of these methods solve impressive tasks, conditions require manual and task-specific design~\cite{Edelkamp2004PDDL2, He2015towards, paxton2019visual}. Moreover, TAMP methods typically operate in state-space~\cite{Garrett2018FFRobLS}, hence, their generalization to sensor-based input in the real world requires non-trivial scene understanding~\cite{Dantam2018AnIC}.
In our work we learn visual policies and directly predict control sequences from image inputs.

Convolutions Neural Networks (CNNs) have significantly advanced visual recognition~\cite{He2016DeepRL, fasterrcnn, newell2016stacked} and robotics, for example in tasks such as tossing objects~\cite{Zeng2019TossingBotLT}, cube stacking~\cite{popov-manipulation}, grasping~\cite{Lampe2013AcquiringLearning} and opening doors~\cite{Gu2016}.
Direct methods for visual control avoid explicit scene reconstruction and derive actions directly from image observations. 
Such methods typically use Reinforcement Learning (RL)~\cite{kobber-survey} with auxiliary rewards~\cite{SACX} or Imitation Learning (IL)~\cite{robot-learning-survey, cad2real} relying on large amounts of demonstrations~\cite{Zhang2017DeepIL, Rahmatizadeh2017VisionBasedMM}. The complexity of problems addressed by direct methods is typically limited by the task length and the number of manipulated objects~\cite{SACX, Zhang2017DeepIL, Rahmatizadeh2017VisionBasedMM}.
Indirect methods first estimate scene parameters~\cite{servoing, Andrychowicz2017HindsightER} such as object positions and orientations, and then deploy state-based planning strategies.
Scene reconstruction from images, however, might be a more challenging task than solving a control task itself~\cite{pinto-grasp, lepetit2018}.
We avoid drawbacks of direct and indirect methods and first solve the task in the simulated state space.
We then use obtained solutions as automatic supervision for learning visual policies in the observation space.
Inspired by~\cite{Levine2015End-to-EndPolicies, Mahler2017DexNet2D}, we render states and train visual policies for a real robot using BC~\cite{Pomerleau1989} and sim2real~\cite{learningsim2real2019, sim2real2017tobin}.
\vspace{-.1cm}
\section{Approach}
\label{sec:approach}
\vspace{-.1cm}

We address the problem of building a 3D object shape by manipulating a set of available primitives with a robot.
The configuration of the primitives on the table defines the state, $s \in \mathcal{S}$.
We assume to have access to a shape classifier function $f_{\mathcal{C}}: \mathcal{S} \rightarrow \{0, 1\}$.
We define the shape as a subset of the state space, $\mathcal{C} = \{s \in \mathcal{S} \vert f_{\mathcal{C}}(s) = 1\}$.
Given $f_{\mathcal{C}}$ and a single shape instance $\hat{s} \in \mathcal{C}$, our method learns a visual policy $\pi$ that generates a robot action given a camera observation $o \in \mathcal{O}$.
The resulting sequence of actions is then used to assemble a shape instance from available primitives.
Our policy operates in the observation space $\mathcal{O}$, i.e.,~it only has access to the image of a current scene before deciding on the next action.
Moreover, the policy is expected to build new object configurations from unseen primitives that resemble building blocks used during training.

\vspace{-.1cm}
\subsection{Overview of the method}
\vspace{-.1cm}
Our method has two stages: (i) generating action sequences in the state space and (ii) learning visual policies in the observation space. 
We use a simulated environment for both stages, however, our visual policies are trained with sim2real data augmentation~\cite{learningsim2real2019} and directly transfer to the real robot.
The overview of the method is presented in Fig.~\ref{fig:overview}.

The first stage aims to find new shape instances and to construct action sequences for building them. It takes as input one 3D shape and a shape classifier. 
We propose to use an {\em unmake} procedure to generate valid action sequences. We disassemble the given shape instance and interpret the resulting sequences as reversed assembly demonstrations. We disassemble objects in multiple ways to find all assembly actions that are possible in the same state.
We refer to the shape classifier $f_{\mathcal{C}}$ as a sparse reward signal which we use to learn a state-value function $V_k$. We generate new shape instances using a state policy $\mu_k$ which is greedy with respect to the learned $V_k$. For a fixed number of iterations, we repeat the unmake procedure using the new set of instances and train an updated value function $V_{k+1}$.
This part of our approach is described in Section~\ref{sec:statespace}.

In the second stage we learn a visual policy that infers appropriate actions from image observations.
We convert states $s_i$ into observations $o_i=R(s_i)$ using a graphics renderer $R$ and train a CNN policy $\pi$ with Behaviour Cloning (BC).
Given the assembly trajectories produced by the first stage, each state is associated with a valid set of actions $\bm{a}_i$ that we turn into a heatmap
$h_i$ to predict all possible actions simultaneously.
Then $\pi$ is trained in a supervised manner using observation-heatmap pairs $(o_i,h_i)$.
This part of our approach is described in Section~\ref{sec:observationspace}.

\vspace{-.1cm}
\subsection{Building objects in state space}
\vspace{-.1cm}
\label{sec:statespace}

We define the full state of our environment by the vector \mbox{$s=(x^1,\ldots,x^m)^{\top}$} with $x\in\mathbb{R}^{12}$ representing parameters for $m$ primitive shapes (our building blocks) in the scene.
Each primitive $x$  is defined by three position coordinates, three orientation angles in the 3D space, three spatial extents (width, height and depth), and three color channels (the building order may depend on the color, see Section~\ref{sec:tasks}).
A robot action $a \in \mathcal{A}$ corresponds to a high-level skill of picking, placing and rotation of a primitive: $a = (x, p, o)$ where $x \in \{x^1, \ldots, x^m\}$ is the primitive to pick, $p \in \mathbb{R}^3$ is the position to place it, and $o \in \mathbb{R}^3$ is the orientation $x$ is placed in.
We restrict orientations of primitives to the three axis-parallel directions and assume that all primitives are located on the surface of a table or on top of each other.
Assuming access to the simulator $T$, applying action $a_t$ in the state $s_t$ would result into $s_{t+1} = T(s_t, a_t)$.

Building an object from a given set of primitives requires finding an appropriate sequence of actions \mbox{$a_{1:n}=\{a_1,\ldots,a_n\}$} that transforms an initial state $s_0$ into the desired shape state $s_n \in \mathcal{C}$.
Finding a correct sequence $a_{1:n}$ is not trivial even in the state space. Given the large space of possible actions and the exponential growth of the number of action sequences depending on $n$, the naive brute-force search works only for building simple objects.

\textit{Making objects by unmaking.}
For an object example defined by the state $\hat{s} \in \mathcal{C}$ we propose to find valid building sequences $a_{1:n}$ via disassembling or {\em unmaking} $\hat{s}$.
We first find a sequence of unmake actions $\tilde{a}_{1:n}=\{\tilde{a}_1,\ldots,\tilde{a}_n\}$, where $\tilde{a}_i$ moves a random non-blocked primitive on an object\footnote{By ``non-blocked primitives on objects`` we refer to primitives that are not on the table surface and that can be freely lifted above their current positions. Such primitives can be easily derived from $s$.} to a random non-occupied location on the table.
If $m$ is the number of primitives constituting object $\hat{s}$, we can disassemble $\hat{s}$ by a sequence of $n=m-1$ unmake actions.
We call an action invertible if for any action $\tilde{a}$ such that $T(s_i, \tilde{a})=s_j$ there exists an inverse action $a=\tilde{a}^{-1}$ such that $T(s_j, a)=s_i$. 
We assume our unmake actions to be invertable and
obtain a valid sequence of actions for building $\hat{s}$ as $a_{1:n}=\{\tilde{a}^{-1}_n,\ldots,\tilde{a}^{-1}_1\}$.

\begin{figure}
  \centering
    %[trim=left bottom right top, clip] 
    \includegraphics[trim=0 15 670 230, clip, width=0.9\linewidth]{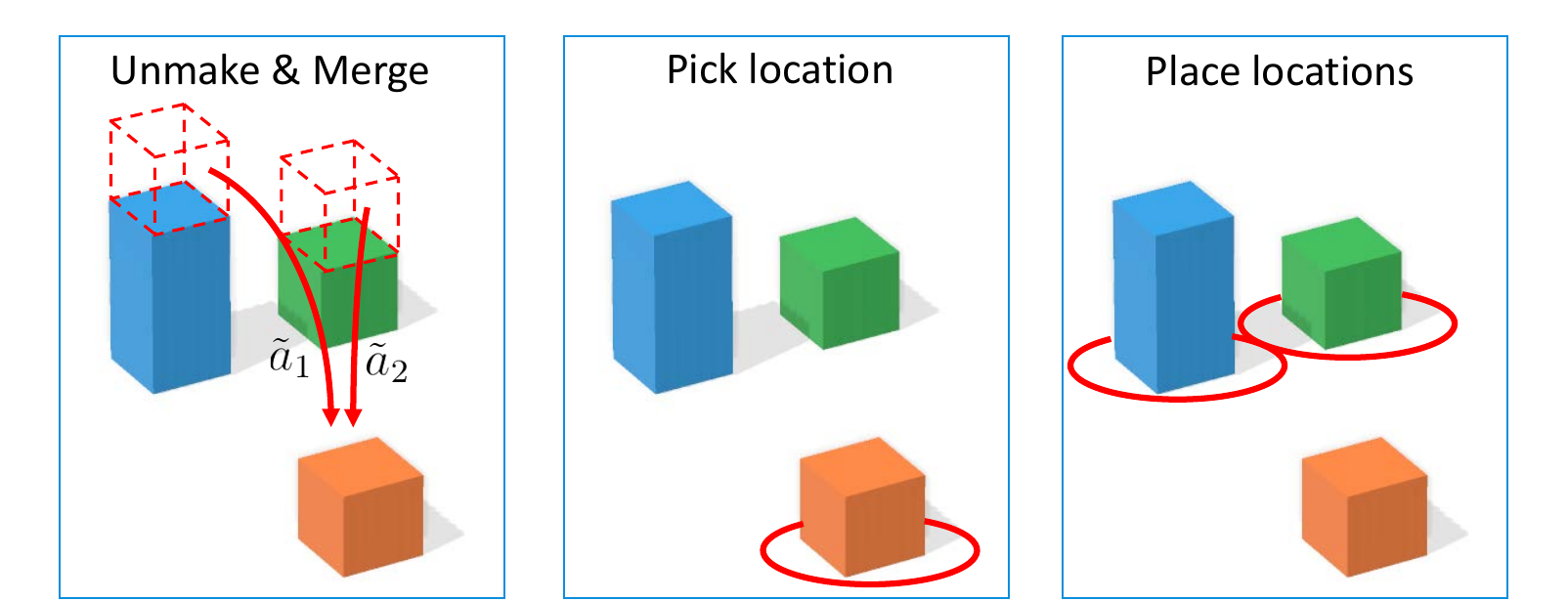}
  \caption{\small The proposed unmake procedure disassembles objects in multiple ways (left) and generates set of pick \& place actions available from each state (right). For visual policies, we use the heatmap representation to predict all the actions simultaneously. \vspace{-0.3cm}
  }
  \label{fig:heatmap_explain}
  \vspace{-0.3cm}
\end{figure}

We use the above randomized unmake procedure to generate a large and diverse set of valid action sequences.
While each action $a_i$ in such sequences is associated with a single initial state $s_i$, there might be \emph{multiple correct assembly actions} in the state $s_i$.
For example, when building an arch, the same cube could be placed both to the left and to the right pillars (see Fig.~\ref{fig:heatmap_explain}).
To each state $s_i$, we associate a set of actions $\bm{a}_i$ available in all states which differ only in positions of primitives located on the table surface (denoted as $\approx$), namely, $\bm{a}_i = \{a_j \vert s_j \approx s_i\}$.

Given $M$ action sequences of length $N$ we collect a dataset $\mathcal{D}_{\mu}$ with state-actions pairs $\mathcal{D}_{\mu}=\{(s_i,\bm{a}_i)\}_{i=1,\ldots,M*N}$ that can be readily used to train a policy for assembling an object instance defined by state $\hat{s}$.

\textit{Finding new object instances.}
To generalize our policies to build new objects of the similar shape given any set of primitives, we construct new instances through learning a value function.
The value function $V_k: \mathcal{S} \rightarrow \mathbb{R}$ estimates the sum of discounted rewards $r: \mathcal{S} \rightarrow \mathbb{R}$ under an assembly policy $\mu_k: \mathcal{S} \rightarrow \mathcal{A}$ where $V_k(s) = \mathbb{E}_{\mu_k} \left[\sum_j \gamma^j r(s_{t+j}) | s_t = s\right]$ and $\gamma \in [0, 1)$ is the discount factor.
We define the reward with the shape classifier $r(s_t) = f_{\mathcal{C}}(s_t)$ which makes the value function learn the actions required to build the shape.
Given a learned value function $V_k$, we deduce the assembly policy by choosing $a_{t+1}$ as  $\mu_{k+1}(s_t) = \arg\max_{a} V_k(T(s_t, a))$.
We sample an initial state $s_0$ with a random set of primitives and apply $\mu_{k+1}$ to
choose an action sequence $a_{1:n}$ resulting in an object of the desired shape: $f_{\mathcal{C}}(s_n) = 1$.
We record all instances of the target shape found by assembling is a set $\mathcal{C}^V_k$.

\textit{Learning value function.}
The value function $V_k$ is learned iteratively using instances found with the greedy policy $\mu_{k-1}$. In the first stage of training, we learn $V_0$ using the input instance $\hat{s}$ only.
We run the unmake procedure for all discovered instances in $\mathcal{C}^V_k$, 
Given a sequence of state-actions pairs $\{(s_1,\bm{a}_1),\ldots,(s_n,\bm{a}_n)\}$, the value function estimate for state $s_i$ is $\hat{V}(s_i) = \gamma^{n - i}$.
We also estimate values of states obtained by applying random disassembly actions $\tilde{a}_j$ to trajectory states $s_i$: $s_i^j = T(s_i, \tilde{a}_j)$.
The value estimate $\hat{V}(s_i^j)$ is known if there exists $s_j \in \{s_1, \ldots, s_n\}$ such that $s_i^j \approx s_j$. Otherwise, we set the value as $\hat{V}(s_i^j) = \gamma \hat{V}(s_i)$ which means that we can reach $s_i$ from $s_i^j$ using a single assembly action $\tilde{a}_j^{-1}$.
We record all state-value pairs to the dataset $\mathcal{D}_V^k$ and learn the value function by minimizing the loss $\hat{\eta}_k = \arg\min_{\eta} \textrm{MSE}(V_k(s_i), \hat{V}(s_i))$,
where $V_k$ is implemented as a fully connected neural network with parameters $\hat{\eta}_k$ and MSE is the mean square error.
Once the training is converged, we discover new shape instances with $\mu_k$, unmake them, recollect $\mathcal{D}_V^{k+1}$ and learn $V_{k+1}$.
After $K$ phases, we run the unmake procedure on the set of discovered instances $\mathcal{C}^V_K$ and record all state-actions pairs to the dataset $\mathcal{D}_{\mu}$.
The overview of our approach in the state space is illustrated by the left part of Fig.~\ref{fig:overview} and lines 14-28 of Algorithm~1 of~\cite{3Dshapes2020}.

\vspace{-.1cm}
\subsection{Learning in observation space}
\vspace{-.1cm}
\label{sec:observationspace}

We want to learn a visual policy $\pi$ for assembling objects from diverse sets of primitives by a real robot. The sole input of the policy is the camera observation of the scene.
We learn the image-action association with a supervised learning approach, Behaviour Cloning (BC)~\cite{Pomerleau1989}, where we obtain supervision with solutions found in the state space.
Given the dataset $\mathcal{D}_{\mu}$ with state-actions pairs $\{(s_i, \bm{a}_i)\}$, we use a {\small{\texttt{pybullet}}}~\cite{Courmans2016} graphics renderer $R$ to generate an RGB-D image $o_i=R(s_i)$ for each state $s_i$ in $\mathcal{D}_{\mu}$.
In order to allow multiple actions for each observation $o_i$, we generate an action-heatmap $h_i \in \mathcal{H}$ given the list of actions $\bm{a}_i = \{a_i^1, \ldots, a_i^l\}$. We record the observation-heatmap pairs to the dataset $\mathcal{D}_{\pi}$.
The policy $\pi: \mathcal{O} \rightarrow \mathcal{H}$ is implemented as a CNN and is trained to predict correct action-heatmaps $\pi(o_i)=h_i$ for all $(o_i, h_i) \in \mathcal{D}_{\pi}$.
We show an advantage of the heatmaps-based architecture over a network that directly predicts positions and orientations in Section~\ref{sec:results:vp:archi}.

For the task of building objects with a real robot, we consider separate {\em pick} and {\em place} actions that are parameterized by positions and orientations of primitives on the 2D plane of a table. For simplicity, we assume that the elevation of a primitive above the table can be estimated by external means such as an overhead depth camera or a force-feedback sensor of the robot arm.
We define the output of our policy by distributions over 2D positions on the table plane and 3 possible orientations of primitives on the table. We represent such distributions as heatmaps $(h^{\text{pick}},h^{\text{place}})$ corresponding to the source and target parameters of primitives. Our heatmaps are 4-channel images with one channel representing position distribution and three other channels representing orientation.
The placing positions and orientations might depend on the picked primitive, hence, we predict pick and place action-heatmaps sequentially.

We render separate observations $(o^{\text{pick}}, o^{\text{place}})$ for pick and place action-heatmaps $(h^{\text{pick}},h^{\text{place}})$ respectively.
For each list of pick-place actions $\bm{a}_i$, we render the observation $o_i^{\text{pick}} = R(s_i)$ and define $h_i^{\text{pick}}$ as an image with Gaussian distributions around positions of all picked objects by $\bm{a}_i$.
For each picked object, we render the observation $o_i^{\text{place}} = R(s_i)$ with the robotic arm picking this object and create the heatmap $h_i^{\text{place}}$ with all possible placing positions in $\bm{a}_i$ (see Fig.~\ref{fig:heatmap_explain}). We record all observation-heatmap pairs in the dataset $\mathcal{D}_{\pi}$.

Heatmaps are often used as CNN outputs in tasks such as human pose estimation~\cite{newell2016stacked,Wei_2016_CVPR} and segmentation~\cite{long2015fully}. We follow~\cite{newell2016stacked} and use  a  HourGlass CNN architecture for heatmap prediction $\pi(o) \mapsto h$.
Given $\mathcal{D}_{\pi}$, we train $\pi$ by minimizing the loss $\hat{\theta} = \arg\min_{\theta} \textrm{MSE}(\pi_{\theta}(o_i), h_i)$,
where $\hat{\theta}$ are parameters of the HourGlass CNN.
We obtain parameters for the pick and place actions by maximizing the obtained location heatmaps over the 2D space and then maximizing orientations over the 3 channels at the selected location.
The overview of our policy learning is illustrated by the right part of Fig.~\ref{fig:overview} and lines 30-41 of Algorithm~1 of~\cite{3Dshapes2020}.

\begin{figure}
  \centering
  \begin{subfigure}{.4\linewidth}
    \centering
    \includegraphics[trim=0 50 0 50, clip, width=\linewidth]{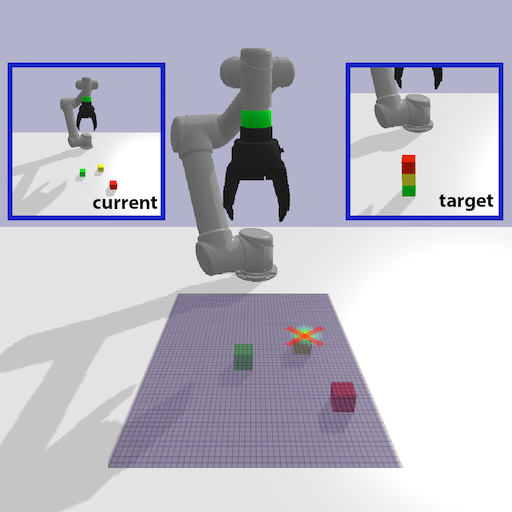}
  \end{subfigure} \hspace{.2cm}
  \begin{subfigure}{.4\linewidth}
    \centering
    \includegraphics[trim=0 50 0 50, clip, width=\linewidth]{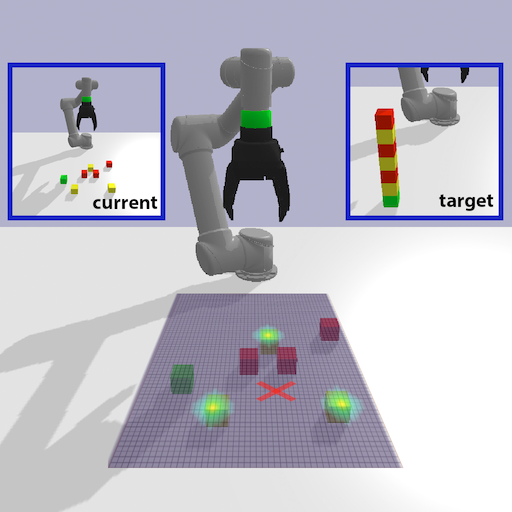}
  \end{subfigure}
  \caption{\small Visualization of predictions of visual policies using HourGlass (green blobs) and ResNet-18 (red crosses) architectures. The policies are trained to build towers and should start by placing a yellow cube on the green one. ResNet predicts the correct picking location when all the cubes have distinct colors (left). Once identical yellow cubes are introduced to the scene (right), ResNet fails to choose between them and predicts an averaged location. HourGlass locates all the cubes correctly in both cases.}
  \label{fig:tower}
  \vspace{-.7cm}
\end{figure}

% \vspace{-.2cm}
\section{Experiments}
% \vspace{-.2cm}
\label{sec:experiments}
In this section we evaluate our approach both in simulation and on
a real UR5 robot. We start with implementation details  in
Section~\ref{sec:impdetails} and present tasks used for evaluation in Section~\ref{sec:tasks}.
Section~\ref{sec:results:vf} confirms the importance of learning the state-value
function for efficient task solving.
Section~\ref{sec:results:vp} evaluates visual policies trained to solve tasks in the observation
space. We validate our proposed network architecture and highlight the importance
of the disassembling procedure.
Additional qualitative results are available in the Appendix of~\cite{3Dshapes2020} and on the project web-page~\cite{pashevich20buildingWeb}.

\begin{table}
\small
\setlength\tabcolsep{3.2pt} 
\centering
\begin{tabular}{lccc}
\toprule
\textit{Arch, state space} & 3U & 4U & 5U \\
\midrule
Random & \footnotesize 3.4e3 $\pm$ 5.0e3 & \footnotesize 1.5e4 $\pm$ 1.2e4 & \footnotesize 6.1e4 $\pm$ 9.6e4 \\
MCTS~\cite{Coulom2006EfficientSA} & \footnotesize 6.3e2 $\pm$ 4.7e2 & \footnotesize 7.0e3 $\pm$ 6.2e3 & \footnotesize 1.6e4 $\pm$ 2.4e4 \\
Ours (state policy)$^*$ & \footnotesize \textbf{4.0e0} $\pmb{\pm}$ \textbf{0.7e0} & \footnotesize \textbf{5.7e0} $\pmb{\pm}$ \textbf{0.7e0} & \footnotesize \textbf{6.9e0} $\pmb{\pm}$ \textbf{1.8e0} \\
\midrule
Oracle & \footnotesize 4.0e0 $\pm$ 0.7e0 & \footnotesize 5.7e0 $\pm$ 0.7e0 & \footnotesize 6.4e0 $\pm$ 1.0e0 \\
\bottomrule
\end{tabular}
\caption{\small Average amount and standard deviation of steps required to build an arch shape for our method, random exploration and MCTS. $^*$The simulation steps used for training of our method are not included.}
\vspace{-.2cm}
\label{table:arches:vp}
\end{table}

\begin{table}[ht]
\vspace{-.2cm}
\small
\centering
\begin{tabular}{lccc}
\toprule
\textit{Tower, simulation} & 3 cubes & 5 cubes & 7 cubes \\
\midrule
ResNet-18~\cite{He2016DeepRL} & 99.2 & 1.4 & 0.0 \\
HourGlass~\cite{newell2016stacked} & \textbf{99.6} & \textbf{97.2} & \textbf{94.8} \\
\midrule
\textit{Tower, real} & 3 cubes & 5 cubes & 7 cubes \\
\midrule
HourGlass~\cite{newell2016stacked} & 95.0 & 90.0 & 90.0 \\
\bottomrule
\end{tabular}
\caption{\small Success rates of visual policies (in percent) trained to build tower instances of 3, 5 and 7 cubes. On the real robot, the policies are evaluated using 20 trials.}
\vspace{-.6cm}
\label{table:towers}
\end{table}

\begin{table}
\vspace{-.4cm}
\small
\setlength\tabcolsep{3.2pt} 
\centering
\begin{tabular}{lccc}
\toprule
\textit{Arch, simulation} & 3U & 4U & 5U \\
& \multicolumn{3}{c}{\textit{category}} \\
\midrule
Single unmake trajectory & 98.6 & 92.8 & 69.6 \\
Multiple unmake trajectories & \textbf{99.0} & \textbf{98.8} & \textbf{95.6} \\
\bottomrule
\end{tabular}
\caption{\small Success rates of visual policies trained to build the arch shape category
  of different heights. The policies are trained on trajectories obtained by disassembling the same object once or multiple times.}
\label{table:arches:merge}
\vspace{-.2cm}
\end{table}

\begin{table}
\small
\setlength\tabcolsep{3.2pt} 
\centering
\begin{tabular}{lccc|ccc}
\toprule
\textit{Arch, simulation} & 3U & 4U & 5U & 3U & 4U & 5U \\
& \multicolumn{3}{c|}{\textit{instance}} & \multicolumn{3}{c}{\textit{category}} \\
\midrule
Instance policies & 99.4 & 97.8 & 96.8 & 61.4 & 54.4 & 32.6 \\
Uni-height policies & \textbf{99.6} & \textbf{98.2} & \textbf{98.0} & \textbf{99.0} & \textbf{98.8} & \textbf{95.6} \\
Multi-height policy & 97.4 & 96.4 & 95.4 & 97.8 & 95.4 & 94.0 \\
\bottomrule
\end{tabular}
\caption{\small Success rates of visual policies trained by disassembling only the input
instance (top) and instances found by state policies. The state policies are trained on arches of the same height (middle) and arches of heights 3-5U (bottom). The policies assemble arches given a fixed set of primitives (left) or various configurations of primitives (right). While instance and uni-height policies need to be trained for each given arch height, a single multi-height policy can assemble arches of various heights.}
\label{table:arches:sim}
% \vspace{-.3cm}
\end{table}

\begin{table}
% \vspace{-.4cm}
\small
\setlength\tabcolsep{3.2pt} 
\centering
\begin{tabular}{lccc|ccc}
\toprule
\textit{Arch, real} & 3U & 4U & 5U & 3U & 4U & 5U \\
& \multicolumn{3}{c|}{\textit{normal scenario}} & \multicolumn{3}{c}{\textit{re-assembling}} \\
\midrule
Uni-height policies & \textbf{95.0} & \textbf{80.0} & \textbf{75.0} & \textbf{90.0} & 75.0 & 70.0 \\
Multi-height policy & 85.0 & \textbf{80.0} & \textbf{75.0} & \textbf{90.0} & \textbf{80.0} & \textbf{80.0} \\
\bottomrule
\end{tabular}
\caption{\small Success rates of visual policies assembling arches of different heights on a real robot. Over 20 trials, the policies are evaluated on a normal Arch task and a "re-assembling" scenario when the agent needs to adapt to additional primitives introduced to the scene after the arch had already been built.}
\label{table:arches:real}
\vspace{-.2cm}
\end{table}

\begin{table}
\small
\setlength\tabcolsep{3.2pt} 
\centering
\begin{tabular}{lccc}
\toprule
\textit{Arch, real} & set 1 & set 2 & set 3 \\
& \multicolumn{3}{c}{\textit{novel primitives}} \\
\midrule
Detect + MCTS~\cite{Coulom2006EfficientSA} & 60.0 & 35.0 & 20.0 \\
Our method & \textbf{90.0} & \textbf{80.0} & \textbf{55.0} \\
\bottomrule
\end{tabular}
\caption{\small Success rates of our method and an MCTS baseline for assembling arches using novel primitives on a real robot. The arches built with primitives from sets 1, 2 and 3 are shown in Fig.~\ref{fig:arch} (last column, top), Fig.~\ref{fig:arch} (last column, bottom) and Fig.~\ref{fig:teaser} (bottom) correspondingly.}
\label{table:arches:unseen}
\vspace{-.7cm}
\end{table}

\vspace{-.2cm}
\subsection{Implementation details}
\label{sec:impdetails}
\vspace{-.1cm}

We control a 6-DoF UR5 robotic arm with a 3 finger Robotiq gripper.
In simulation, we model the robot and its environment with the {\small{\texttt{pybullet}}} physics simulator~\cite{Courmans2016}.
Given the positions and orientations of primitives to be manipulated, we use
standard path planing
methods~\cite{lavalle2006-palnning-algorithms,sucan2012the-open-motion-planning-library}
to implement the pick and place actions. The elevation of primitives above the
table surface is obtained with a Microsoft Kinect-2 camera located above the
table.

Our observations are depth images recorded with another Kinect-2 camera placed in front of the robot arm. We use the same parameters for real and simulation cameras (location and calibration).
Visual policies receive the depth image and color segmentation masks corresponding to the colors of primitives.
While visual policies in simulation have average errors of less than 5mm, the
sim2real gap increases this value up to 2-3cm on the real robot. Given that
stacking multiple primitives requires high precision, we apply a correction procedure using the depth camera as explained in Appendix of~\cite{3Dshapes2020}.

The neural network $V_{\eta}$ for Value Function has five fully-connected layers with 128 hidden units, ReLU activations and Batch Normalization~\cite{Ioffe2015BatchShift}. We train $V_{\eta}$ for 20 iterations of 30 epochs each using Adam and \mbox{LR=1e-3}.
The CNN $\pi_{\theta}$ contains one HourGlass~\cite{newell2016stacked} module
which we compare to ResNet~\cite{He2016DeepRL} in Section~\ref{sec:results:vp:archi}.
The spatial dimensions of HourGlass input and output are 256x256 and 64x64 pixels
respectively. We train $\pi_{\theta}$ using Adam and \mbox{LR=2.5e-4} for 50 epochs.
For both value and visual networks, we use datasets of size 200.000
value-state and heatmap-observation pairs correspondingly.
To enable the transfer of policies to the real robot, we use
sim2real~\cite{learningsim2real2019} to augment synthetic depth maps during
training.
Color segmentation masks are augmented with Bernoulli noise.
During rendering, we also randomize shapes of primitives by adding noise to spatial coordinates of points that define cube meshes.
We use 500 episodes for evaluation in simulation and 20 trials on the real robot.

\vspace{-.2cm}
\subsection{Tasks}
\label{sec:tasks}
\vspace{-.1cm}

\textit{Tower.}
The goal of the agent is to stack cubes in a specific order of colors (see
Fig.~\ref{fig:tower}). In the beginning of the task, green, yellow and red cubes of size 1 unit (1U) are randomly distributed on the surface of a table.
The unit corresponds to a physical size of 4.5 cm.
The lowest cube is always green, the rest of the tower is defined as alternating
yellow and red cubes. We use the Tower task to compare HourGlass and ResNet
architectures in Section~\ref{sec:results:vp:archi}.

\begin{figure*}
  \centering
  %[trim=left bottom right top, clip]
  % 3u sim
  \includegraphics[trim=110 50 180 140, clip, width=.1185\linewidth]{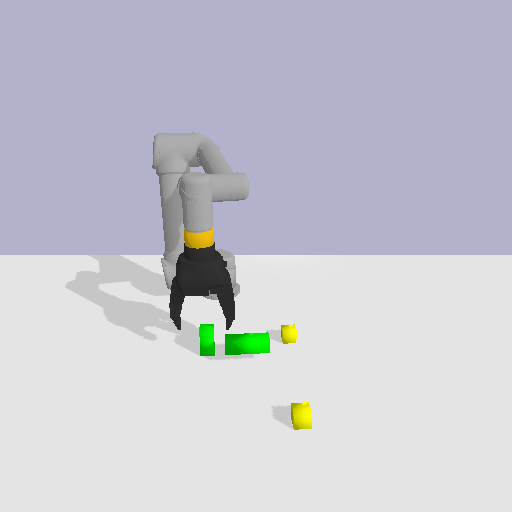}\hspace{-.06cm}
  \includegraphics[trim=110 50 180 140, clip, width=.1185\linewidth]{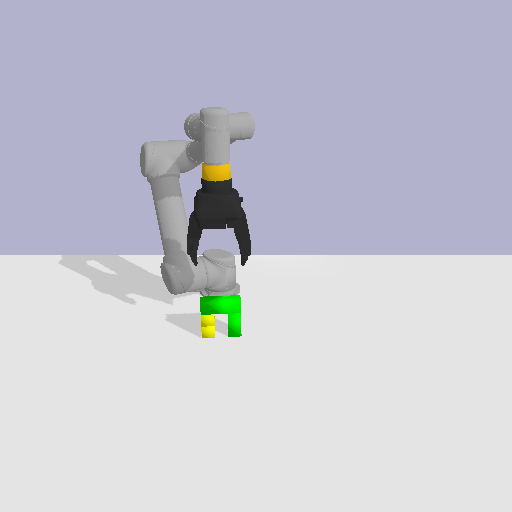}\hspace{.1cm}
  % 4u sim
  \includegraphics[trim=110 50 180 140, clip, width=.1185\linewidth]{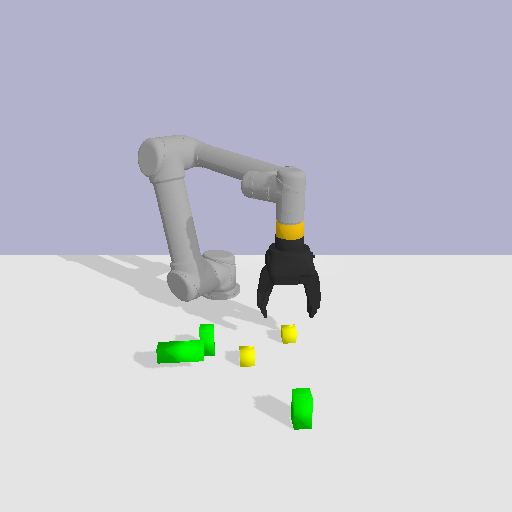}\hspace{-.06cm}
  \includegraphics[trim=110 50 180 140, clip, width=.1185\linewidth]{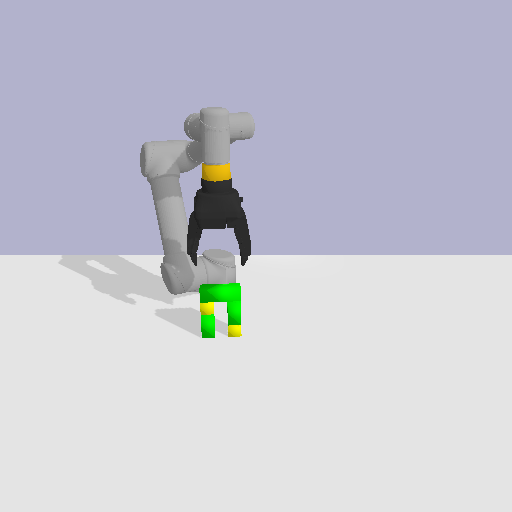}\hspace{.1cm}
  % 5u sim
  \includegraphics[trim=110 50 180 140, clip, width=.1185\linewidth]{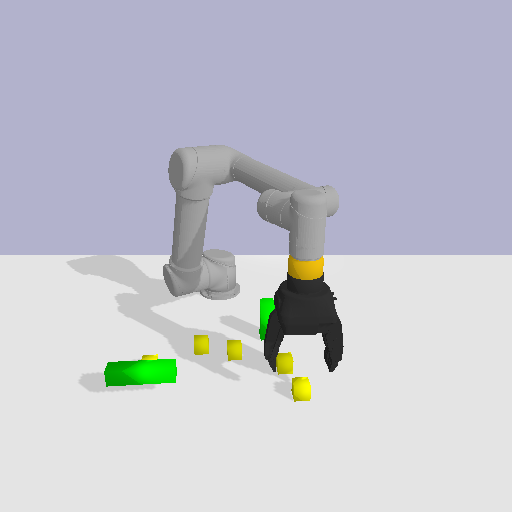}\hspace{-.06cm}
  \includegraphics[trim=110 50 180 140, clip, width=.1185\linewidth]{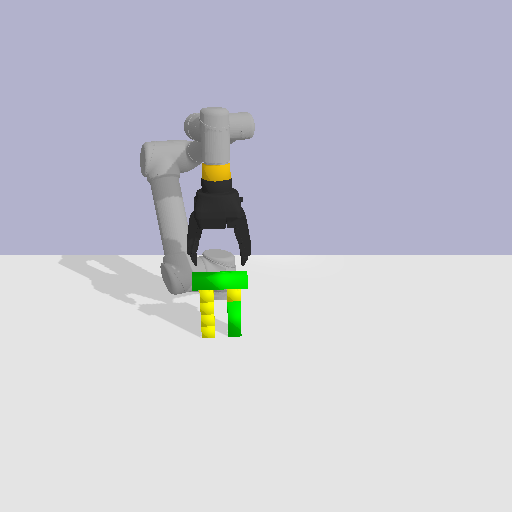}\hspace{.1cm}
  % unseen primitives
  \includegraphics[trim=220 0 291 25, clip, width=.1185\linewidth]{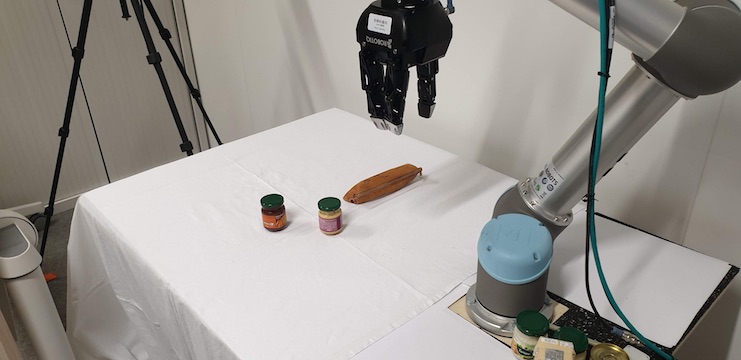}\hspace{-.06cm}
  \includegraphics[trim=261 0 250 25, clip, width=.1185\linewidth]{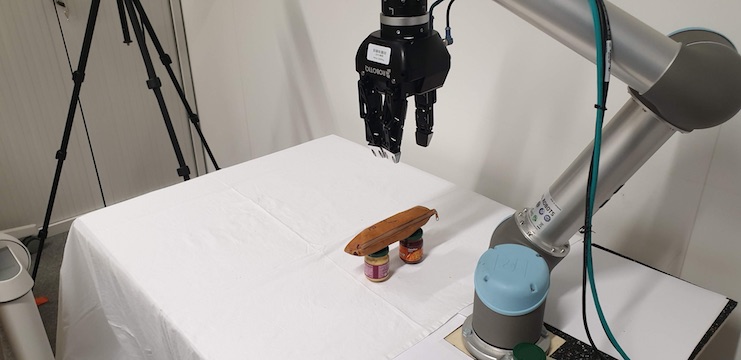}
  \\
  % 3u real
  \includegraphics[trim=220 0 190 20, clip, width=.1185\linewidth]{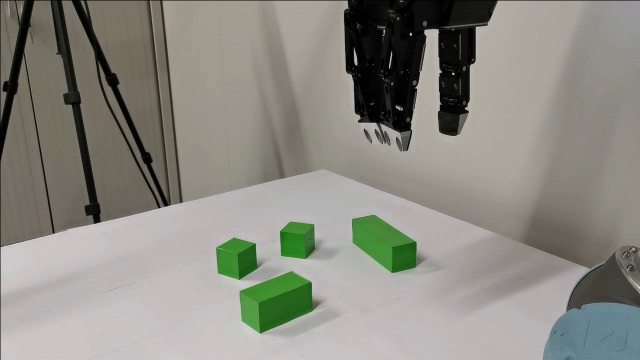}\hspace{-.06cm}
  \includegraphics[trim=260 0 150 20, clip, width=.1185\linewidth]{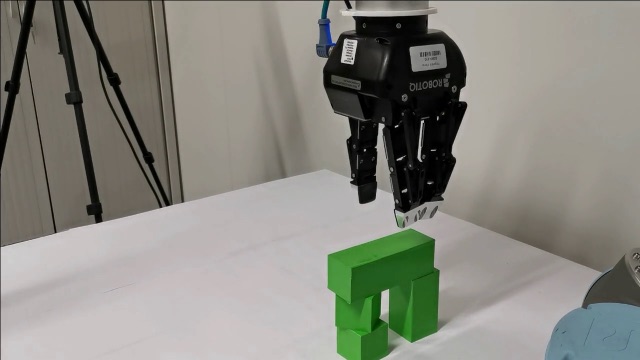}\hspace{.1cm}
  % 4u real
  \includegraphics[trim=205 0 205 20, clip, width=.1185\linewidth]{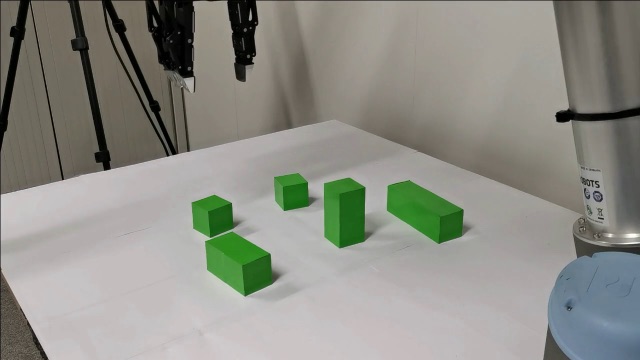}\hspace{-.06cm}
  \includegraphics[trim=280 0 130 20, clip, width=.1185\linewidth]{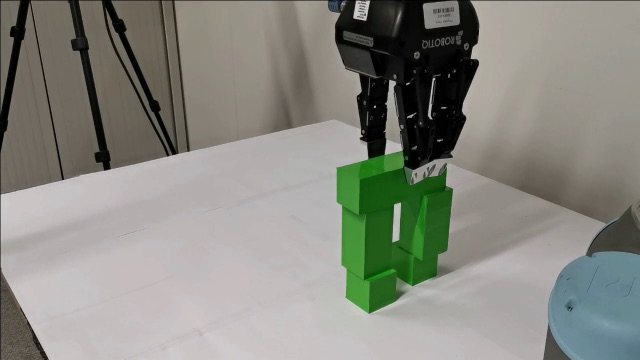}\hspace{.1cm}
  % 5u real
  \includegraphics[trim=170 0 240 20, clip, width=.1185\linewidth]{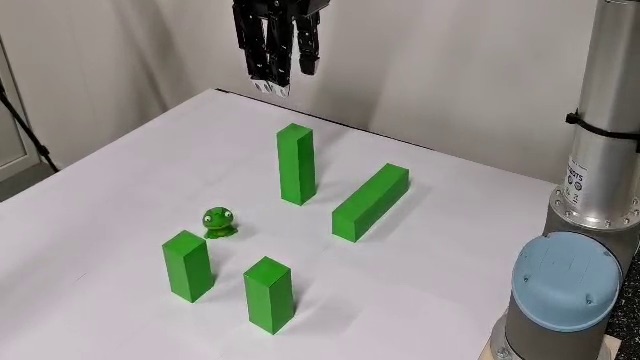}\hspace{-.06cm}
  \includegraphics[trim=210 0 200 20, clip, width=.1185\linewidth]{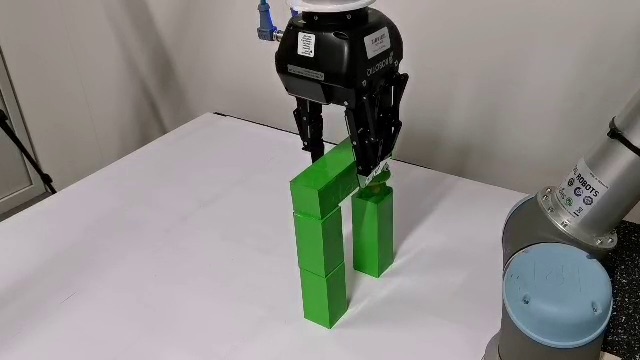}\hspace{.1cm}
  % unseen primitives
  \includegraphics[trim=225 0 286 20, clip, width=.1185\linewidth]{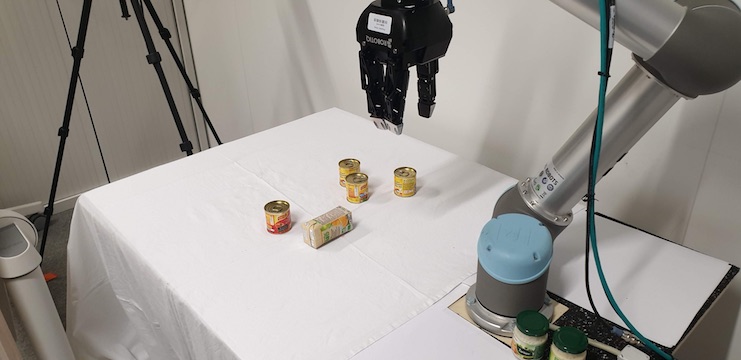}\hspace{-.06cm}
  \includegraphics[trim=261 0 250 20, clip, width=.1185\linewidth]{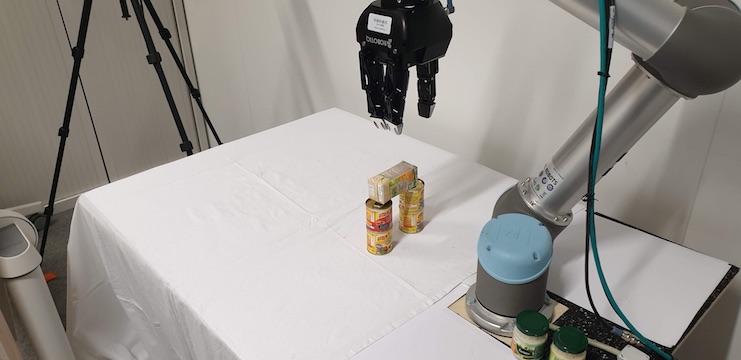}
  \caption{\small Assembling arches of 3-5 units in simulation (first 6 columns, top) and with a real robot (first 6 columns, bottom). Assembling arches with never-seen primitives resembling cubes and beams on the real robot (last 2 columns). The image pairs correspond to initial and final states.}
  \label{fig:arch}
  \vspace{-.9cm}
\end{figure*}

\textit{Arch.}
The agent needs to use all primitives available on a table to build an arch (see Fig.~\ref{fig:arch}).
The construction primitives are cubes of size 1U and beams of length 2U and 3U.
The arch shape category is defined as two symmetrical pillars with a bar bridging them.
For example, pillars of an arch could be constructed from a 3U beam,
three cubes or one cube and a 2U beam. Initially, all primitives are randomly
distributed on the surface of the table. The beams can have three axes-parallel
orientations. The primitives can have any color that differ from the color of the 
table. The location of pillars on the table is pre-defined. There are 49, 16 and 4 instances for 5U, 4U and 3U arch shapes correspondingly. We use the Arch task
to evaluate the generalization of our method to the shape category and to show
advantages of unmaking procedure in Sections~\ref{sec:results:vf} and \ref{sec:results:vp}.

\vspace{-.2cm}
\subsection{Learning in state space}
\label{sec:results:vf}
\vspace{-.1cm}

\begin{figure}
  \vspace{.3cm}
  \centering
    %[trim=left bottom right top, clip] 
  \includegraphics[trim=155 45 255 35, clip, width=.235\linewidth]{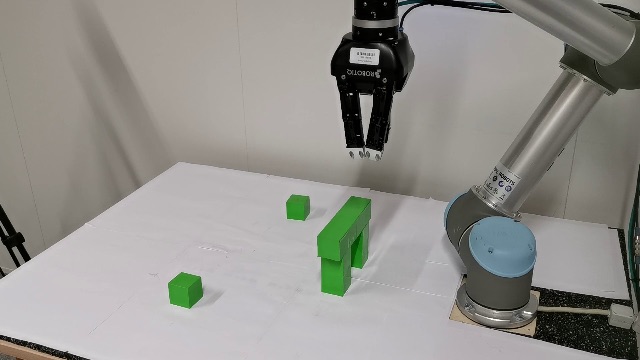}
  \includegraphics[trim=160 45 250 35, clip, width=.235\linewidth]{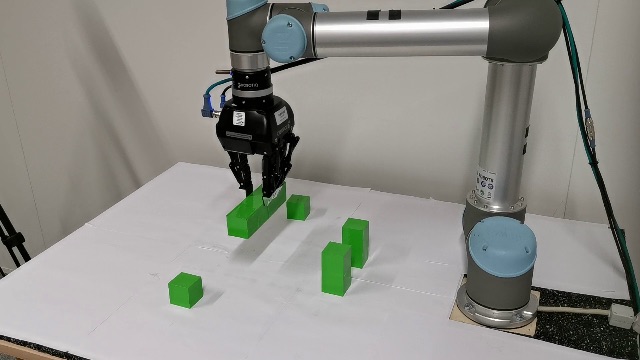}
  \includegraphics[trim=160 45 250 35, clip, width=.235\linewidth]{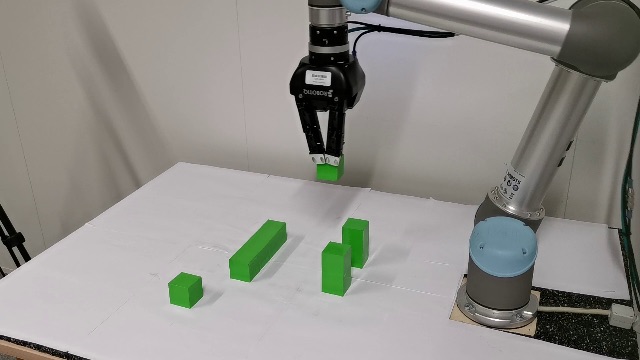}
  \includegraphics[trim=220 45 190 35, clip, width=.235\linewidth]{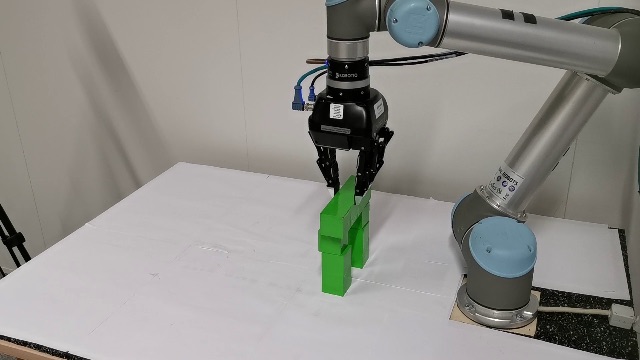}
  \caption{\small Our visual policies successfully re-build shapes using primitives added to the table.}
  \label{fig:reassemble}
  \vspace{-.7cm}
\end{figure}

% intro
This section evaluates how efficient our approach in generalizing to 3D shapes in the state space.
For evaluation, we use a simulated Arch task.
% describe our method
Our approach receives a single shape instance as an input and learns a state-value function by
unmaking this instance. Given the trained value function and the simulator, we obtain the state
policy by iterating over all possible actions and taking the one that maximizes
the value function prediction.
This state policy is then used to discover new object instances.
We iteratively repeat the state-value network training after unmaking the discovered instances during 20 iterations.
% describe the baselines
We compare our approach to MCTS~\cite{Coulom2006EfficientSA} and random exploration.
Similarly to the state policy, both baselines choose an object for picking and its placing location on top of other objects.
The baselines also choose the placing orientation among the three axis-parallel directions.
For MCTS, we use a shape matching score which is defined by a percentage of how much the arch shape is completed with primitives.

% efficiency
We estimate an average amount of steps required by our method and the
baselines to build an arch, and report results in Tab.~\ref{table:arches:vp}. 
We also report the minimum number of steps required to build an arch (Oracle).
Note that the Oracle has a non-zero variance since arches can be composed from different numbers of primitives.
While the efficiency of our approach is comparable to Oracle, baselines require orders of magnitude more steps to find a correct solution. 
For example, to build a 5U arch, MCTS and the random exploration require 1.6e4 and 6.1e4 steps respectively while our method solves the task in 6.9 steps on average.
We expect our method to scale well to more complex tasks where the complexity of baselines will prohibit their use.

\vspace{-.2cm}
\subsection{Evaluation of visual policies}
\label{sec:results:vp}
\vspace{-.1cm}
This section evaluates visual policies trained to solve tasks given images as input.
We validate the HourGlass architecture in
Section~\ref{sec:results:vp:archi}, show benefits of the proposed unmaking approach over
prior work in Section~\ref{sec:results:vp:unmake}, evaluate generalization of
our method to the shape category in Section~\ref{sec:results:vp:category} and
present an evaluation on the real robot in Section~\ref{sec:results:vp:real}.

\subsubsection{Visual policies architecture}
\label{sec:results:vp:archi}
% intro
We compare our heatmap-based architecture to a ResNet-18 network that 
directly outputs source and target parameters of manipulated primitives.
% describe our method
Our approach uses HourGlass~\cite{newell2016stacked} to predict a multi-modal distribution of picking and placing actions. The first heatmap corresponds to 2D locations on the robot workspace, the three additional heatmaps encode
orientations of a primitive.
% describe the alternative
ResNet-18~\cite{He2016DeepRL} predicts five values corresponding to 2D locations and three orientations of objects.

% discuss the results
We train HourGlass and ResNet-18 networks to build towers of 3,
5 and 7 cubes of green, yellow and red colors. For ResNet-18, we adapt the learning rate to 1e-3 and the input image size to 224.
Tab.~\ref{table:towers} indicates that both HourGlass and ResNet policies achieve
almost perfect accuracy for towers of 3 cubes (Fig.~\ref{fig:tower}, left).
Given more complex scenes with multiple primitives of the same color, ResNet fails due to its unimodal prediction. As illustrated in Fig.~\ref{fig:tower}, it outputs a mean location of relevant primitives.
The HourGlass-based policy builds towers of 7 cubes with a failure rate of 6\%
where the errors are mainly caused by occlusions.
On the real world Tower task with 7 cubes, the HourGlass-based policy succeeds in 18 trials out of 20.
The two failure cases were caused by an occlusion and
misidentifying a cube due to the sim2real gap. Empirically, we did not find any performance improvement when using multiple HourGlass modules.

\subsubsection{Learning by unmaking}
\label{sec:results:vp:unmake}
This section compares the proposed approach of unmaking assembled objects 
with prior
work~\cite{disassembly2019}. While our method disassembles objects in multiple ways,
\cite{disassembly2019} proposes to use a single disassembly trajectory.
Such disassembly trajectories consist of pairs of state and corresponding action. However, there could be multiple correct actions possible in a state as shown in
Fig.~\ref{fig:heatmap_explain}.
We address this by computing several disassembly paths and merging actions that correspond
to states, where the only difference
comes from positions of primitives located on the table surface.

We train visual policies on observation-heatmap pairs obtained with and without multiple
unmake trajectories.
Tab.~\ref{table:arches:merge} shows that using multiple unmake paths
significantly improves the performance. The performance difference  is 26\% on the hardest task of building 5U arches. Using multiple unmake paths make the policy learn multiple hypothesis of picking and placing locations. This property becomes critical
when multiple identical objects are used or if there exist several shape instances where identical primitives are assembled
in different configurations.

\subsubsection{Learning to build a 3D category shapes}
\label{sec:results:vp:category}
This section evaluates the generalization of our approach to a shape category.
We train visual policies on trajectories of unmaking 3 sets of arch instances: (i) only the input
instance, (ii)~instances obtained by a state policy trained on arches of the
same height, (iii) instances obtained by a state policy trained on arches of
heights 3-5U.
We refer to these policies as (i) instance, (ii) uni-height and (iii) multi-height.
For instance and uni-height policies, we train separate networks to build arches of each height.
For multi-height policy, the same network is used to build arches of varying heights.

We evaluate all the policies separately on the set of primitives corresponding to the
input instance (Tab.~\ref{table:arches:sim}, left) and on different
sets corresponding to the entire category (Tab.~\ref{table:arches:sim}, right).
In the first case, all the policies assemble arches with less than 5\% of failures.
The results of uni--height policies are higher compared to the instance policies.
We believe that this improvement is due to a higher variation of the category instances that can be seen as a form of data augmentation.
However, the performance of the instance policies rapidly drops when they are exposed to
unseen sets of primitives. The uni-height policies have 2-3\% higher success rates than the multi-height policy. Their success rates on 5U arches are 95.6\% and 94.0\% correspondingly. However, we need to train only a single network for the multi-height policy
which is then able to reason about available primitives on the table and decide which arch height to build.

\subsubsection{Real robot evaluation}
\label{sec:results:vp:real}
This section evaluates the performance of our method on the real world Arch task using two scenarios: (i) "normal" scenario that matches the simulation, (ii) "re-assembling" scenario.
In the "normal" scenario the robot has to assemble arches from varying sets of primitives.
Tab.~\ref{table:arches:real}~(left) shows that uni-height policies have similar performance to the multi-height policy in the standard scenario except for 3U arches.
In the "re-assembling" scenario the task starts with an assembled arch and additional primitives on the table (see Fig.~\ref{fig:reassemble}). The agent is expected to re-assemble the arch by using all available primitives on the table.
Our method is able to automatically re-assemble shapes without additional training due to the presence of random pick-place actions in the train set.
Similarly to the value function dataset described in Section~\ref{sec:statespace}, we record observation-actions pairs corresponding to the inverse of the one-step random actions that include examples of re-assembling.
In the "re-assembling" scenario (Tab.~\ref{table:arches:real}~(right)) the multi-height policy has higher success rates for 4U and 5U arches. We illustrate the assembly of arches with the real UR5 robot arm in Figures~\ref{fig:arch}, \ref{fig:reassemble} and on the project web-page~\cite{pashevich20buildingWeb}.
Failure cases with incorrect assembly are typically caused by occlusions and the gap between simulated and real environments.

Finally, we test the generalization of our approach to new building primitives and compare it to the MCTS baseline. We use three different sets of primitives that resemble cuboids used during training. The sets contain (i) 2 jars and a pencil case (Fig.~\ref{fig:arch} last column, top), (ii) 4 cans and a juice box (Fig.~\ref{fig:arch} last column, bottom), (iii) 5 stones (Fig.~\ref{fig:teaser} bottom).
The input we provide to the MCTS baseline is the state of primitives in terms of their sizes and locations. 
Size and location are determined by clustering 3D points, which are obtained based on depth image coordinates above the table similarly to the prediction correction procedure used for our method (see Appendix of~\cite{3Dshapes2020}).
We estimate the spatial dimensions of primitives by fitting bounding boxes to depth points associated to each 3D cluster.
Tab.~\ref{table:arches:unseen} shows that our method significantly outperforms the MCTS baseline with the performance gap of up to 45\%.
In all cases the failures of MCTS are caused by errors in the estimation of size and location of the primitives. Given the incorrect estimates, MCTS cannot find an assembly path to build a correct arch. In contrast, our method relies on position correction based on direct image input and does not require spatial estimation of the primitives. 
Both methods often fail on the third set of primitives with stones given the substantial difference of stones to cuboid primitives used during training. Qualitative results for additional sets of primitives are presented in Appendix of~\cite{3Dshapes2020}.

\vspace{-.1cm}
\section{Conclusion}
\vspace{-.1cm}
We proposed an approach to build 3D object shapes using a robotic arm and varying sets of primitives. Our method efficiently learns to solve the task in the state space and then uses the obtained solutions as supervision to train visual policies in the observation space. We demonstrate successful application of our method to the new task of assembling shape categories and show promising results on a real robot.

While the disassembling procedure explored in this work may not be directly applicable to all tasks, we note that it could generalize even to physically irreversible actions by learning appropriate backward models in state space. Future work will explore this direction for a wide range of tasks including cooking and other more complex assembling tasks.
We also note that the complexity of the proposed unmake procedure grows exponentially with the number of primitives. To address more complex tasks, it will be interesting to extend our method to a fixed number of sampled disassembly trajectories.

\noindent
\textbf{Acknowledgements.} This work was funded in part by the French government under management of Agence Nationale de la Recherche as part of the ”Investissements d’avenir” program, reference ANR19-P3IA-0001 (PRAIRIE 3IA Institute).

\vspace{-0.4cm}
\bibliographystyle{myplain}
\bibliography{ref}

\begin{thebibliography}{10}

\bibitem{pashevich20buildingWeb}
Project webpage.
\newblock \url{http://pascal.inrialpes.fr/data2/3D-shapes/}.

\bibitem{agrawal2016learning}
P. Agrawal, A.~V. Nair, P. Abbeel, J. Malik, and S. Levine.
\newblock Learning to poke by poking: Experiential learning of intuitive
  physics.
\newblock In {\em NIPS}, 2016.

\bibitem{Andrychowicz2017HindsightER}
M. Andrychowicz, D. Crow, A. Ray, J. Schneider, R.~H. Fong, P. Welinder, B.
  McGrew, J. Tobin, P. Abbeel, and W. Zaremba.
\newblock Hindsight experience replay.
\newblock {\em NIPS}, 2017.

\bibitem{robot-learning-survey}
B.~D. Argall, S. Chernova, M. Veloso, and B. Browning.
\newblock A survey of robot learning from demonstration.
\newblock {\em RAS}, 57(5), May 2009.

\bibitem{Chen_2019_ICCV}
Z. Chen, D. Guo, T. Xiao, S. Xie, X. Chen, H. Yu, J. Gray, K. Srinet, H. Fan,
  J. Ma, C. Qi, S. Tulsiani, A. Szlam, and L. Zitnick.
\newblock {Order-Aware Generative Modeling Using the 3D-Craft Dataset}.
\newblock In {\em ICCV}, 2019.

\bibitem{codevilla2018end}
F. Codevilla, M. Miiller, A. L{\'o}pez, V. Koltun, and A. Dosovitskiy.
\newblock End-to-end driving via conditional imitation learning.
\newblock {\em ICRA}, 2018.

\bibitem{Coulom2006EfficientSA}
R. Coulom.
\newblock Efficient selectivity and backup operators in monte-carlo tree
  search.
\newblock In {\em Computers and Games}, 2006.

\bibitem{Courmans2016}
E. Courmans and Y. Bai.
\newblock {PyBullet}, {P}ython module for physics simulation, robotics and
  machine learning, 2016.

\bibitem{Dantam2018AnIC}
N.~T. Dantam, Z.~K. Kingston, S. Chaudhuri, and L.~E. Kavraki.
\newblock An incremental constraint-based framework for task and motion
  planning.
\newblock {\em IJRR}, 37, 2018.

\bibitem{onshot-imitation}
Y. Duan, M. Andrychowicz, B. Stadie, J. Ho, J. Schneider, I. Sutskever, P.
  Abbeel, and W. Zaremba.
\newblock One-shot imitation learning.
\newblock {\em NIPS}, 2017.

\bibitem{Ebert2018VisualFM}
F. Ebert, C. Finn, S. Dasari, A. Xie, A.~X. Lee, and S. Levine.
\newblock {Visual Foresight: Model-Based Deep Reinforcement Learning for
  Vision-Based Robotic Control}.
\newblock {\em arXiv}, 2018.

\bibitem{Edelkamp2004PDDL2}
S. Edelkamp and J. Hoffmann.
\newblock {PDDL2.2: The Language for the Classical Part of IPC-4}.
\newblock 2004.

\bibitem{servoing}
B. {Espiau}, F. {Chaumette}, and P. {Rives}.
\newblock A new approach to visual servoing in robotics.
\newblock {\em IEEE Transactions on Robotics and Automation}, 1992.

\bibitem{FIKES1971189}
R. Fikes and N. Nilsson.
\newblock Strips: A new approach to the application of theorem proving to
  problem solving.
\newblock {\em Artificial Intelligence}, 1971.

\bibitem{florensa2017reverse}
C. Florensa, D. Held, M. Wulfmeier, M. Zhang, and P. Abbeel.
\newblock Reverse curriculum generation for reinforcement learning.
\newblock In {\em CoRL}, 2017.

\bibitem{gandhi2017learning}
D. Gandhi, L. Pinto, and A. Gupta.
\newblock Learning to fly by crashing.
\newblock In {\em IROS}, 2017.

\bibitem{garrett2015ffrob}
C.~R. Garrett, T. Lozano-P{\'e}rez, and L.~P. Kaelbling.
\newblock Ffrob: An efficient heuristic for task and motion planning.
\newblock In {\em Algorithmic Foundations of Robotics XI}, pages 179--195.
  Springer, 2015.

\bibitem{Garrett2018FFRobLS}
C.~R. Garrett, T. Lozano-P{\'e}rez, and L.~P. Kaelbling.
\newblock {FFRob: Leveraging symbolic planning for efficient task and motion
  planning}.
\newblock {\em IJRR}, 37:104--136, 2018.

\bibitem{garrett2018sampling}
C.~R. Garrett, T. Lozano-P{\'e}rez, and L.~P. Kaelbling.
\newblock Sampling-based methods for factored task and motion planning.
\newblock {\em IJRR}, 37, 2018.

\bibitem{lepetit2018}
A. {Grabner}, P.~M. {Roth}, and V. {Lepetit}.
\newblock {3D Pose Estimation and 3D Model Retrieval for Objects in the Wild}.
\newblock In {\em CVPR}, 2018.

\bibitem{Gu2016}
S. Gu, E. Holly, T. Lillicrap, and S. Levine.
\newblock Deep reinforcement learning for robotic manipulation.
\newblock In {\em ICML}, 2016.

\bibitem{He2015towards}
K. {He}, M. {Lahijanian}, L. {Kavraki}, and M. {Vardi}.
\newblock Towards manipulation planning with temporal logic specifications.
\newblock In {\em ICRA}, 2015.

\bibitem{He2016DeepRL}
K. He, X. Zhang, S. Ren, and J. Sun.
\newblock Deep residual learning for image recognition.
\newblock In {\em CVPR}, 2016.

\bibitem{hosu2016playing}
I.-A. Hosu and T. Rebedea.
\newblock Playing atari games with deep reinforcement learning and human
  checkpoint replay.
\newblock In {\em ECAI Workshop on Evaluating General Purpose AI}, 2016.

\bibitem{Huang2018NeuralTG}
D.-A. Huang, S. Nair, D. Xu, Y. Zhu, A. Garg, L. Fei-Fei, S. Savarese, and
  J.~C. Niebles.
\newblock Neural task graphs: Generalizing to unseen tasks from a single video
  demonstration.
\newblock In {\em CVPR}, 2018.

\bibitem{Ioffe2015BatchShift}
S. Ioffe and C. Szegedy.
\newblock Batch normalization: Accelerating deep network training by reducing
  internal covariate shift.
\newblock {\em ICML}, 2015.

\bibitem{rlblogpost}
A. Irpan.
\newblock Deep reinforcement learning doesn't work yet, 2018.

\bibitem{Jaderberg2016ReinforcementLW}
M. Jaderberg, V. Mnih, W. Czarnecki, T. Schaul, J.~Z. Leibo, D. Silver, and K.
  Kavukcuoglu.
\newblock Reinforcement learning with unsupervised auxiliary tasks.
\newblock {\em arXiv}, 2016.

\bibitem{janner2019reasoning}
M. Janner, S. Levine, W.~T. Freeman, J.~B. Tenenbaum, C. Finn, and J. Wu.
\newblock Reasoning about physical interactions with object-oriented prediction
  and planning.
\newblock In {\em ICLR}, 2019.

\bibitem{kobber-survey}
J. Kober, J. Bagnell, and J. Peters.
\newblock Reinforcement learning in robotics: A survey.
\newblock {\em IJRR}, 2013.

\bibitem{Lampe2013AcquiringLearning}
T. Lampe and M. Riedmiller.
\newblock Acquiring visual servoing reaching and grasping skills using neural
  reinforcement learning.
\newblock {\em IJCNN}, 2013.

\bibitem{lavalle2006-palnning-algorithms}
S. LaValle.
\newblock {\em Planning Algorithms}.
\newblock Cambridge University, 2006.

\bibitem{Levine2015End-to-EndPolicies}
S. Levine, C. Finn, T. Darrell, and P. Abbeel.
\newblock End-to-end training of deep visuomotor policies.
\newblock {\em JMLR}, 2015.

\bibitem{long2015fully}
J. Long, E. Shelhamer, and T. Darrell.
\newblock Fully convolutional networks for semantic segmentation.
\newblock In {\em CVPR}, 2015.

\bibitem{lozano2014constraint}
T. Lozano-P{\'e}rez and L.~P. Kaelbling.
\newblock A constraint-based method for solving sequential manipulation
  planning problems.
\newblock In {\em IROS}, 2014.

\bibitem{Mahler2017DexNet2D}
J. Mahler, J. Liang, S. Niyaz, M. Laskey, R. Doan, X. Liu, J.~A. Ojea, and
  K.~Y. Goldberg.
\newblock {Dex-Net 2.0: Deep Learning to Plan Robust Grasps with Synthetic
  Point Clouds and Analytic Grasp Metrics}.
\newblock {\em RSS}, 2017.

\bibitem{nair2018overcoming}
A. Nair, B. McGrew, M. Andrychowicz, W. Zaremba, and P. Abbeel.
\newblock Overcoming exploration in reinforcement learning with demonstrations.
\newblock In {\em ICRA}, 2018.

\bibitem{Nair2018TimeRA}
S. Nair, M. Babaeizadeh, C. Finn, S. Levine, and V. Kumar.
\newblock Time reversal as self-supervision.
\newblock {\em ICRA}, 2020.

\bibitem{newell2016stacked}
A. Newell, K. Yang, and J. Deng.
\newblock Stacked hourglass networks for human pose estimation.
\newblock In {\em ECCV}, 2016.

\bibitem{ModulatedPH}
A. Pashevich, D. Hafner, J. Davidson, R. Sukthankar, and C. Schmid.
\newblock Modulated policy hierarchies.
\newblock {\em NIPS Deep RL workshop}, 2018.

\bibitem{3Dshapes2020}
A. Pashevich, I. Kalevatykh, I. Laptev, and C. Schmid.
\newblock Learning visual policies for building 3d shape categories.
\newblock {\em arXiv}, 2020.

\bibitem{learningsim2real2019}
A. Pashevich, R. Strudel, I. Kalevatykh, I. Laptev, and C. Schmid.
\newblock Learning to augment synthetic images for sim2real policy transfer.
\newblock In {\em IROS}, 2019.

\bibitem{paxton2019visual}
C. {Paxton}, Y. {Barnoy}, K. {Katyal}, R. {Arora}, and G.~D. {Hager}.
\newblock Visual robot task planning.
\newblock In {\em ICRA}, 2019.

\bibitem{pinto-grasp}
L. Pinto and A. Gupta.
\newblock Supersizing self-supervision: Learning to grasp from 50k tries and
  700 robot hours.
\newblock ICRA 2016.

\bibitem{Pomerleau1989}
D.~A. Pomerleau.
\newblock {Alvinn: {A}n autonomous land vehicle in a neural network}.
\newblock In {\em NIPS}, 1989.

\bibitem{popov-manipulation}
I. Popov, N. Heess, T. Lillicrap, R. Hafner, G. Barth{-}Maron, M.
  Vecer{\'{\i}}k, T. Lampe, Y. Tassa, T. Erez, and M. Riedmiller.
\newblock Data-efficient deep reinforcement learning for dexterous
  manipulation.
\newblock {\em arXiv}, 2017.

\bibitem{Rahmatizadeh2017VisionBasedMM}
R. Rahmatizadeh, P. Abolghasemi, L. B{\"o}l{\"o}ni, and S. Levine.
\newblock Vision-based multi-task manipulation for inexpensive robots using
  end-to-end learning from demonstration.
\newblock {\em ICRA}, 2018.

\bibitem{fasterrcnn}
S. Ren, K. He, R. Girshick, and J. Sun.
\newblock {Faster R-CNN: Towards Real-time Object Detection with Region
  Proposal Networks}.
\newblock In {\em NIPS}, 2015.

\bibitem{SACX}
M.~A. Riedmiller, R. Hafner, T. Lampe, M. Neunert, J. Degrave, T.~V. de~Wiele,
  V. Mnih, N. Heess, and J.~T. Springenberg.
\newblock Learning by playing - {S}olving sparse reward tasks from scratch.
\newblock {\em MLR}, 2018.

\bibitem{cad2real}
F. Sadeghi and S. Levine.
\newblock {CAD2RL: Real Single-Image Flight Without a Single Real Image}.
\newblock 2017.

\bibitem{srivastava2014combined}
S. Srivastava, E. Fang, L. Riano, R. Chitnis, S. Russell, and P. Abbeel.
\newblock Combined task and motion planning through an extensible
  planner-independent interface layer.
\newblock In {\em ICRA}, 2014.

\bibitem{suarez2018can}
F. Su{\'a}rez-Ruiz, X. Zhou, and Q.-C. Pham.
\newblock Can robots assemble an ikea chair?
\newblock {\em Science Robotics}, 3(17), 2018.

\bibitem{sucan2012the-open-motion-planning-library}
I.~A. {\c{S}}ucan, M. Moll, and L.~E. Kavraki.
\newblock The {O}pen {M}otion {P}lanning {L}ibrary.
\newblock {\em Robotics \& Automation Magazine}, 19(4):72--82, 2012.

\bibitem{Sukhbaatar2017IntrinsicMA}
S. Sukhbaatar, I. Kostrikov, A. Szlam, and R. Fergus.
\newblock Intrinsic motivation and automatic curricula via asymmetric
  self-play.
\newblock {\em ICLR}, 2017.

\bibitem{sim2real2017tobin}
J. Tobin, R. Fong, A. Ray, J. Schneider, W. Zaremba, and P. Abbeel.
\newblock Domain randomization for transferring deep neural networks from
  simulation to the real world.
\newblock {\em IROS}, 2017.

\bibitem{toussaint2015logic}
M. Toussaint.
\newblock Logic-geometric programming: An optimization-based approach to
  combined task and motion planning.
\newblock {\em IJACI}, 2015.

\bibitem{Wang2019LearningRM}
A. Wang, T. Kurutach, K. Liu, P. Abbeel, and A. Tamar.
\newblock Learning robotic manipulation through visual planning and acting.
\newblock {\em arXiv}, 2019.

\bibitem{Wei_2016_CVPR}
S.-E. Wei, V. Ramakrishna, T. Kanade, and Y. Sheikh.
\newblock Convolutional pose machines.
\newblock In {\em CVPR}, 2016.

\bibitem{disassembly2019}
K. Zakka, A. Zeng, J. Lee, and S. Song.
\newblock {Form2Fit: Learning Shape Priors for Generalizable Assembly from
  Disassembly}.
\newblock {\em arXiv}, 2019.

\bibitem{Zeng2019TossingBotLT}
A. Zeng, S. Song, J. Lee, A. Rodr{\'i}guez, and T.~A. Funkhouser.
\newblock Tossingbot: Learning to throw arbitrary objects with residual
  physics.
\newblock {\em RSS}, 2019.

\bibitem{Zhang2017DeepIL}
T. Zhang, Z. McCarthy, O. Jow, D. Lee, K. Goldberg, and P. Abbeel.
\newblock Deep imitation learning for complex manipulation tasks from virtual
  reality teleoperation.
\newblock {\em ICRA}, 2017.

\end{thebibliography}

\clearpage
 \section{Appendix}
\label{sec:appendix}

\subsection{Real robot implementation details}
\label{sec:appendix:real_robot}

\begin{figure}[h!]
  \centering
  %[trim=left bottom right top, clip]
  \includegraphics[trim=0 130 310 0, clip, width=\linewidth]{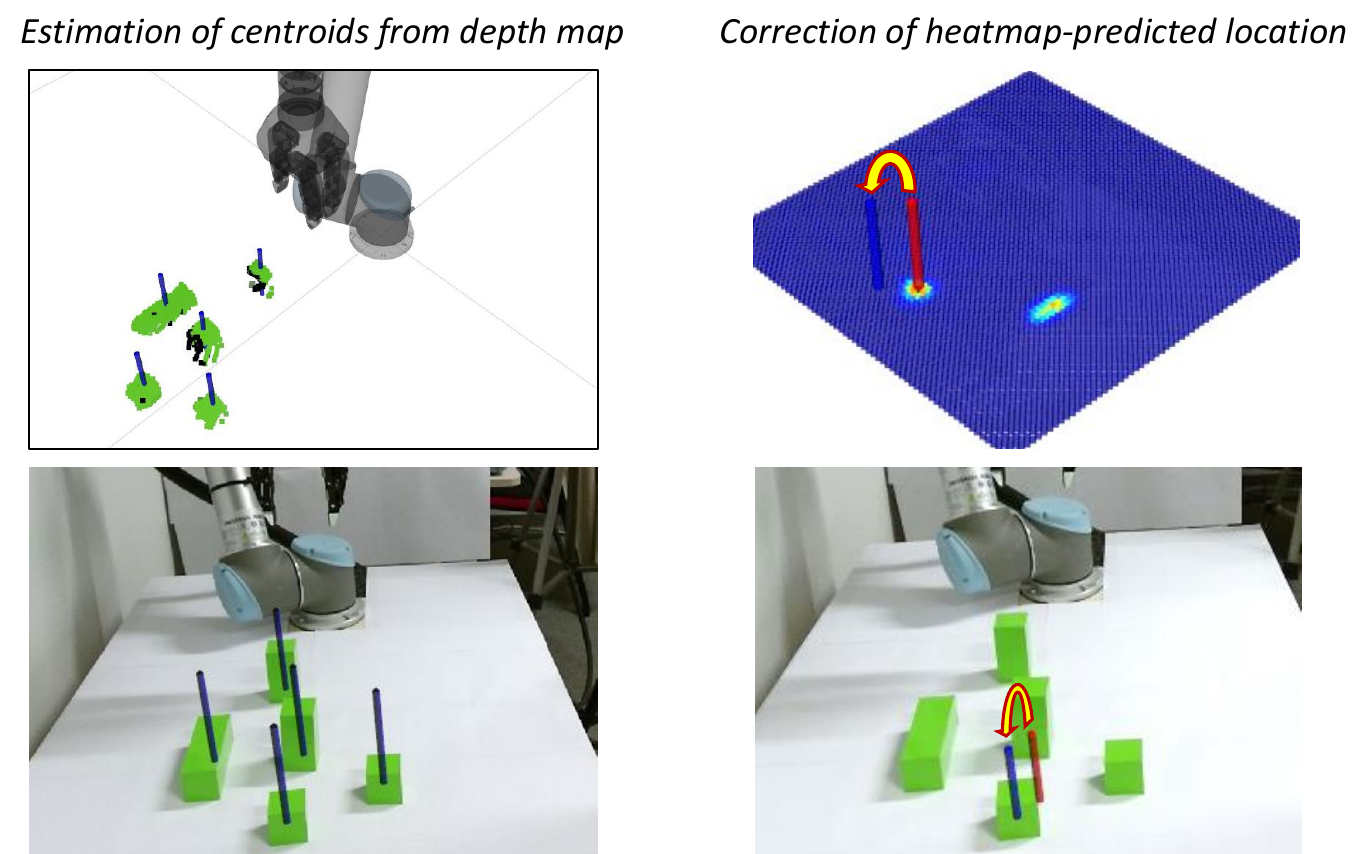}
  \caption{Correction procedure applied to the spatial maxima of predicted heatmaps. Left: object centroids (blue bars) are estimated by the spatial clustering of depth-map locations above the table surface. Right: The location of the heatmap maxima (red bar) is corrected to the location of the closest object centroid. Refer to Section~\ref{sec:appendix:real_robot} for further explanations.}
  \label{fig:correction}
\end{figure}

The goal of our method is to build a 3D object by selecting the right primitives and performing correct construction actions. While the learned policies typically plan correct actions, object locations predicted by the heatmaps in the real robot setup may miss the primitive by a few centimeters. Such errors could be addressed, for example, by learning additional policies for location prediction. Here we chose a simpler solution and make the correction of predicted locations using depth images. As illustrated in Figure~\ref{fig:correction}, given a depth map, we first obtain object centroids by clustering points above the table surface. Next, we predict the spatial location for the next action by maximizing the heatmap produced by the policy. The predicted location is then corrected to the location of the nearest cluster centroid. The applied corrections are typically in the order of 2-3 centimeters.

\subsection{Qualitative results}
\label{sec:appendix:qualitative}

\begin{figure*}
  \centering
  %[trim=left bottom right top, clip]
  \includegraphics[trim=100 0 140 0, clip, width=.16\linewidth]{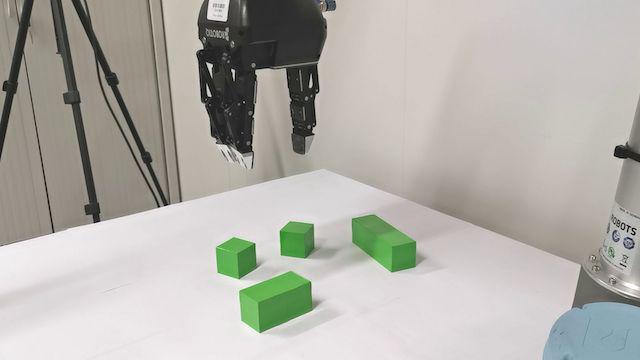}
  \includegraphics[trim=100 0 140 0, clip, width=.16\linewidth]{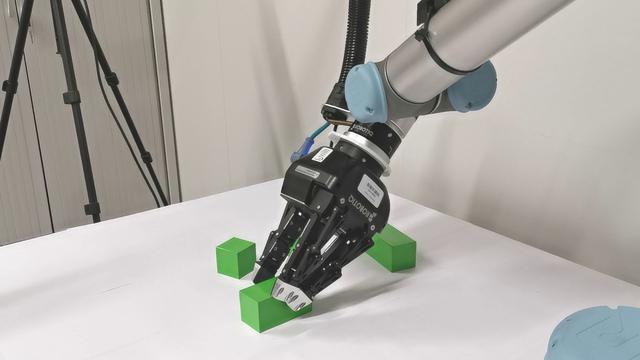}
  \includegraphics[trim=100 0 140 0, clip, width=.16\linewidth]{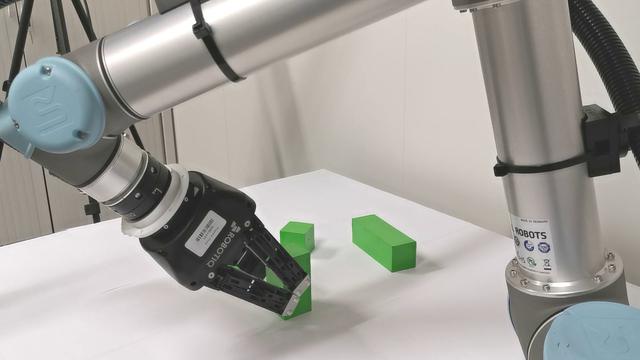}
  \includegraphics[trim=100 0 140 0, clip, width=.16\linewidth]{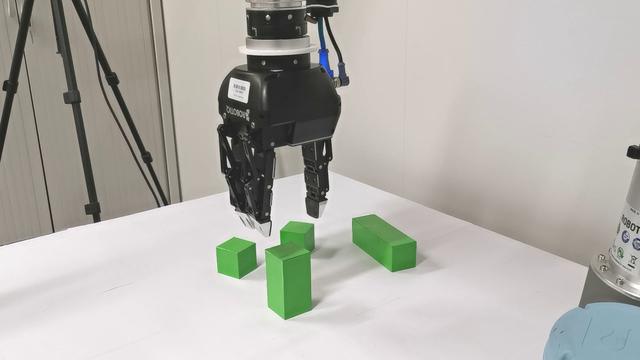}
  \includegraphics[trim=100 0 140 0, clip, width=.16\linewidth]{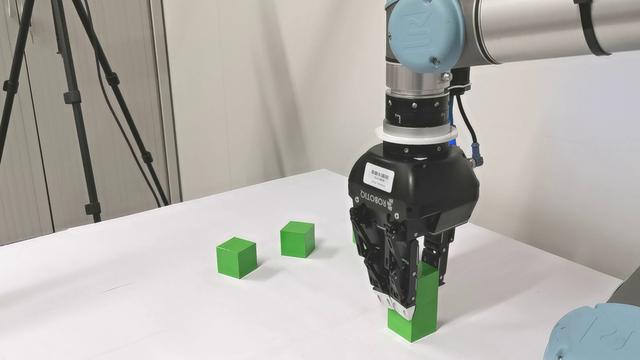}
  \includegraphics[trim=100 0 140 0, clip, width=.16\linewidth]{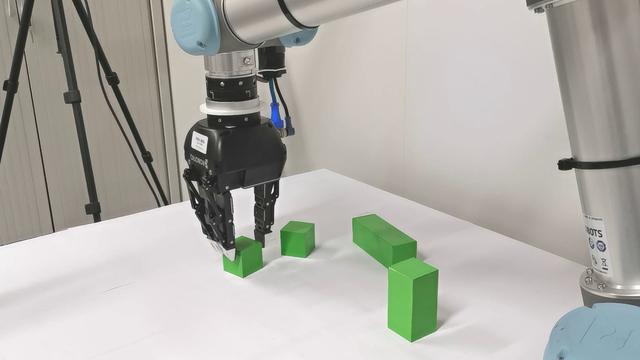}
  \includegraphics[trim=100 0 140 0, clip, width=.16\linewidth]{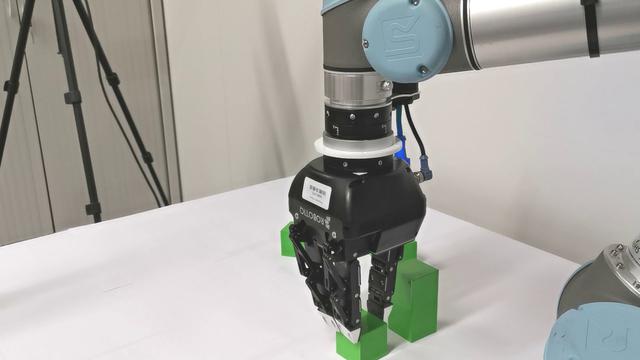}
  \includegraphics[trim=100 0 140 0, clip, width=.16\linewidth]{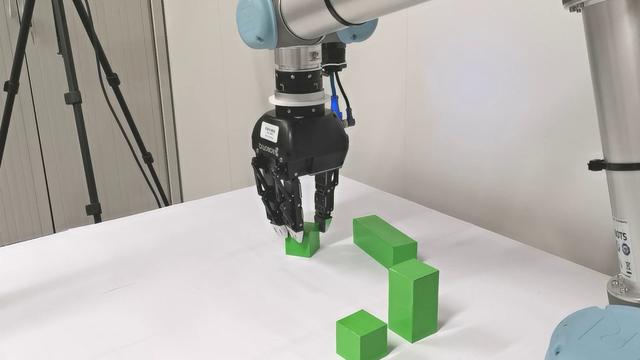}
  \includegraphics[trim=100 0 140 0, clip, width=.16\linewidth]{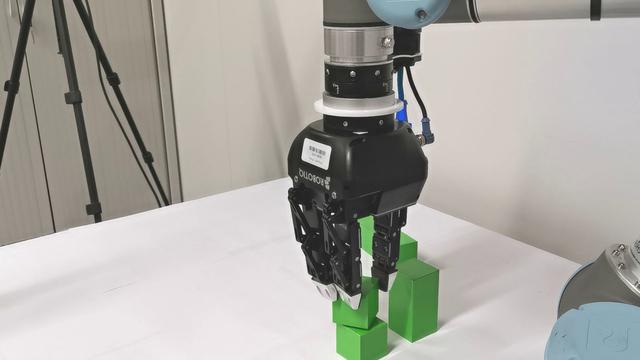}
  \includegraphics[trim=100 0 140 0, clip, width=.16\linewidth]{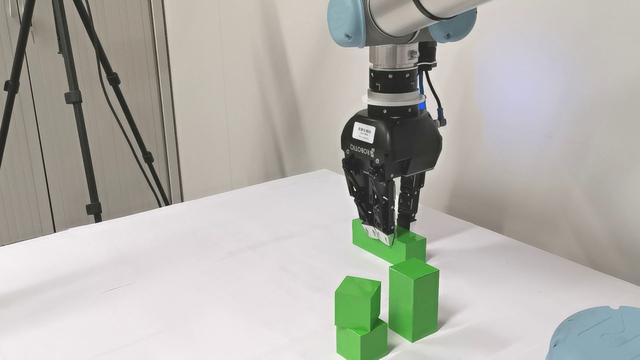}
  \includegraphics[trim=100 0 140 0, clip, width=.16\linewidth]{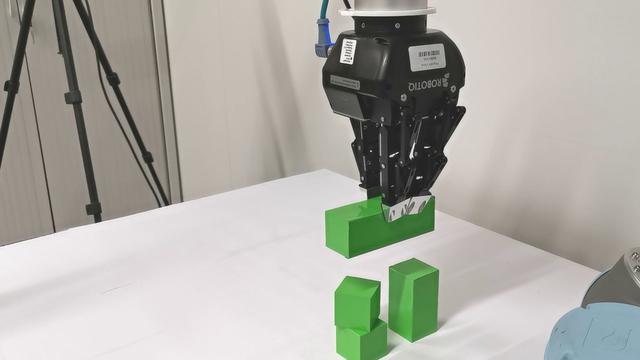}
  \includegraphics[trim=300 0 160 198, clip, width=.16\linewidth]{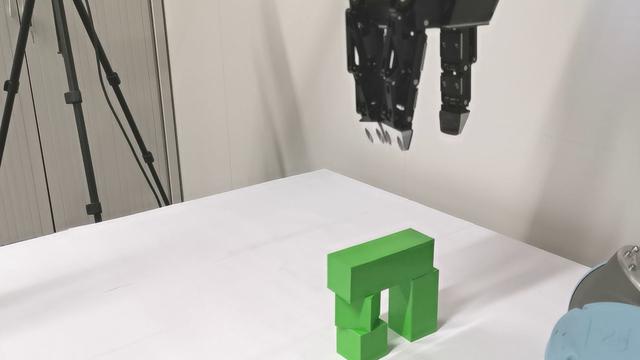}\vspace{.8cm}
  \includegraphics[trim=100 0 140 0, clip, width=.16\linewidth]{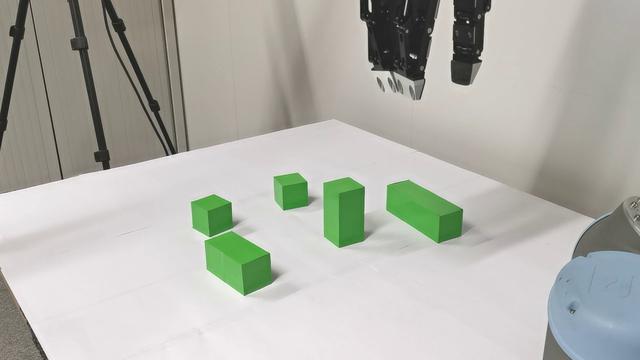}
  \includegraphics[trim=100 0 140 0, clip, width=.16\linewidth]{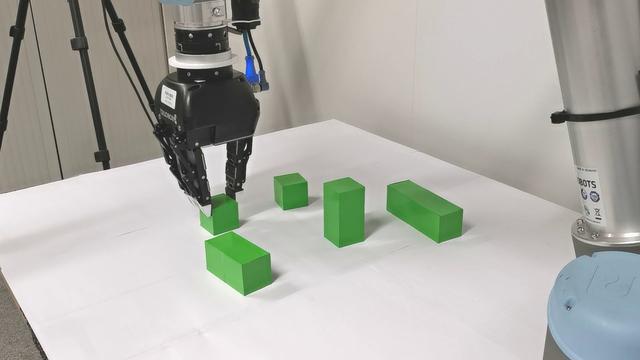}
  \includegraphics[trim=100 0 140 0, clip, width=.16\linewidth]{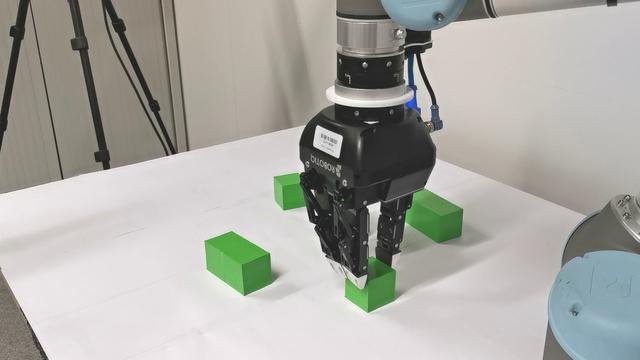}
  \includegraphics[trim=100 0 140 0, clip, width=.16\linewidth]{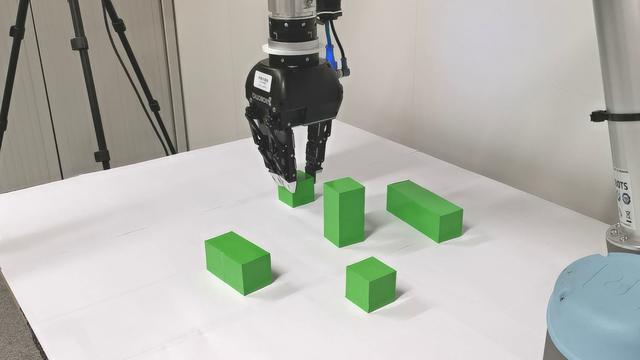}
  \includegraphics[trim=100 0 140 0, clip, width=.16\linewidth]{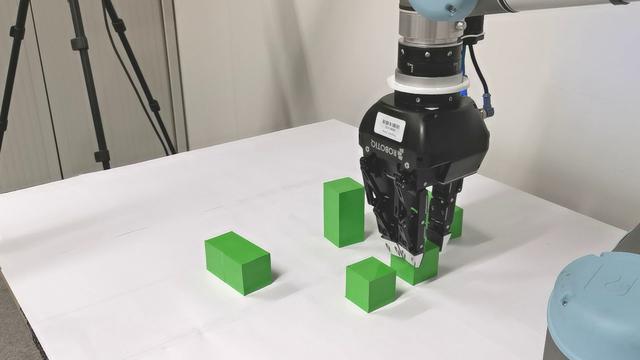}
  \includegraphics[trim=100 0 140 0, clip, width=.16\linewidth]{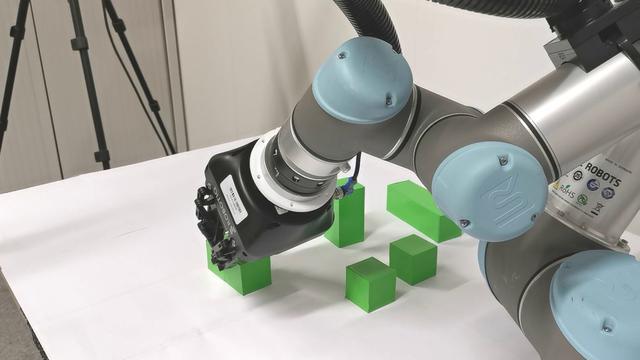}
  \includegraphics[trim=100 0 140 0, clip, width=.16\linewidth]{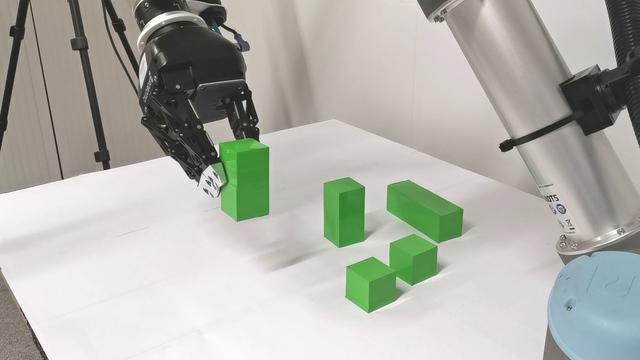}
  \includegraphics[trim=100 0 140 0, clip, width=.16\linewidth]{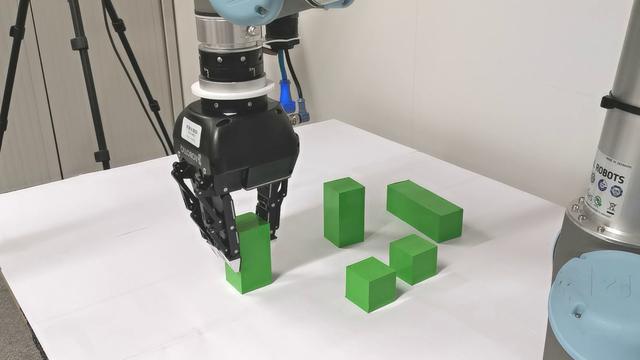}
  \includegraphics[trim=100 0 140 0, clip, width=.16\linewidth]{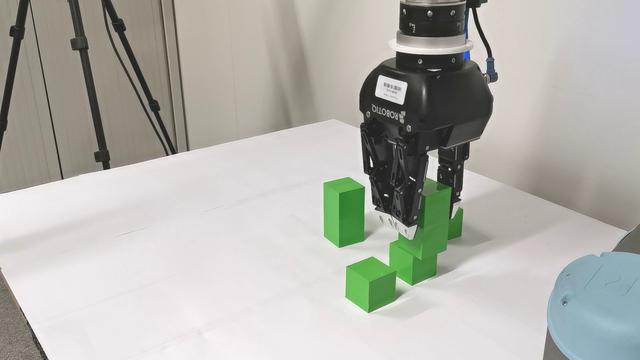}
  \includegraphics[trim=100 0 140 0, clip, width=.16\linewidth]{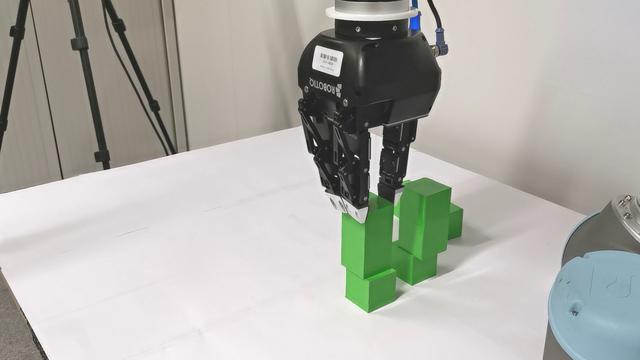}
  \includegraphics[trim=100 0 140 0, clip, width=.16\linewidth]{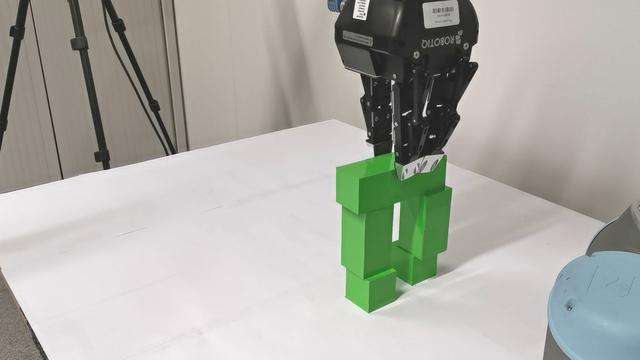}
  \includegraphics[trim=283 30 134 130, clip, width=.16\linewidth]{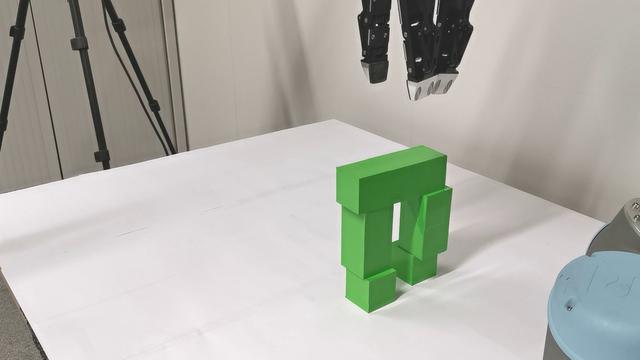}\vspace{.8cm}
  \includegraphics[trim=150 0 180 20, clip, width=.16\linewidth]{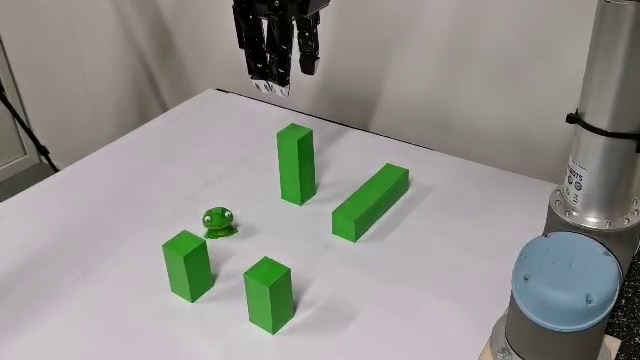}
  \includegraphics[trim=150 0 180 20, clip, width=.16\linewidth]{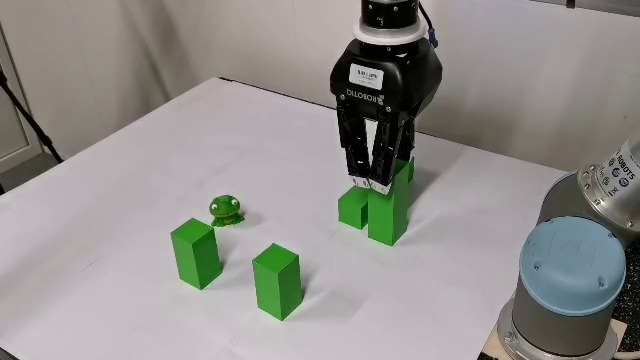}
  \includegraphics[trim=150 0 180 20, clip, width=.16\linewidth]{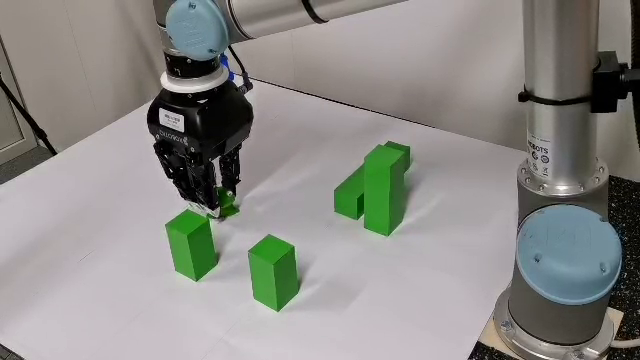}
  \includegraphics[trim=150 0 180 20, clip, width=.16\linewidth]{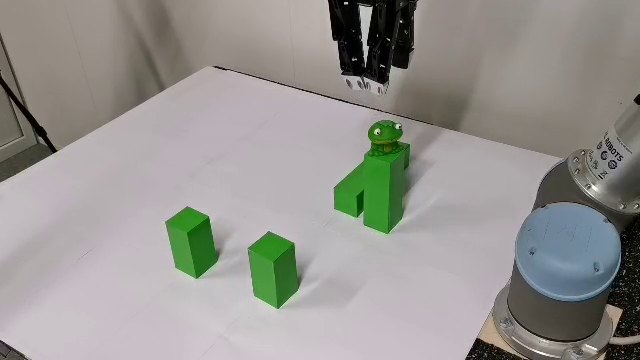}
  \includegraphics[trim=150 0 180 20, clip, width=.16\linewidth]{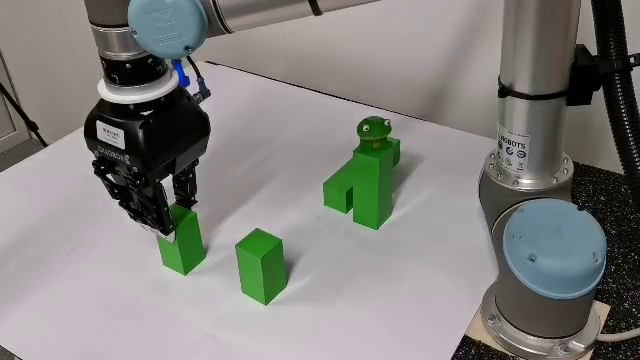}
  \includegraphics[trim=150 0 180 20, clip, width=.16\linewidth]{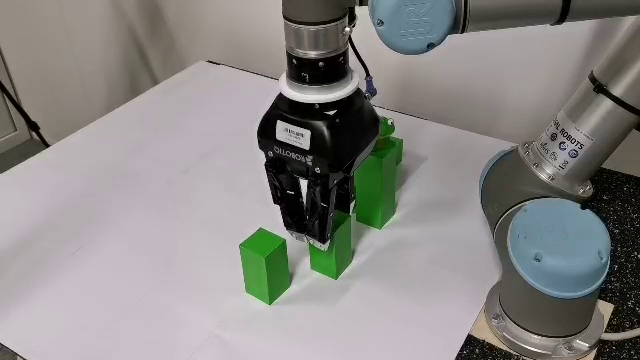}
  \includegraphics[trim=150 0 180 20, clip, width=.16\linewidth]{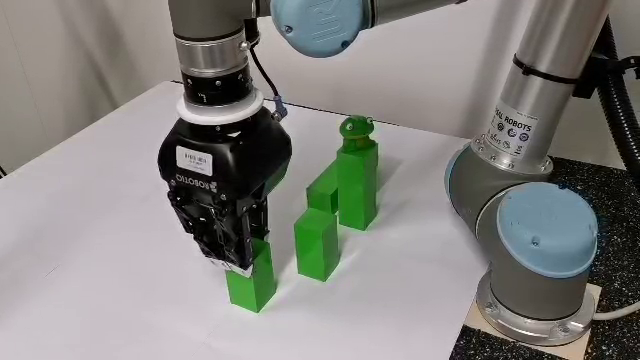}
  \includegraphics[trim=150 0 180 20, clip, width=.16\linewidth]{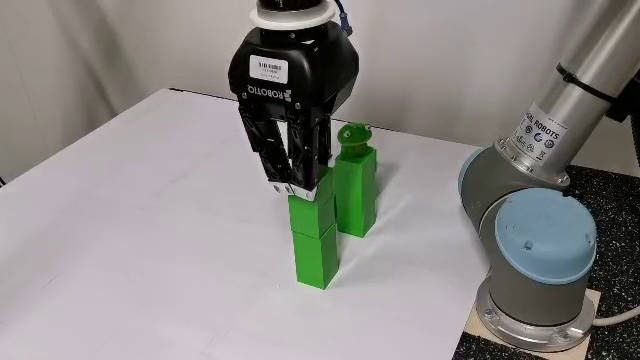}
  \includegraphics[trim=150 0 180 20, clip, width=.16\linewidth]{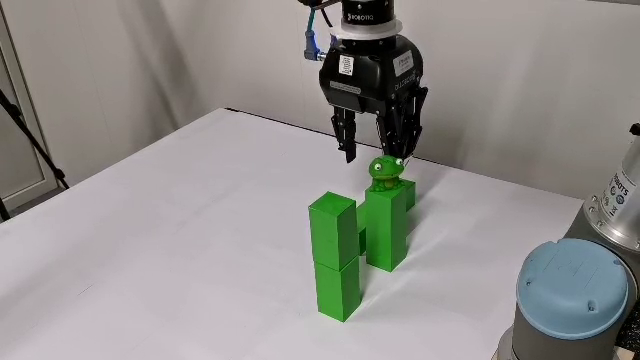}
  \includegraphics[trim=150 0 180 20, clip, width=.16\linewidth]{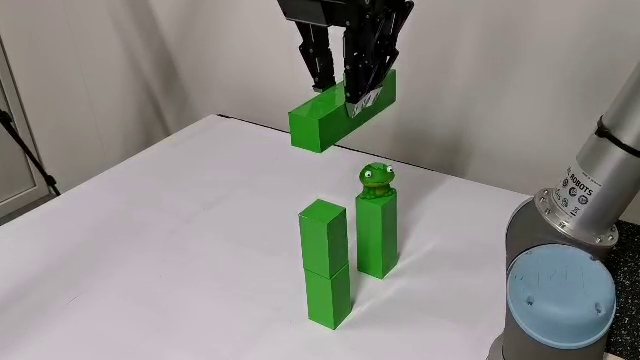}
  \includegraphics[trim=150 0 180 20, clip, width=.16\linewidth]{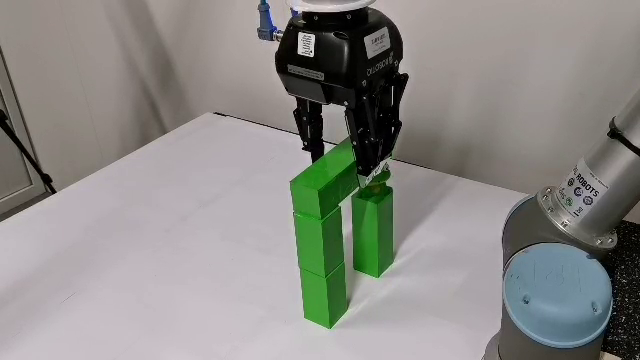}
  \includegraphics[trim=150 0 180 20, clip, width=.16\linewidth]{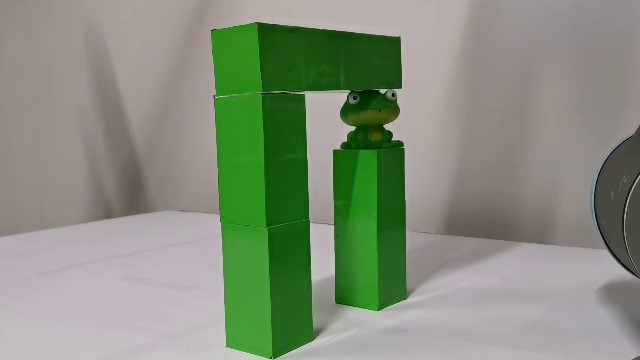}
  \caption{Successful construction of arches from blocks with a real robot. Note the rotation movements performed by the robot to change the orientation of certain blocks. }
  \label{fig:arch_blocks}
\end{figure*}

\begin{figure*}
  \centering
  %[trim=left bottom right top, clip]
  \includegraphics[trim=100 0 180 0, clip, width=.16\linewidth]{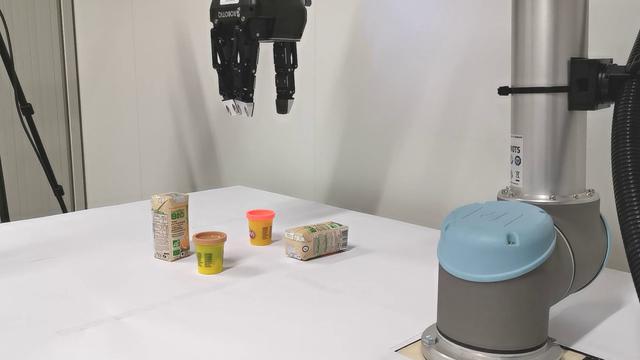}
  \includegraphics[trim=100 0 180 0, clip, width=.16\linewidth]{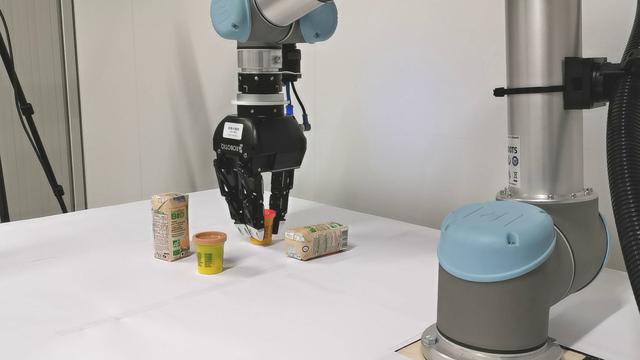}
  \includegraphics[trim=100 0 180 0, clip, width=.16\linewidth]{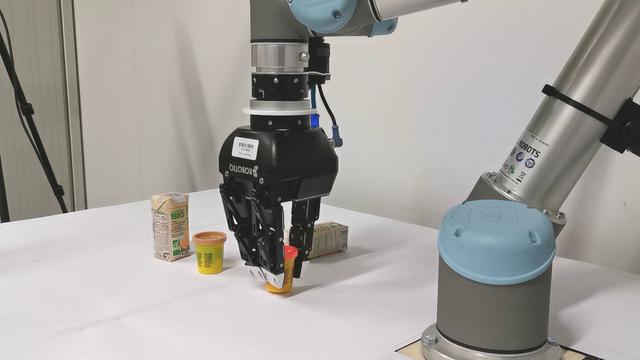}
  \includegraphics[trim=100 0 180 0, clip, width=.16\linewidth]{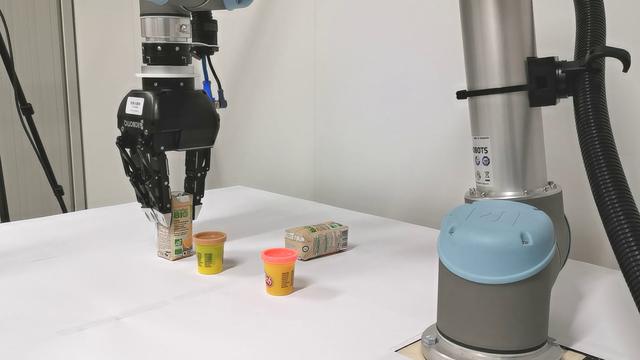}
  \includegraphics[trim=100 0 180 0, clip, width=.16\linewidth]{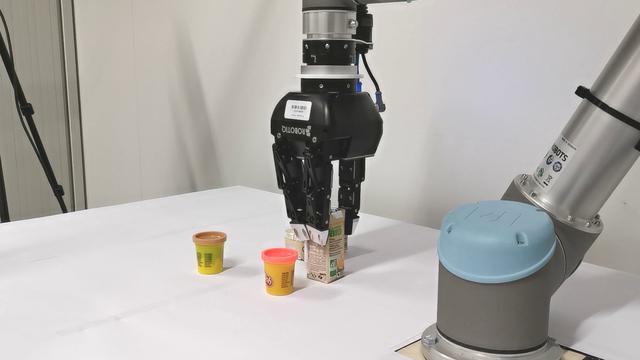}
  \includegraphics[trim=100 0 180 0, clip, width=.16\linewidth]{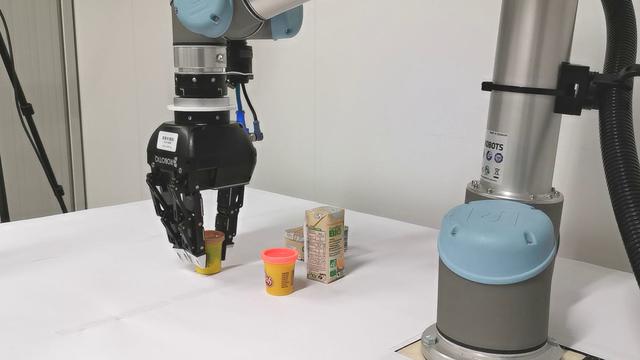}
  \includegraphics[trim=100 0 180 0, clip, width=.16\linewidth]{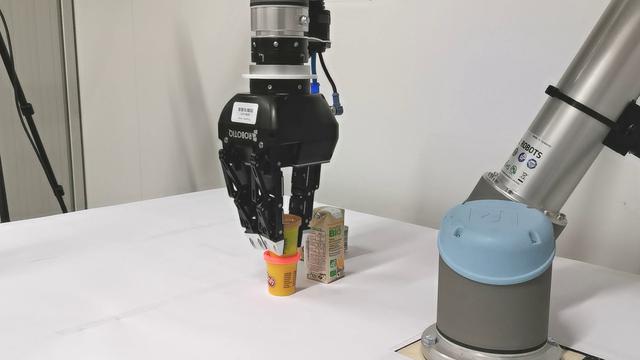}
  \includegraphics[trim=100 0 180 0, clip, width=.16\linewidth]{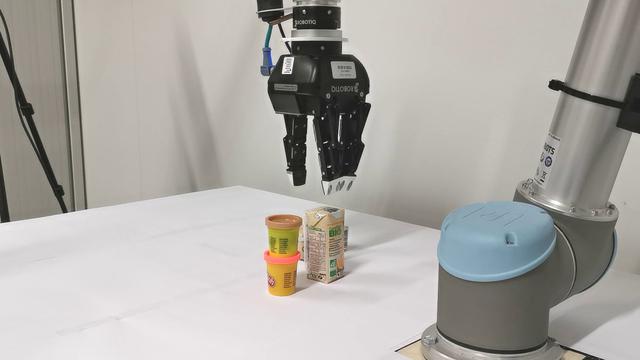}
  \includegraphics[trim=100 0 180 0, clip, width=.16\linewidth]{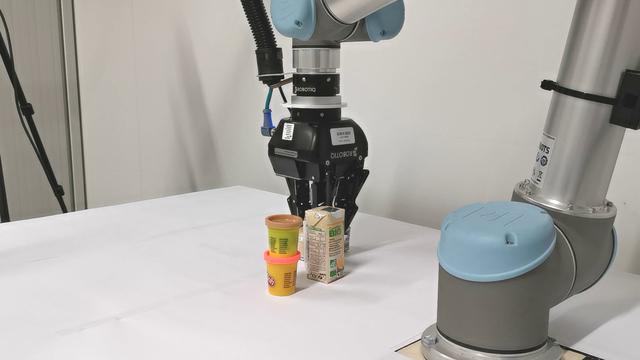}
  \includegraphics[trim=100 0 180 0, clip, width=.16\linewidth]{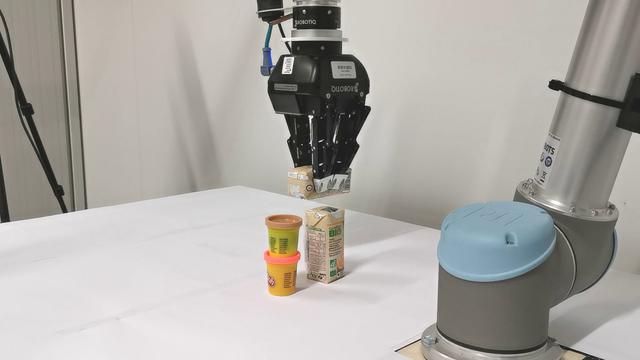}
  \includegraphics[trim=100 0 180 0, clip, width=.16\linewidth]{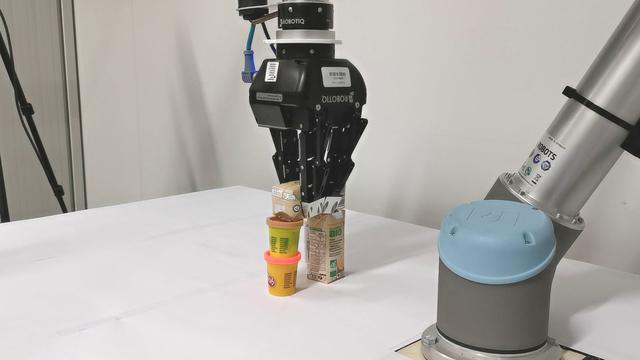}
  \includegraphics[trim=216 46 250 140, clip, width=.16\linewidth]{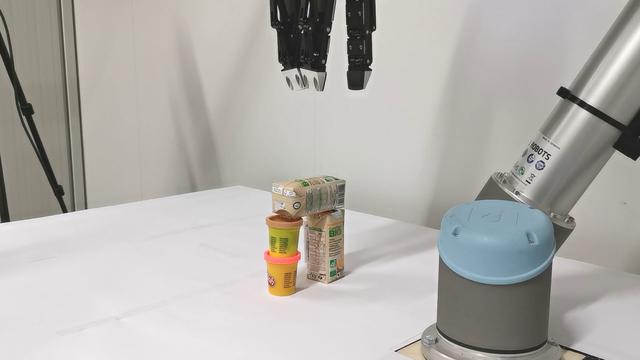}\vspace{.8cm}
  \includegraphics[trim=100 0 140 0, clip, width=.16\linewidth]{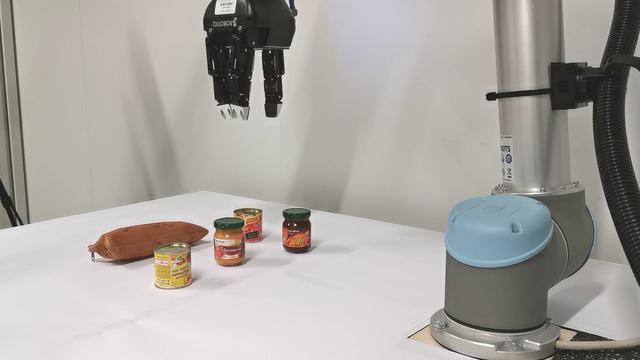}
  \includegraphics[trim=100 0 140 0, clip, width=.16\linewidth]{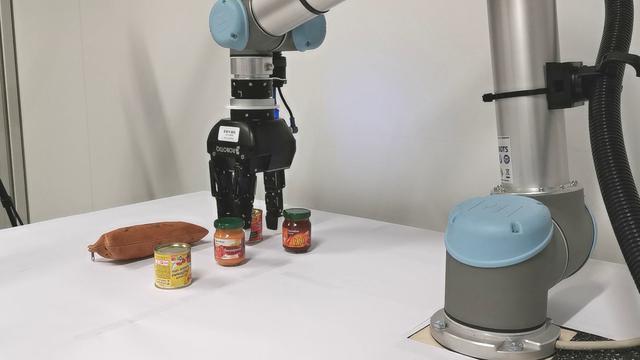}
  \includegraphics[trim=100 0 140 0, clip, width=.16\linewidth]{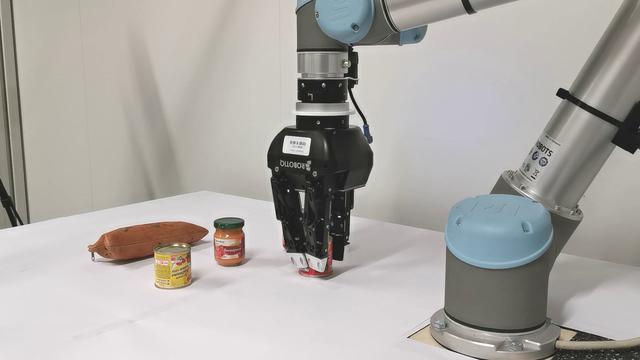}
  \includegraphics[trim=100 0 140 0, clip, width=.16\linewidth]{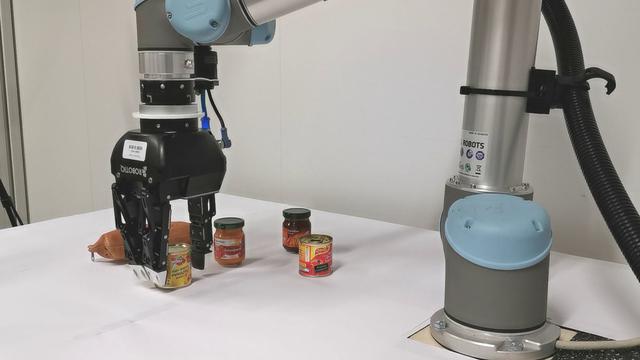}
  \includegraphics[trim=100 0 140 0, clip, width=.16\linewidth]{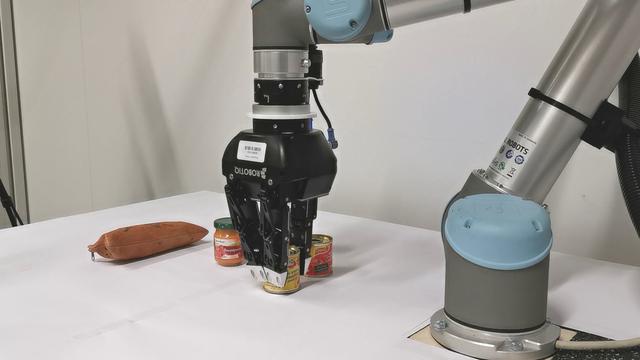}
  \includegraphics[trim=100 0 140 0, clip, width=.16\linewidth]{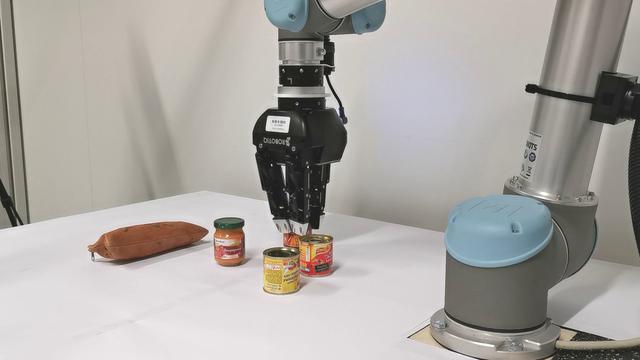}
  \includegraphics[trim=100 0 140 0, clip, width=.16\linewidth]{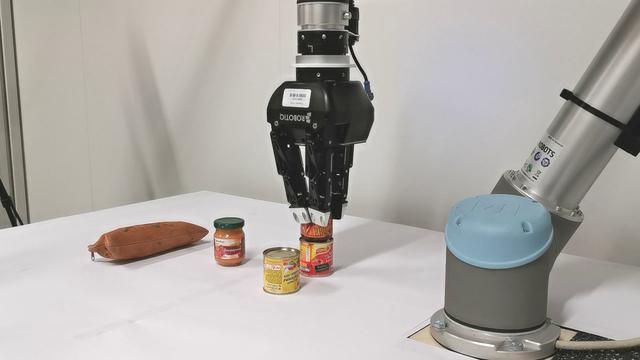}
  \includegraphics[trim=100 0 140 0, clip, width=.16\linewidth]{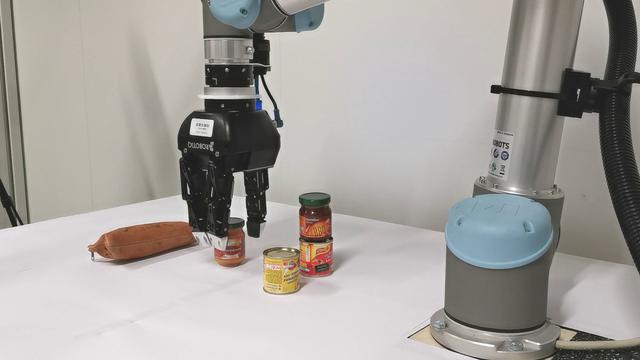}
  \includegraphics[trim=100 0 140 0, clip, width=.16\linewidth]{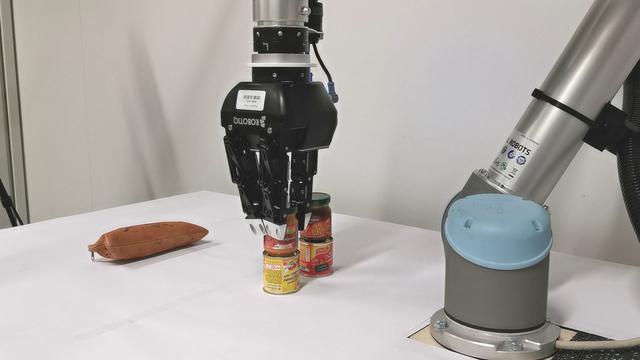}
  \includegraphics[trim=100 0 140 0, clip, width=.16\linewidth]{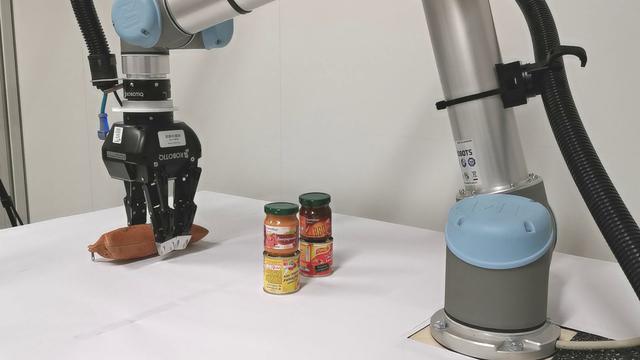}
  \includegraphics[trim=100 0 140 0, clip, width=.16\linewidth]{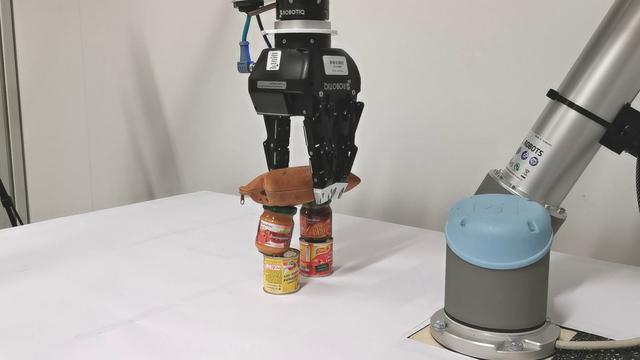}
  \includegraphics[trim=216 60 250 144, clip, width=.16\linewidth]{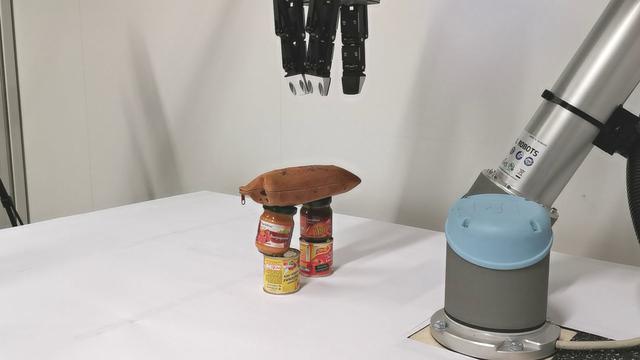}\vspace{.8cm}
  \includegraphics[trim=100 0 140 0, clip, width=.16\linewidth]{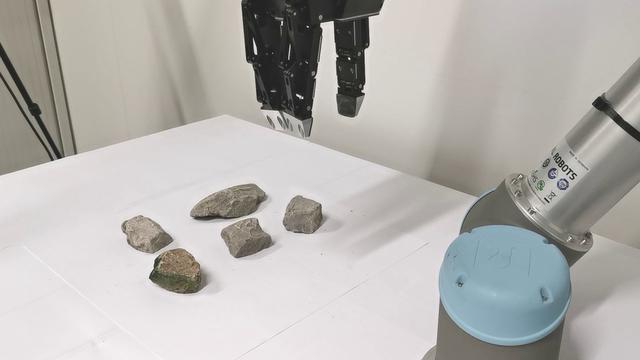}
  \includegraphics[trim=100 0 140 0, clip, width=.16\linewidth]{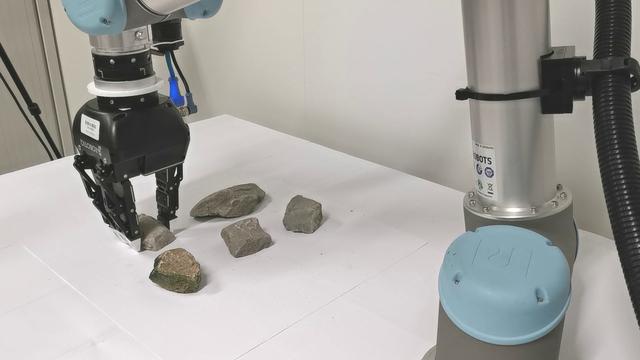}
  \includegraphics[trim=100 0 140 0, clip, width=.16\linewidth]{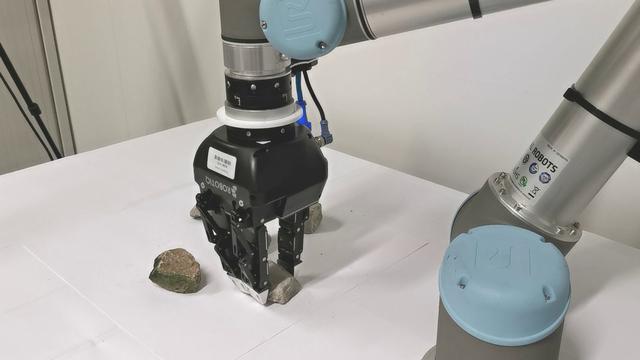}
  \includegraphics[trim=100 0 140 0, clip, width=.16\linewidth]{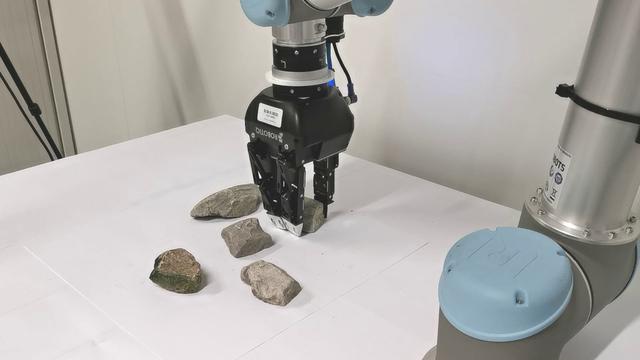}
  \includegraphics[trim=100 0 140 0, clip, width=.16\linewidth]{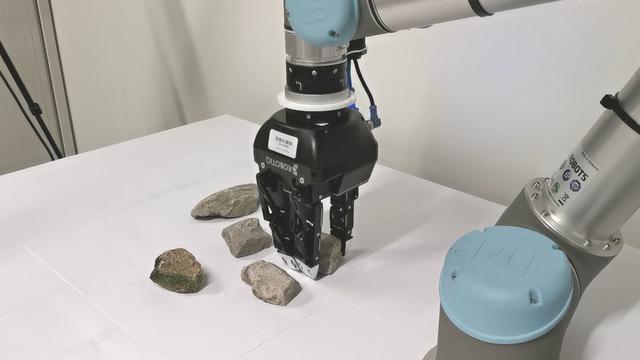}
  \includegraphics[trim=100 0 140 0, clip, width=.16\linewidth]{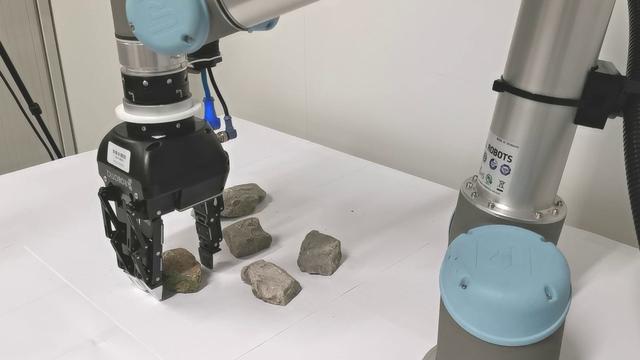}
  \includegraphics[trim=100 0 140 0, clip, width=.16\linewidth]{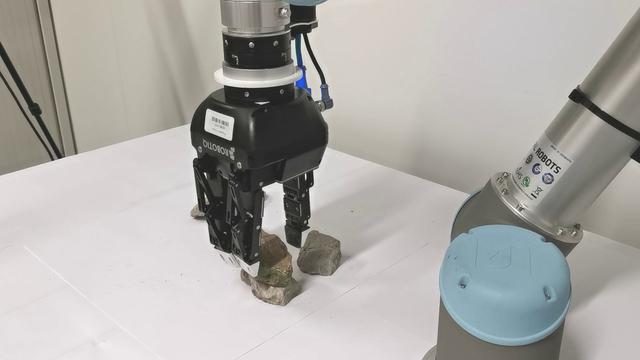}
  \includegraphics[trim=100 0 140 0, clip, width=.16\linewidth]{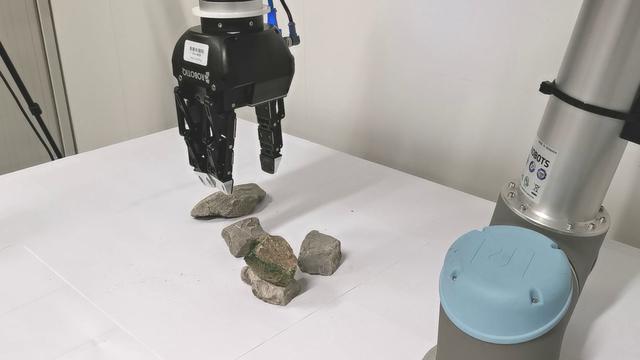}
  \includegraphics[trim=100 0 140 0, clip, width=.16\linewidth]{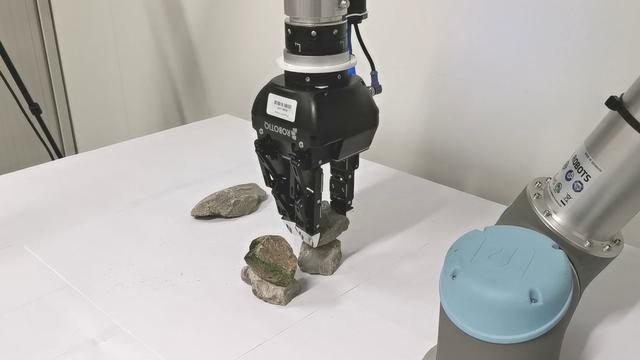}
  \includegraphics[trim=100 0 140 0, clip, width=.16\linewidth]{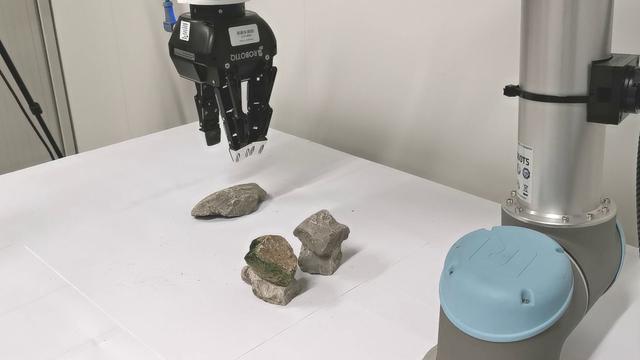}
  \includegraphics[trim=100 0 140 0, clip, width=.16\linewidth]{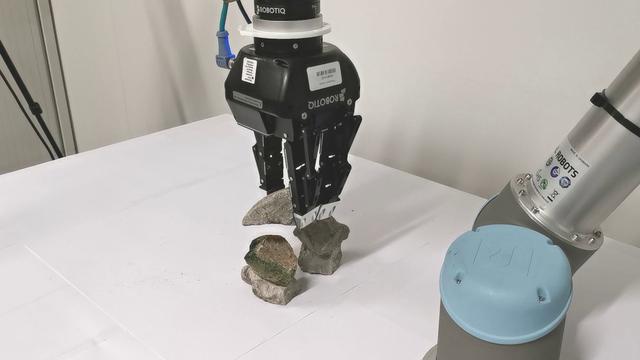}
  \includegraphics[trim=153 0 86 0, clip, width=.16\linewidth]{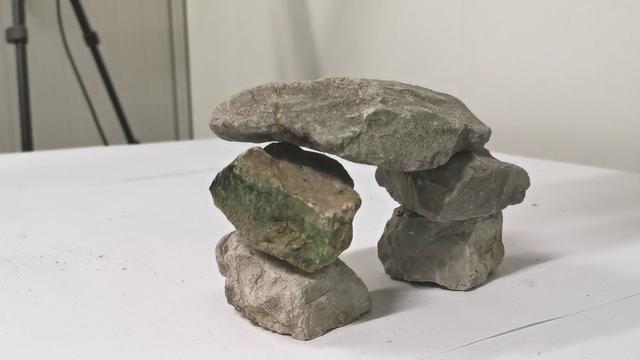}
  \caption{Construction of arches from new primitives that have not been observed during training. Despite the fact that our visual policy has been trained only with block-shaped primitives and only in simulation, it is able to generalize to new diverse primitives and to construct arches in a real robot setup.}
  \label{fig:arch_new}
\end{figure*}

\begin{figure*}
  \centering
  %[trim=left bottom right top, clip]
  \includegraphics[trim=100 0 140 0, clip, width=.16\linewidth]{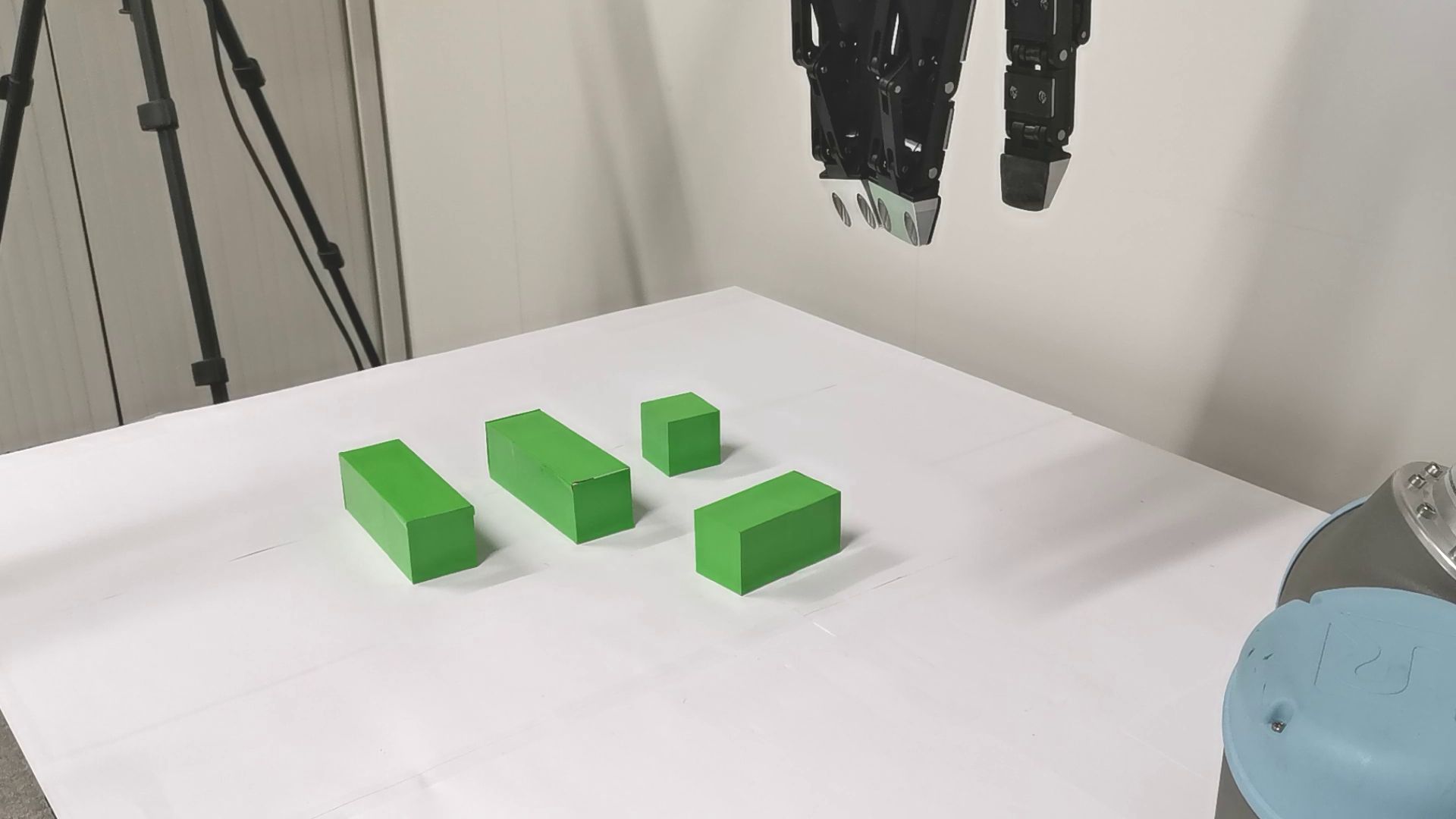}
  \includegraphics[trim=100 0 140 0, clip, width=.16\linewidth]{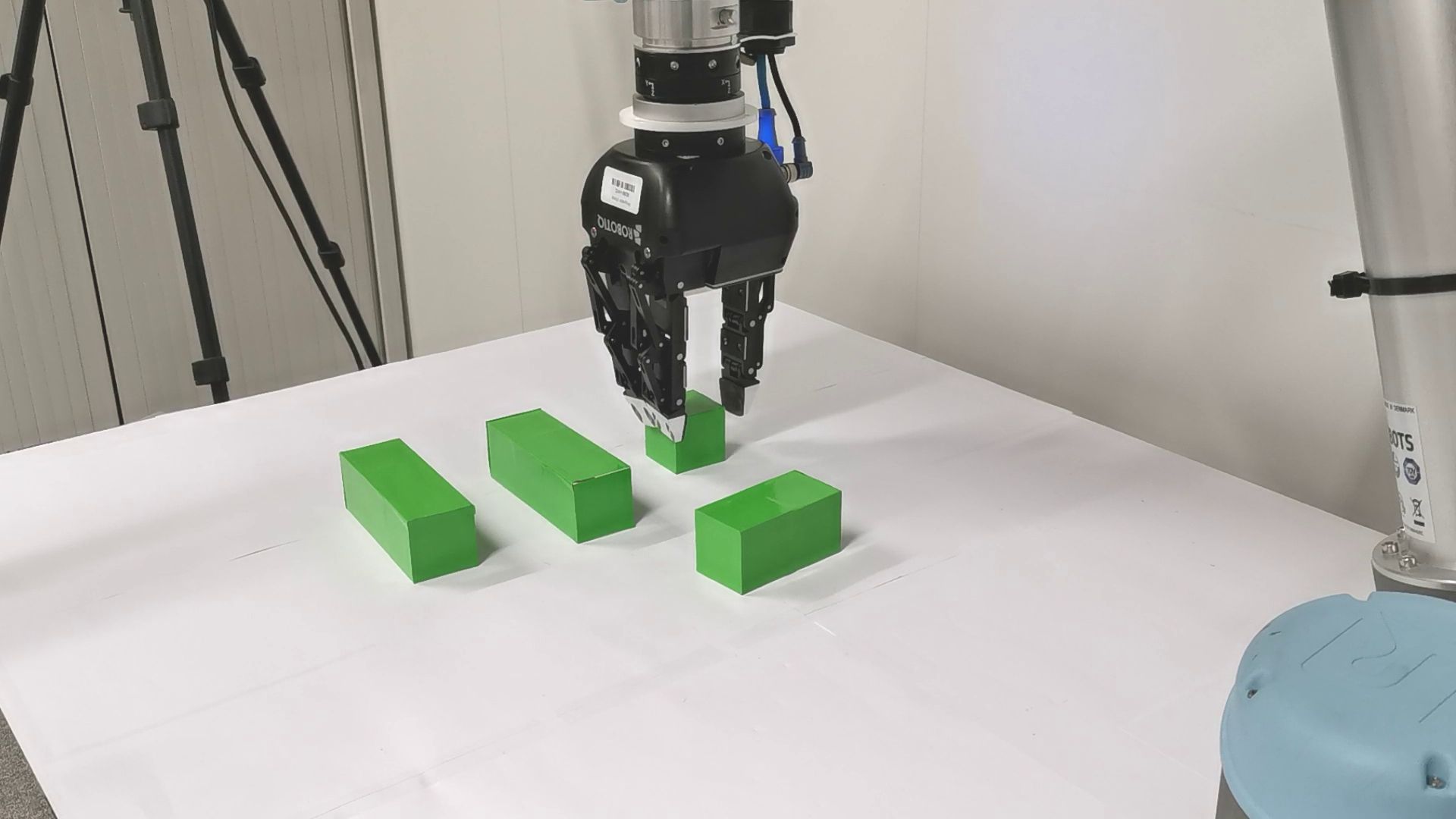}
  \includegraphics[trim=100 0 140 0, clip, width=.16\linewidth]{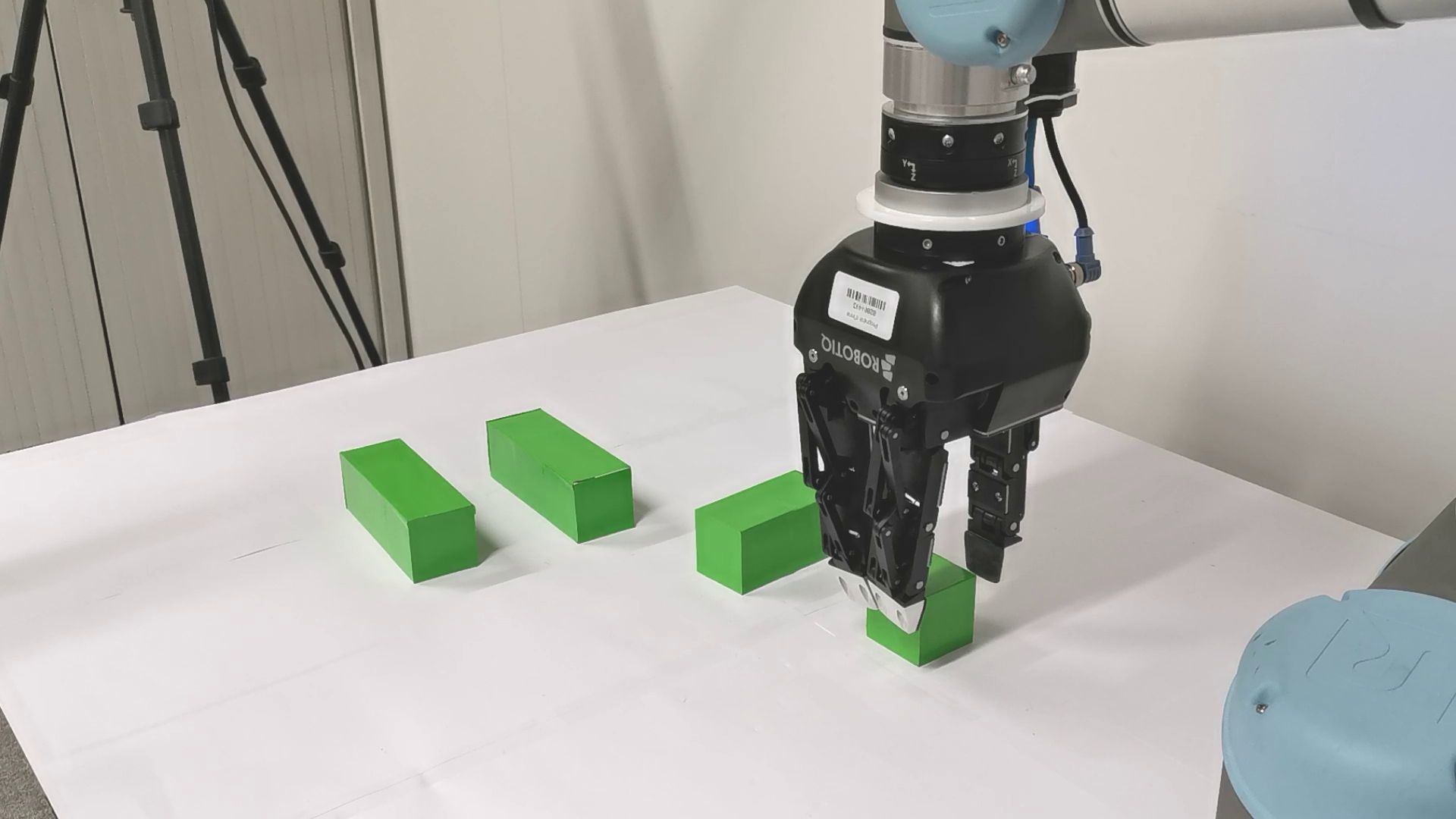}
  \includegraphics[trim=100 0 140 0, clip, width=.16\linewidth]{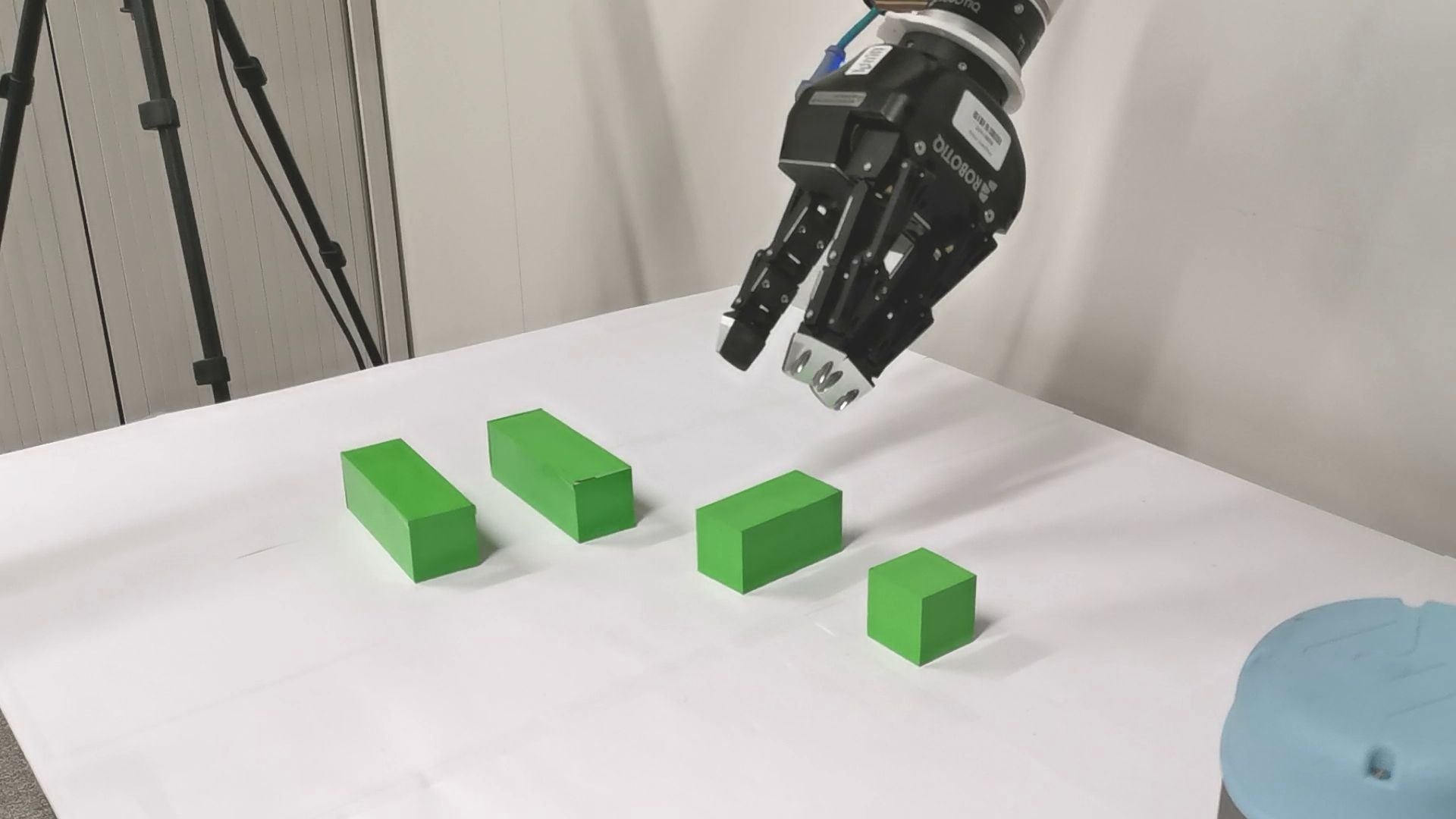}
  \includegraphics[trim=100 0 140 0, clip, width=.16\linewidth]{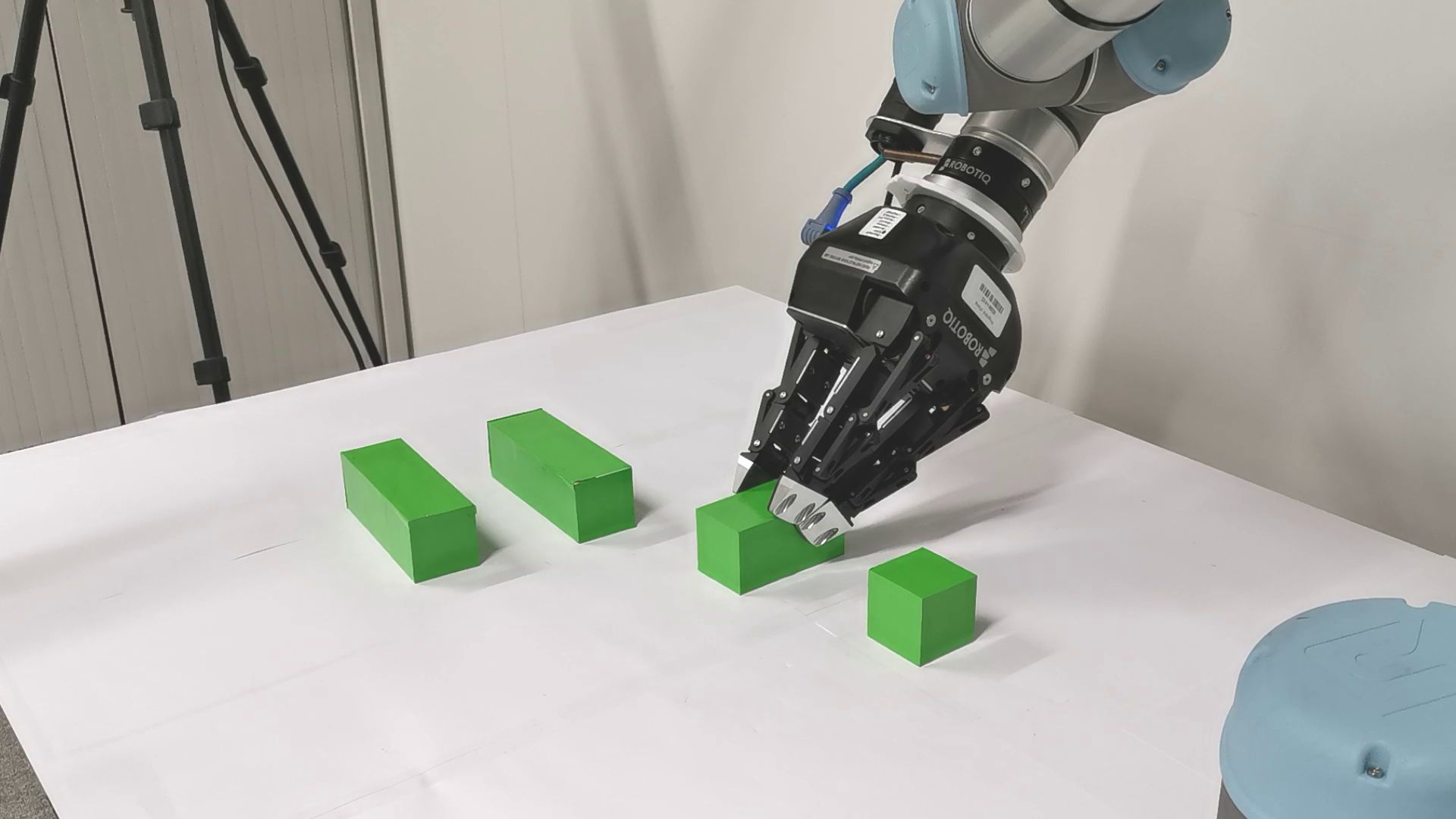}
  \includegraphics[trim=100 0 140 0, clip, width=.16\linewidth]{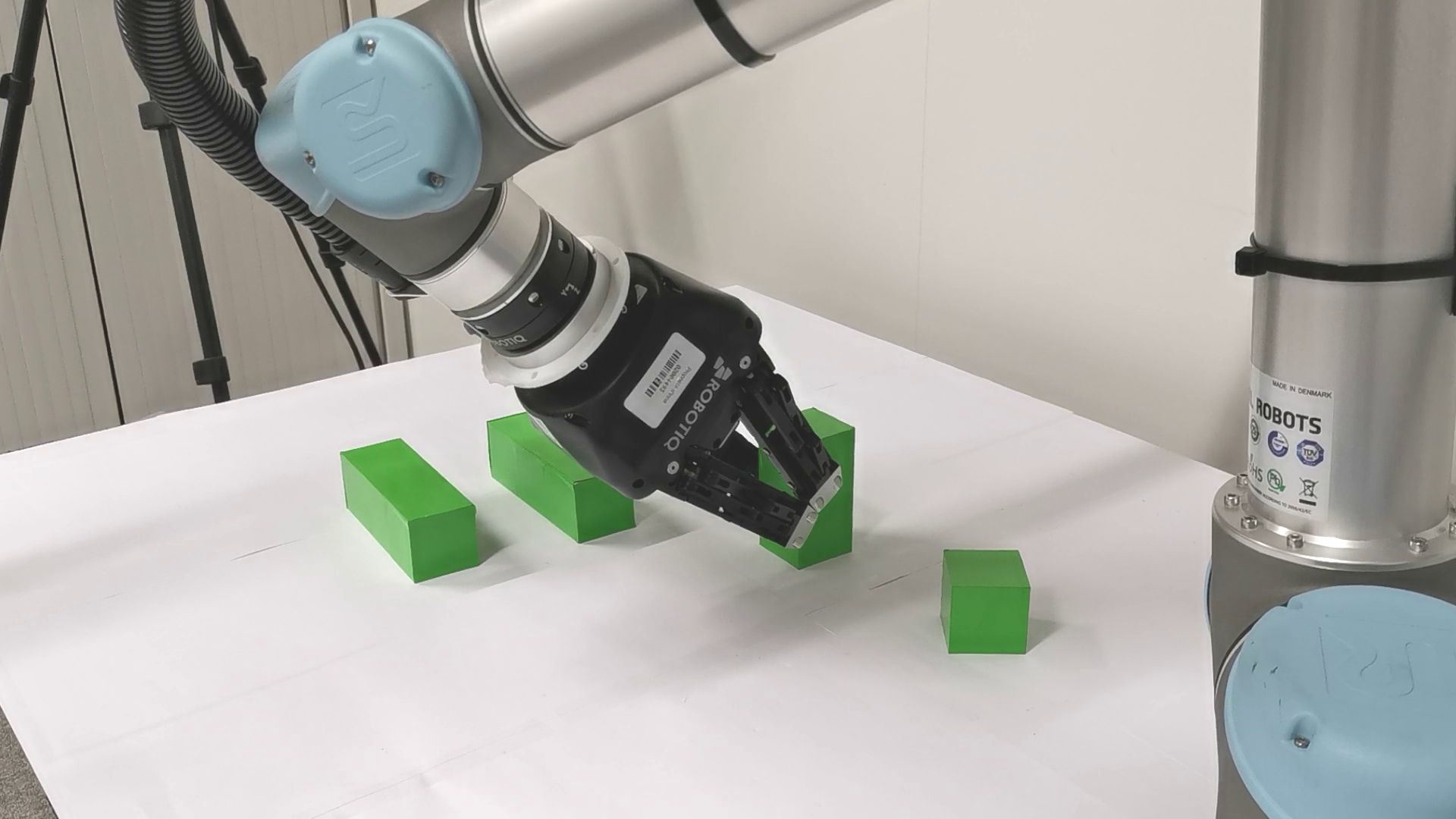}
  \includegraphics[trim=100 0 140 0, clip, width=.16\linewidth]{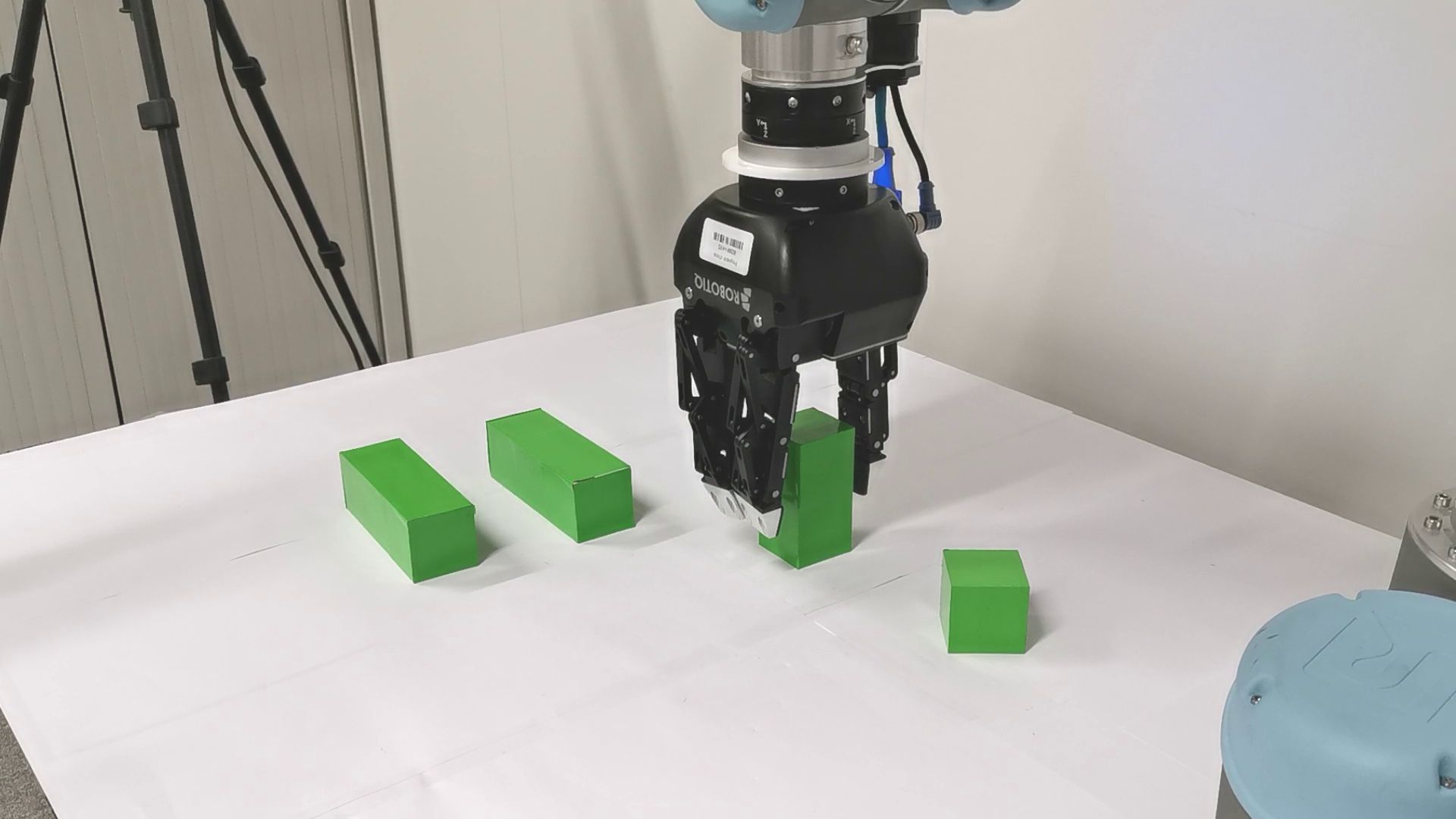}
  \includegraphics[trim=100 0 140 0, clip, width=.16\linewidth]{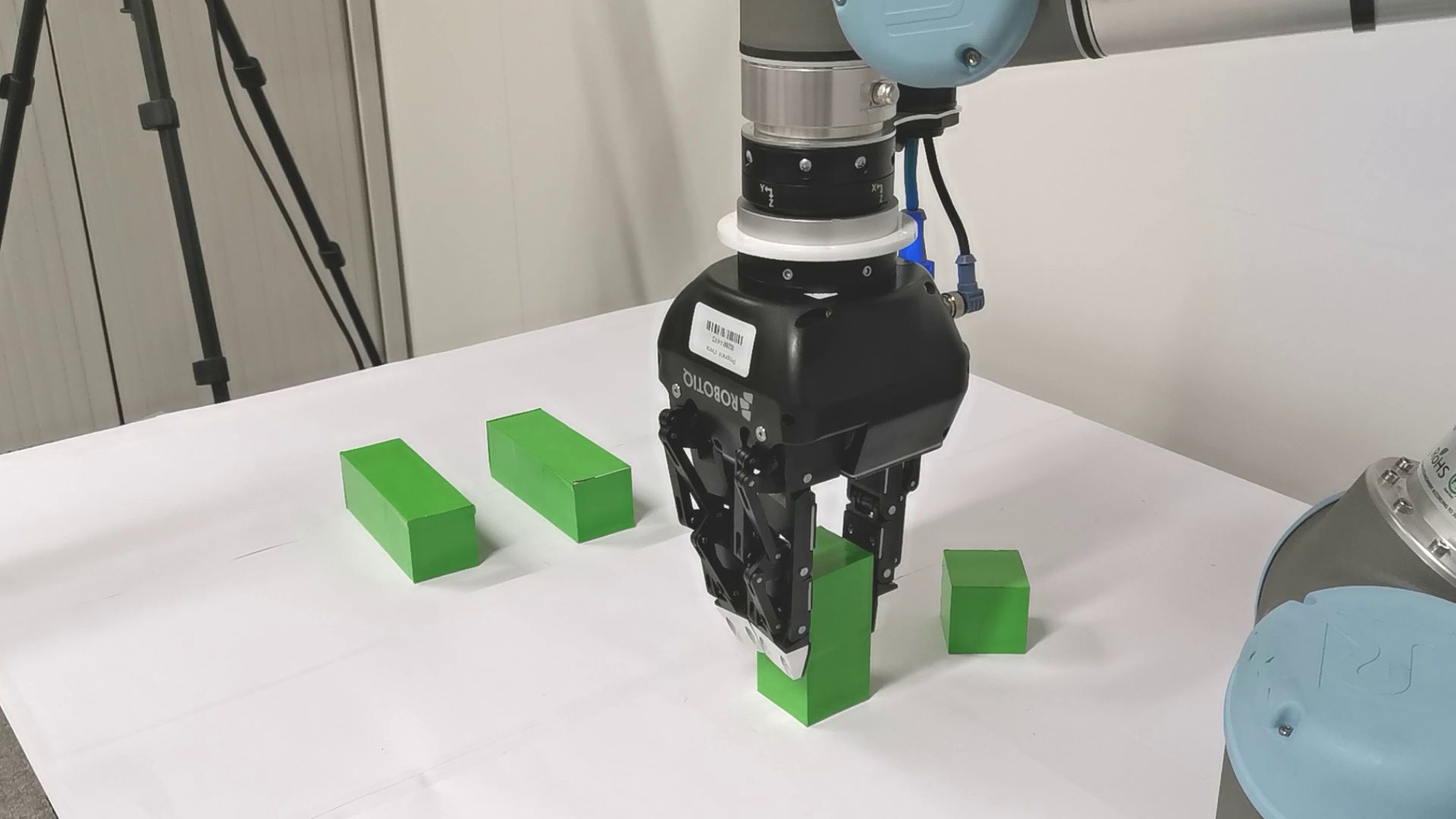}
  \includegraphics[trim=100 0 140 0, clip, width=.16\linewidth]{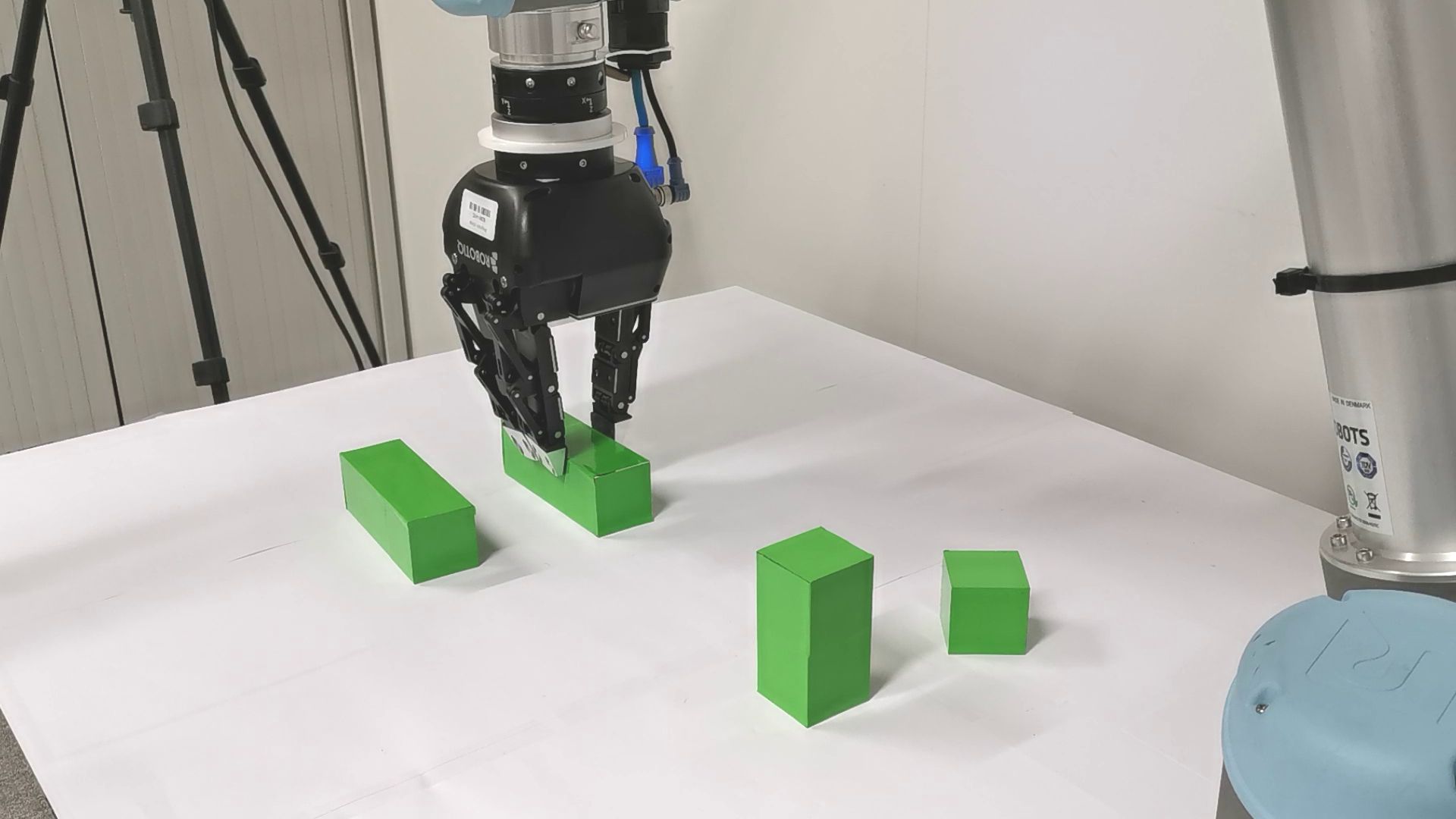}
  \includegraphics[trim=100 0 140 0, clip, width=.16\linewidth]{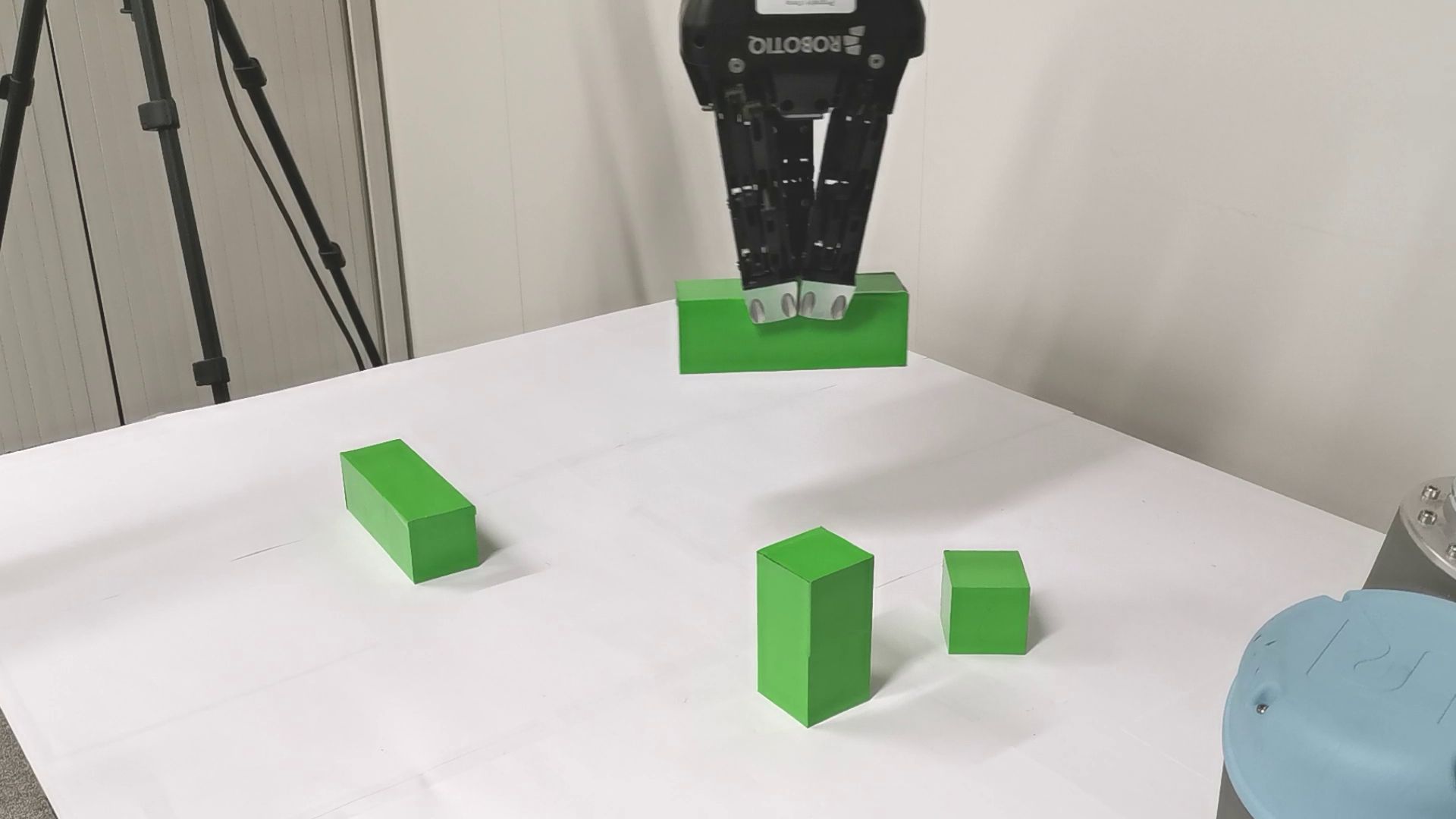}
  \includegraphics[trim=100 0 140 0, clip, width=.16\linewidth]{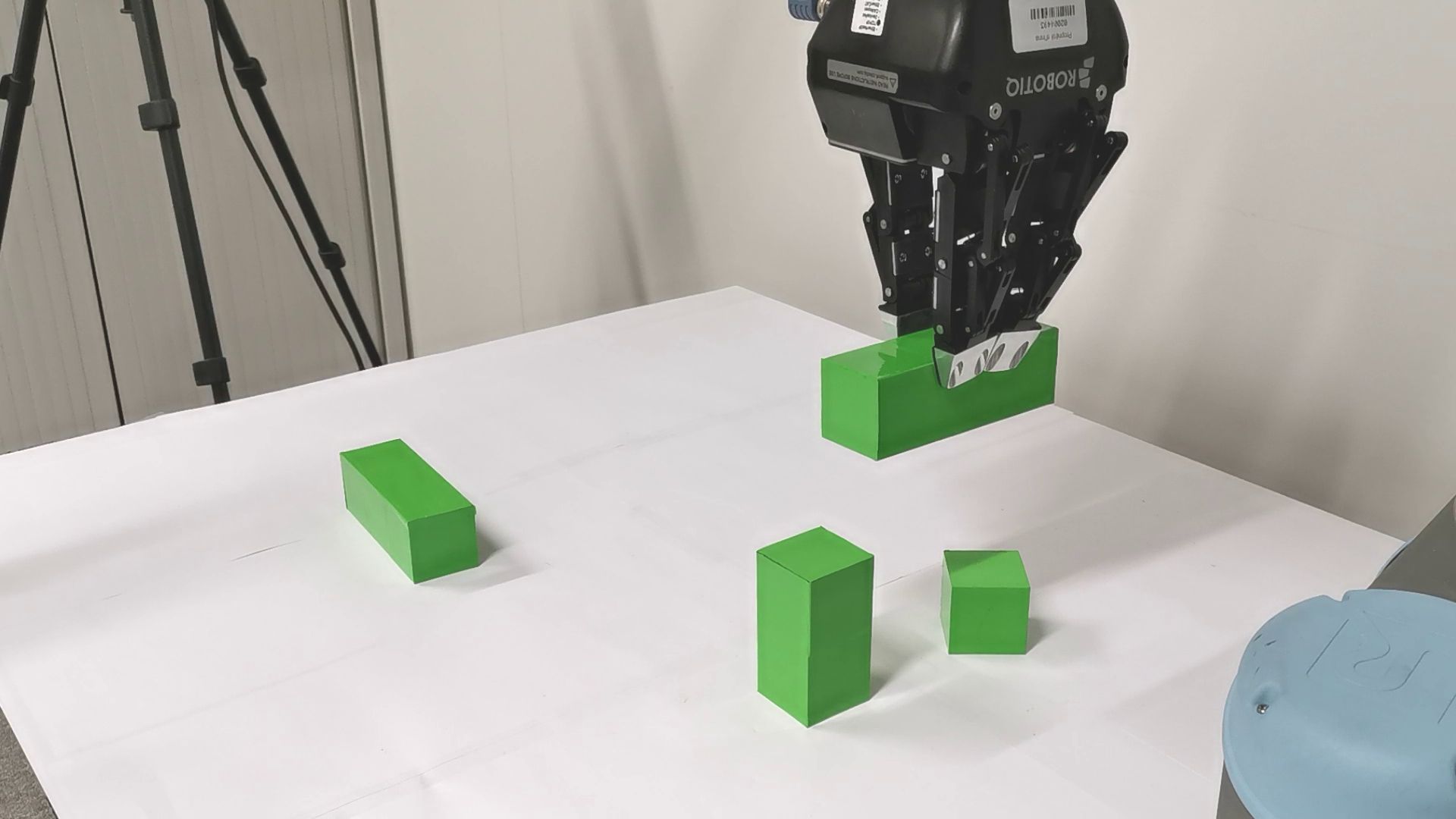}
  \includegraphics[trim=100 0 140 0, clip, width=.16\linewidth]{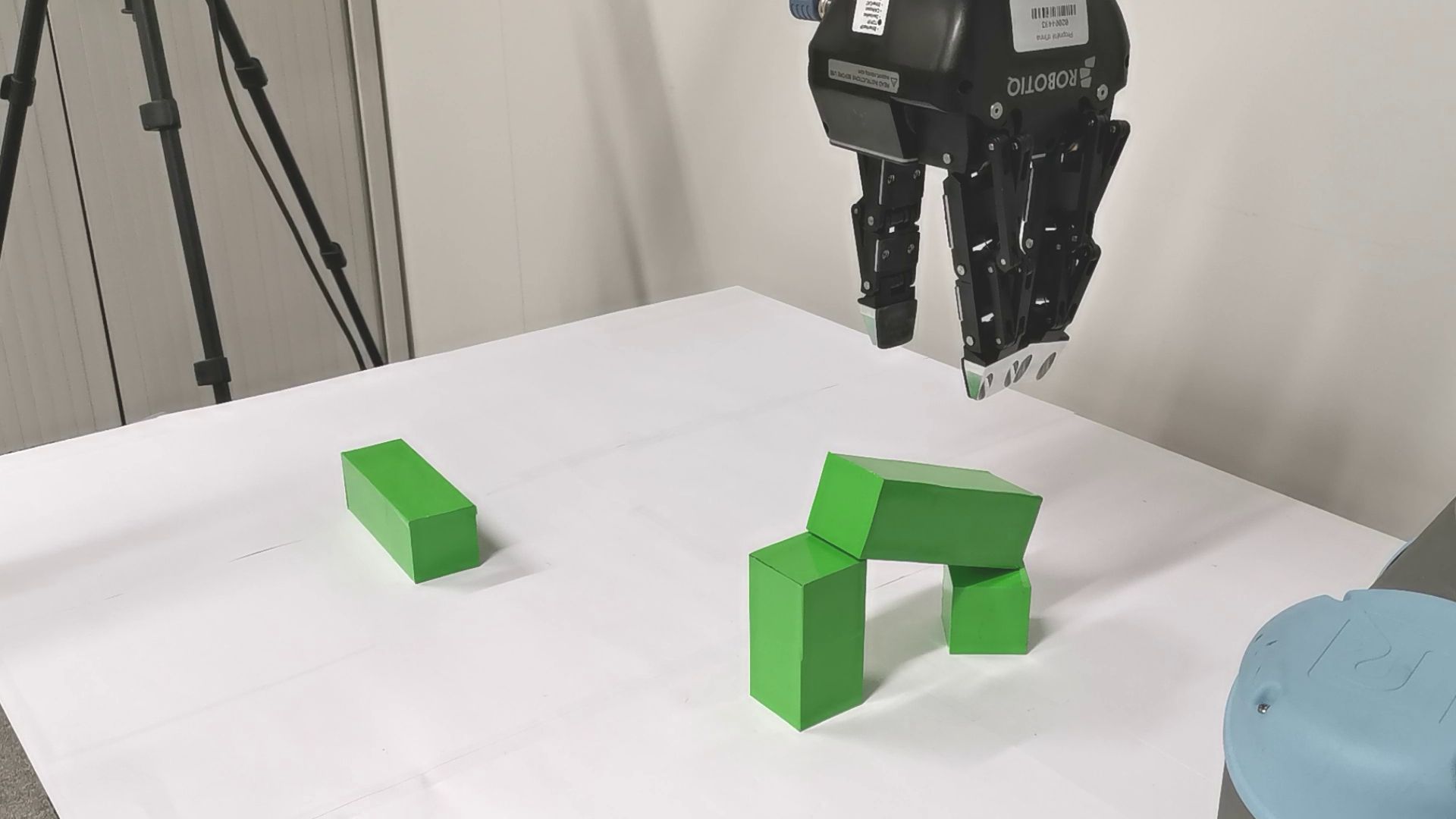}\vspace{.8cm}
  \includegraphics[trim=100 0 140 0, clip, width=.16\linewidth]{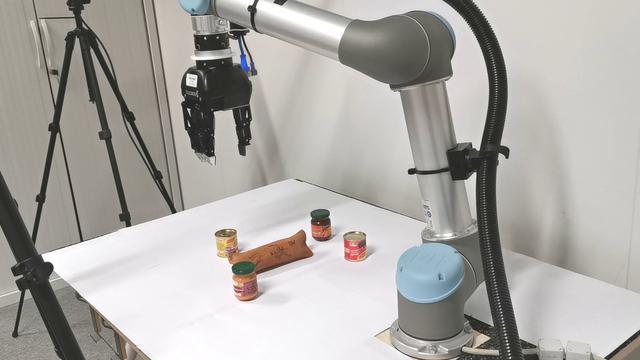}
  \includegraphics[trim=100 0 140 0, clip, width=.16\linewidth]{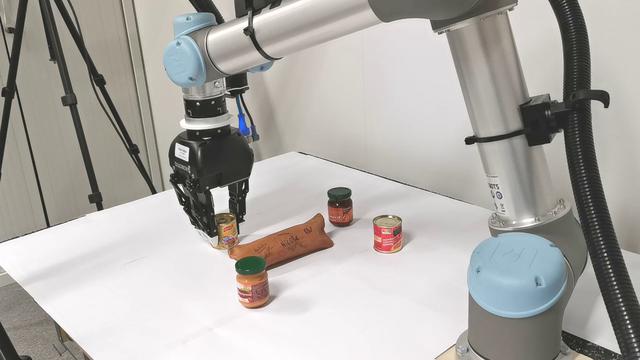}
  \includegraphics[trim=100 0 140 0, clip, width=.16\linewidth]{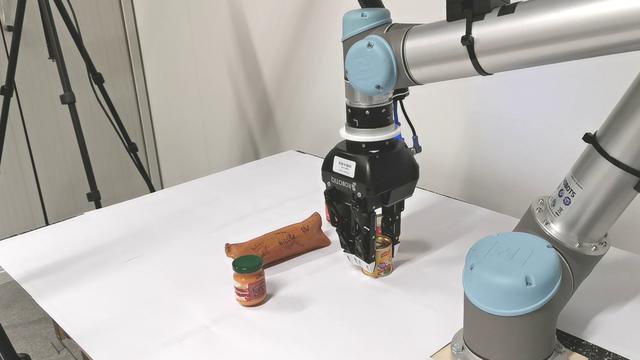}
  \includegraphics[trim=100 0 140 0, clip, width=.16\linewidth]{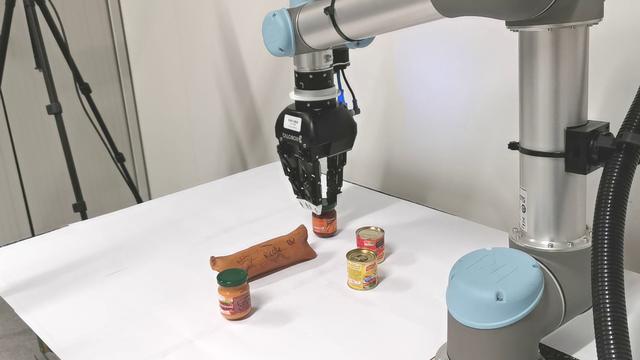}
  \includegraphics[trim=100 0 140 0, clip, width=.16\linewidth]{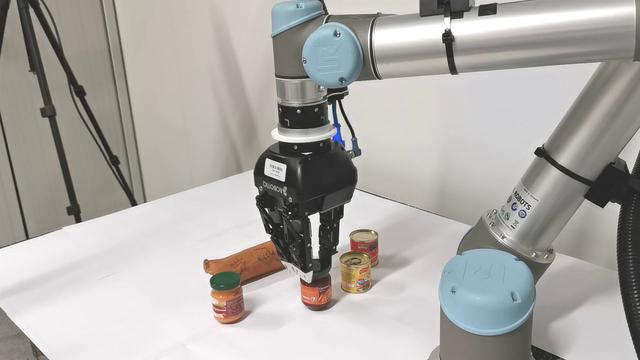}
  \includegraphics[trim=100 0 140 0, clip, width=.16\linewidth]{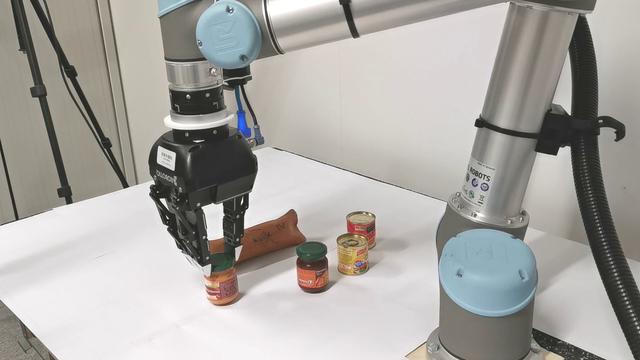}
  \includegraphics[trim=100 0 140 0, clip, width=.16\linewidth]{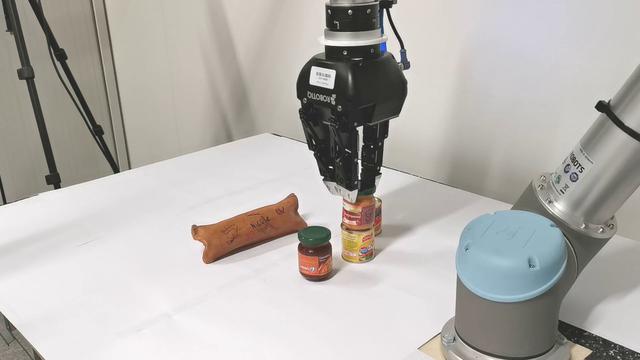}
  \includegraphics[trim=100 0 140 0, clip, width=.16\linewidth]{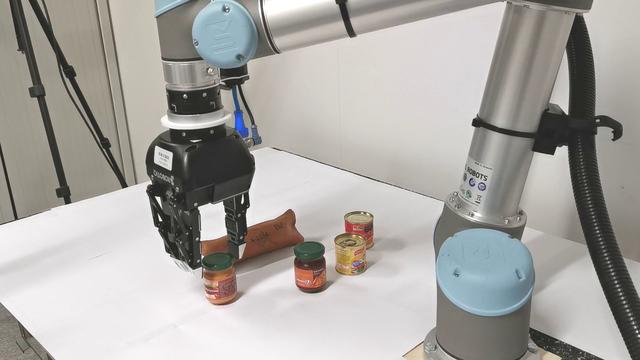}
  \includegraphics[trim=100 0 140 0, clip, width=.16\linewidth]{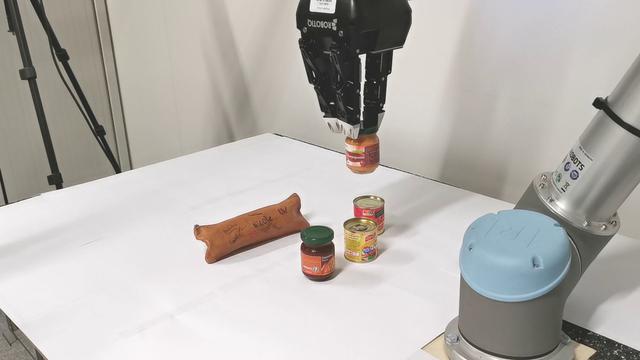}
  \includegraphics[trim=100 0 140 0, clip, width=.16\linewidth]{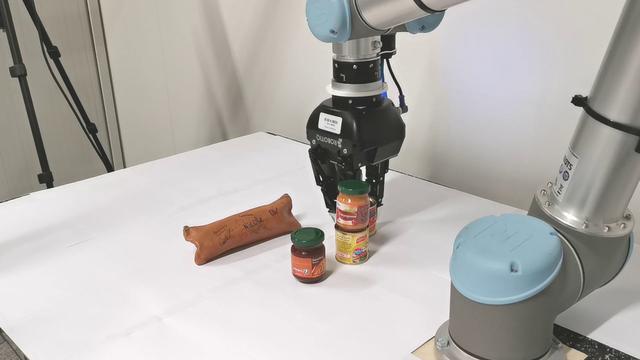}
  \includegraphics[trim=100 0 140 0, clip, width=.16\linewidth]{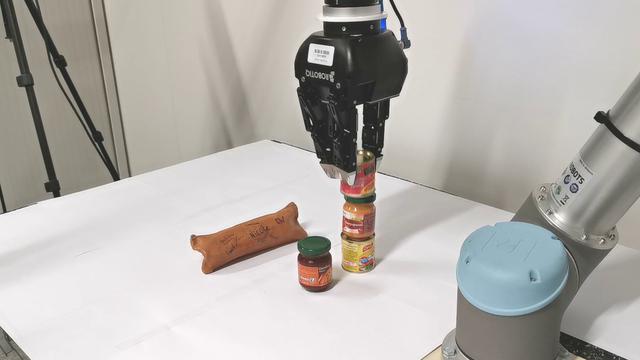}
  \includegraphics[trim=100 0 140 0, clip, width=.16\linewidth]{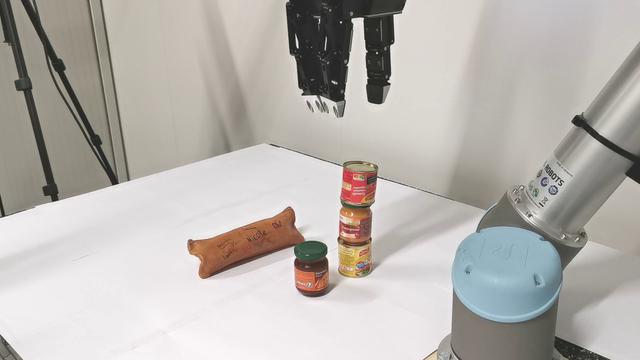}\vspace{.8cm}
  \includegraphics[trim=100 0 140 0, clip, width=.16\linewidth]{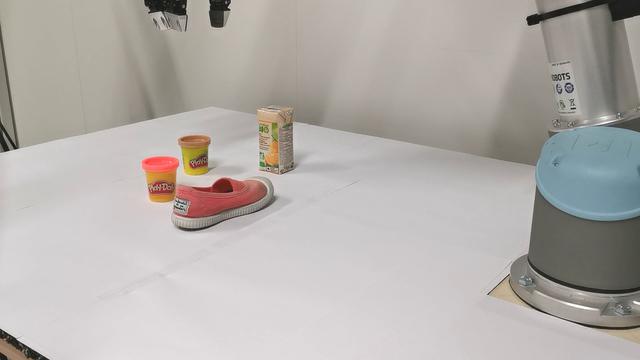}
  \includegraphics[trim=100 0 140 0, clip, width=.16\linewidth]{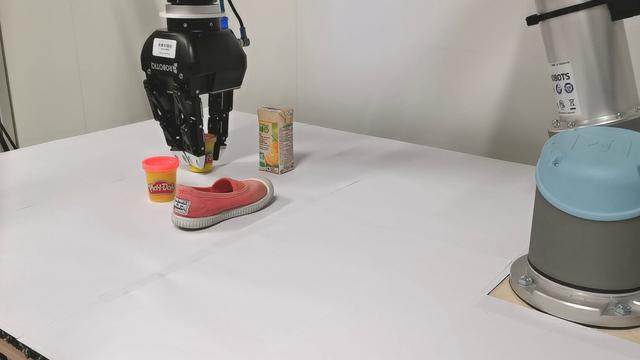}
  \includegraphics[trim=100 0 140 0, clip, width=.16\linewidth]{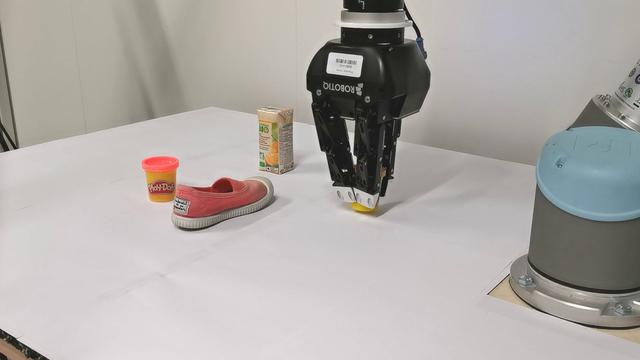}
  \includegraphics[trim=100 0 140 0, clip, width=.16\linewidth]{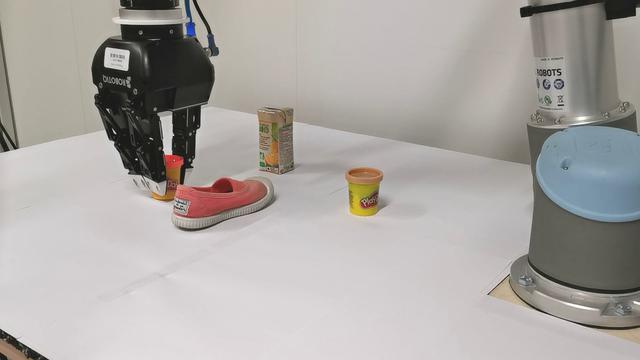}
  \includegraphics[trim=100 0 140 0, clip, width=.16\linewidth]{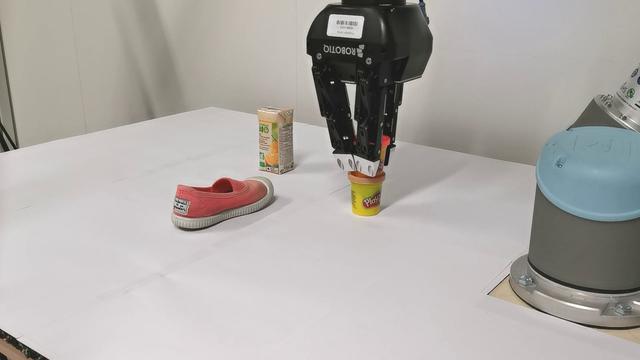}
  \includegraphics[trim=100 0 140 0, clip, width=.16\linewidth]{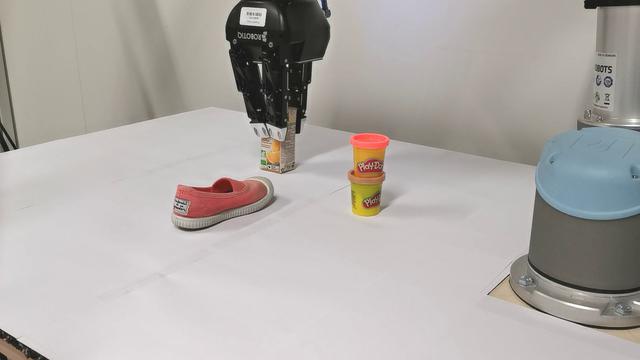}
  \includegraphics[trim=100 0 140 0, clip, width=.16\linewidth]{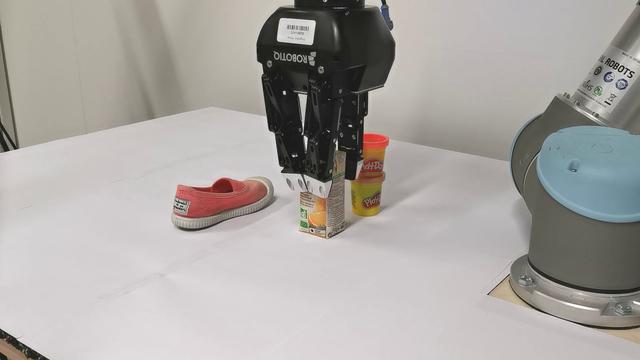}
  \includegraphics[trim=100 0 140 0, clip, width=.16\linewidth]{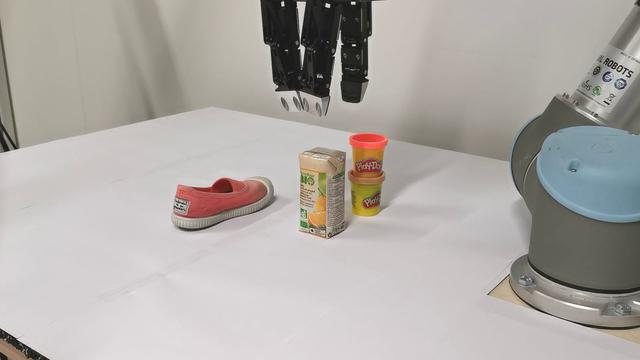}
  \includegraphics[trim=100 0 140 0, clip, width=.16\linewidth]{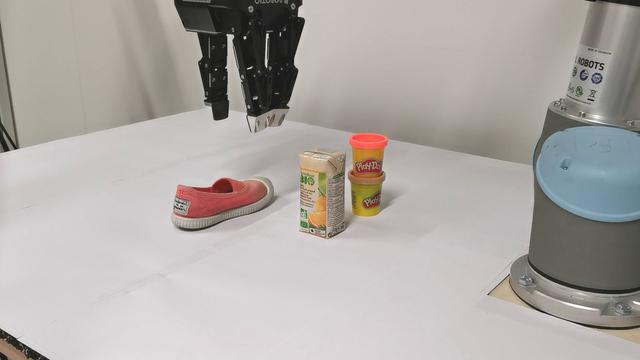}
  \includegraphics[trim=100 0 140 0, clip, width=.16\linewidth]{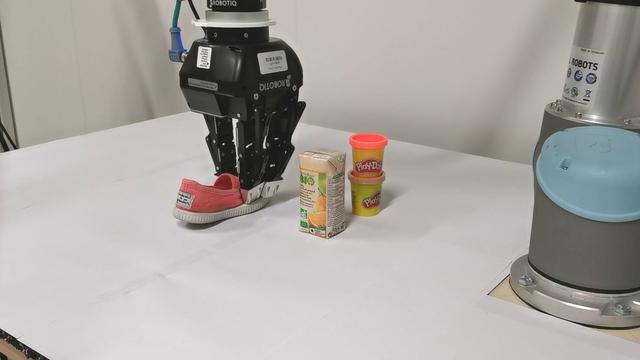}
  \includegraphics[trim=100 0 140 0, clip, width=.16\linewidth]{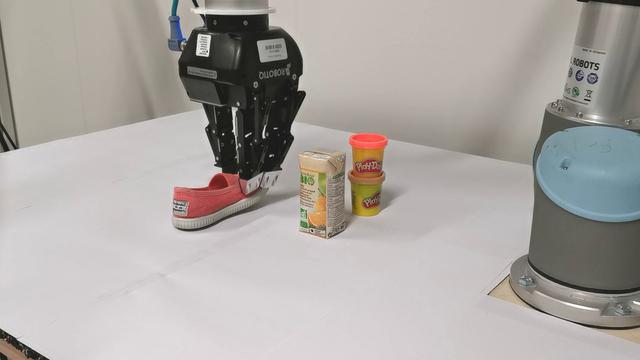}
  \includegraphics[trim=100 0 140 0, clip, width=.16\linewidth]{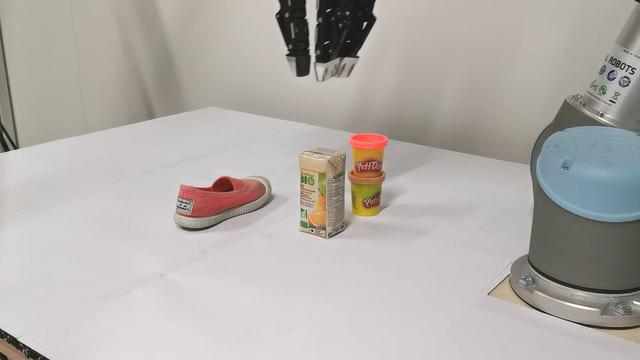}
  \caption{Failure cases for arch construction with a real robot. The failures originate from the wrong choice of a primitive (top example), wrong choice of the target location (middle example), and a failure to grasp a soft object (bottom example).}
  \label{fig:arch_failure}
\end{figure*}

Figures~\ref{fig:arch_blocks}-\ref{fig:arch_failure} demonstrate application of our method to the construction of arches on a real robot. Figure~\ref{fig:arch_blocks} presents the construction of arches from blocks. The blocks are initially arranged in random positions and orientations. The policy changes positions and orientations of blocks, leading to the successful construction of arches of different sizes.
Figure~\ref{fig:arch_new} demonstrates arch constructions from new primitives that have not been observed during training. We emphasize that our learned visual policies directly control the robot without intermediate geometric reconstruction of the world state. Nevertheless, the policy can handle diverse object shapes. The robustness to new shapes could be further improved by using various shapes for policy training in simulation. Finally, Figure~\ref{fig:arch_failure} illustrates few failure cases originating from incorrectly estimated pick and place locations as well as from a failure to grasp a soft object (a shoe).

\end{document}